\documentclass[twoside,11pt]{article}

\usepackage{color,graphicx}
\usepackage[dvipsnames]{xcolor}  

\usepackage{enumitem}  
\usepackage{changepage}  
\usepackage[normalem]{ulem}   

\usepackage[%
    font={small,sf},
    labelfont=bf,
    format=hang,    
    format=plain,
    margin=0pt,
    width=0.8\textwidth,
]{caption}
\usepackage[list=true]{subcaption}
\usepackage{wrapfig}  

\usepackage{floatrow}
\newfloatcommand{capbtabbox}{table}[][\FBwidth]

\usepackage{tabularx, booktabs}
\usepackage{colortbl}

\usepackage{amsmath}
\usepackage{amssymb, dsfont}  
\usepackage{mathtools} 
\usepackage{bm}   
\usepackage{commath}  
\usepackage{expl3, xparse}

\ExplSyntaxOn
\keys_define:nn { mbfinternal / mbf } {
  latin          .tl_set:N = \mbfinternal_mbf_latin_tl,
  greek          .tl_set:N = \mbfinternal_mbf_greek_tl,
  latin-alphabet .tl_set:N = \mbfinternal_mbf_latin_alphabet_tl,
  greek-alphabet .tl_set:N = \mbfinternal_mbf_greek_alphabet_tl,
}
\cs_new:Nn \mbfinternal_mbf:n {
  \tl_if_in:NnTF \mbfinternal_mbf_latin_alphabet_tl { #1 } {
    \mbfinternal_mbf_latin_tl { #1 }
  } {
    \tl_if_in:NnT \mbfinternal_mbf_greek_alphabet_tl { #1 } {
      \mbfinternal_mbf_greek_tl { #1 }
    }
  }
}
\NewDocumentCommand \mbf { O{} m } {
  \group_begin:
  \keys_set:nn { mbfinternal / mbf } {
    latin-alphabet = abcdefghijklmnopqrstuvwxyz
                     ABCDEFGHIJKLMNOPQRSTUVWXYZ
                     0123456789,
    greek-alphabet = \alpha\beta\delta\epsilon\zeta
                     \phi\gamma\eta\iota\theta
                     \kappa\lambda\mu\nu\pi\chi
                     \rho\sigma\tau\omega\xi\psi\xi
                     \Alpha\Beta\Delta\Epsilon  
                     \Phi\Gamma\Eta\Iota\Theta  
                     \Kappa\Lambda\Mu\Nu\Pi\Chi 
                     \Rho\Sigma\Tau\Omega\Xi\Psi\Xi\Zeta, 
    latin = \mathbf,
    greek = \boldsymbol,
    #1
  }
  \mbfinternal_mbf:n { #2 }
  \group_end:
}
\ExplSyntaxOff    

\usepackage[ruled]{algorithm2e}

\usepackage{tikz}
\usetikzlibrary{arrows, arrows.meta, calc}
\usetikzlibrary{backgrounds,automata,fit,shapes.geometric}
\definecolor{mtlhlcolor}{rgb}{0.494,0.286,0.714}
\definecolor{purplevis}{RGB}{185,164,219}
\definecolor{myblue}{RGB}{25,166,184}

\tikzset{%
    latentcts/.style={%
        line width=1.25pt,
        circle,minimum size=30pt,draw
    },
    gatenode/.style={%
        minimum size=30pt,draw, color=lstmcolgate, fill=lstmcolgate, text=white
    },
    latentctswhite/.style={%
        line width=1.25pt,
        circle,minimum size=30pt,draw,fill=white
    },
    latentdiscrete/.style={%
        line width=1.25pt,
        minimum size=30pt,draw
    },
    latentdeterministic/.style={%
        line width=1.25pt,
        diamond, minimum size=30pt,fill=white, draw, inner sep=2pt
    },
    indirectcts/.style={%
        circle,minimum size=30pt,fill=gray!30, draw
    },
    visiblects/.style={%
        line width=1.25pt,
        circle,minimum size=30pt,fill=purplevis!50!white, draw
    },
    visiblediscrete/.style={%
        minimum size=30pt,fill=purplevis!50!white, draw
    },
    visibledeterministic/.style={%
        line width=1.25pt,
        diamond, minimum size=30pt,fill=purplevis!50!white, draw
    },
    invisiblects/.style={%
        circle,minimum size=30pt, opacity=100
    },
    invis/.style={%
        minimum size=0pt,fill=white, inner sep=0pt
    },
    invistrans/.style={%
        minimum size=0pt, inner sep=0pt
    },
    blackdot/.style={%
        circle,minimum size=4pt,fill=black, inner sep=0pt
    },
    blackdotmed/.style={%
        circle,minimum size=6pt,fill=black, inner sep=0pt
    },
    nnnode/.style={%
        circle, draw, minimum size=8pt, inner sep=0pt
    },
    multnode/.style={%
        circle, minimum size=0pt,inner sep=0,outer sep=0,fill=white,draw
    },
    thickerarrow/.style={%
        line width=1.25pt,
        -{Latex[length=3mm]}, black
    },
    stdarrow/.style={%
        line width=1.25pt,
        -{Latex[length=2.8mm]}, black
    },
    stdline/.style={%
        line width=1.25pt,
        black
    },
    blackarrow/.style={%
        -{Latex[length=2mm]},line cap=round, black
    },
    bluearrow/.style={%
        -{Latex[length=2mm]},line cap=round, myblue
    },
    gatearrow/.style={%
        -{Latex[length=2mm]}, black,line cap=round, opacity=0.7, line width=1.0pt
    },
    gateline/.style={%
        black,line cap=round, opacity=0.7, line width=1.0pt
    },
    blacklinethick/.style={%
        black, line width=1.8pt,line cap=round
    },
    blackarrowthick/.style={%
        -{Latex[length=2mm]}, black,line cap=round, line width=1.8pt
    },
    blackarrowdashed/.style={%
        -{Latex[length=2mm]}, black,line cap=round, dashed
    },
    highlightfree/.style={%
        line width=1.8pt, color=mtlhlcolor
    },
    highlightfreearrow/.style={%
        -{Latex[length=2.8mm]}, line width=1.8pt, color=mtlhlcolor
    },
      declare function={
    atan3(\a,\b)=ifthenelse(atan2(0,1)==90, atan2(\a,\b), atan2(\b,\a));},
  kinky cross radius/.initial=+.125cm,
  @kinky cross/.initial=+, kinky crosses/.is choice,
  kinky crosses/left/.style={@kinky cross=-},kinky crosses/right/.style={@kinky cross=+},
  kinky cross/.style args={(#1)--(#2)}{
    to path={
      let \p{@kc@}=($(\tikztotarget)-(\tikztostart)$),
          \n{@kc@}={atan3(\p{@kc@})+180} in
      -- ($(intersection of \tikztostart--{\tikztotarget} and #1--#2)!%
             \pgfkeysvalueof{/tikz/kinky cross radius}!(\tikztostart)$)
      arc [ radius     =\pgfkeysvalueof{/tikz/kinky cross radius},
            start angle=\n{@kc@},
            delta angle=\pgfkeysvalueof{/tikz/@kinky cross}180 ]
      -- (\tikztotarget)}
     }
}

\makeatletter
\protected\def\specialmergetwolists{%
  \begingroup
  \@ifstar{\def\cnta{1}\@specialmergetwolists}
    {\def\cnta{0}\@specialmergetwolists}%
}
\def\@specialmergetwolists#1#2#3#4{%
  \def\tempa##1##2{%
    \edef##2{%
      \ifnum\cnta=\@ne\else\expandafter\@firstoftwo\fi
      \unexpanded\expandafter{##1}%
    }%
  }%
  \tempa{#2}\tempb\tempa{#3}\tempa
  \def\cnta{0}\def#4{}%
  \foreach \x in \tempb{%
    \xdef\cnta{\the\numexpr\cnta+1}%
    \gdef\cntb{0}%
    \foreach \y in \tempa{%
      \xdef\cntb{\the\numexpr\cntb+1}%
      \ifnum\cntb=\cnta\relax
        \xdef#4{#4\ifx#4\empty\else,\fi\x#1\y}%
        \breakforeach
      \fi
    }%
  }%
  \endgroup
}

\newcommand{\gettikzxy}[3]{%
  \tikz@scan@one@point\pgfutil@firstofone#1\relax
  \edef#2{\the\pgf@x}%
  \edef#3{\the\pgf@y}%
}
\makeatother

\newcommand{\citeay}[1]{\citeauthor{#1}, \citeyear{#1}}

\newcommand{\largeptfont}{\fontsize{14.5}{17}\selectfont}

\newcommand*{\Tr}{^{\mkern-1.5mu\mathsf{T}}}

\renewcommand{\eqref}[1]{eq.\ (\ref{#1})}
\renewcommand{\secref}[1]{Section \ref{#1}}

\newcommand{\defeq}{\vcentcolon=}

\newcommand{\R}{\mathbb{R}}
\newcommand{\E}{\mathbb{E}}

\newcommand{\Ls}{\mathcal{L}}

\newcommand{\x}{\mathbf{x}}
\newcommand{\y}{\mathbf{y}}
\newcommand{\z}{\mathbf{z}}
\newcommand{\bu}{\mathbf{u}}
\newcommand{\bv}{\mathbf{v}}
\newcommand{\bw}{\mathbf{w}}

\newcommand{\hphi}{\mathbf{h}_{\bfphi}}

\newcommand{\ii}{{(i)}}

\newcommand{\Normal}[1]{\mathcal{N}\left(#1\right)}

\newcommand{\Expect}[2][]{%
    \ifthenelse{\isempty{#1}}{%
        \mathbb{E}\left[#2\right]
    }{%
        \mathbb{E}_{#1}\left[#2\right]
    }
}
    
\newcommand{\KLDiv}[2]{\text{KL} \left(#1\, \middle\|\, #2 \right)}

\DeclareMathSymbol{\shortminus}{\mathbin}{AMSa}{"39}

\usepackage{xifthen}

\newcommand{\bfpsi}{{\boldsymbol{\psi}}}

\newcommand{\bfphi}{{\boldsymbol{\phi}}}
\newcommand{\bftheta}{{\boldsymbol{\theta}}}

\newcommand{\bfeta}{{\boldsymbol{\eta}}}

\newcommand{\bflambda}{{\boldsymbol{\lambda}}}
\newcommand{\bfbeta}{{\boldsymbol{\beta}}}

\newcommand{\Dtrain}{\mathcal{D}}

\newcommand{\Tenc}{T_{\textrm{\scriptsize enc}}}

\definecolor{medgrey}{rgb}{0.7,0.7,0.7}

\definecolor{tab10_1}{rgb}{1.0, 0.49804, 0.05490}
\definecolor{tab10_2}{rgb}{0.17255, 0.62745, 0.17255}
\definecolor{tab10_3}{rgb}{0.8392, 0.1529, 0.1569}
\definecolor{tab10_7_orig}{rgb}{0.498, 0.498, 0.498}
\definecolor{tab10_7}{rgb}{0.37, 0.37, 0.37}

\newcommand{\xtn}[2][]{%
    \ifthenelse{\isempty{#2}}{%
    \ifthenelse{\isempty{#1}}{x_{t}^{(n)}}{x_{#1}^{(n)}}
    }{
    \ifthenelse{\isempty{#1}}{\mathbf{x}_{t}^{(n)}}{\mathbf{x}_{#1}^{(n)}}
    }}
\newcommand{\ytn}[2][]{%
    \ifthenelse{\isempty{#2}}{%
    \ifthenelse{\isempty{#1}}{y_{t}^{(n)}}{y_{#1}^{(n)}}
    }{
    \ifthenelse{\isempty{#1}}{\mathbf{y}_{t}^{(n)}}{\mathbf{y}_{#1}^{(n)}}
    }}
\newcommand{\utn}[2][]{%
    \ifthenelse{\isempty{#2}}{%
    \ifthenelse{\isempty{#1}}{u_{t}^{(n)}}{u_{#1}^{(n)}}
    }{
    \ifthenelse{\isempty{#1}}{\mathbf{u}_{t}^{(n)}}{\mathbf{u}_{#1}^{(n)}}
    }}
\newcommand{\epstn}[2][]{%
    \ifthenelse{\isempty{#2}}{%
    \ifthenelse{\isempty{#1}}{\epsilon_{t}^{(n)}}{\epsilon_{#1}^{(n)}}
    }{
    \ifthenelse{\isempty{#1}}{\boldsymbol{\epsilon}_{t}^{(n)}}{\boldsymbol{\epsilon}_{#1}^{(n)}}
    }}

\newcommand{\zn}[2][]{%
    \ifthenelse{\isempty{#2}}{%
    \ifthenelse{\isempty{#1}}{z^{(n)}}{z_{#1}^{(n)}}
    }{
    \ifthenelse{\isempty{#1}}{\mathbf{z}^{(n)}}{\mathbf{z}_{#1}^{(n)}}
    }}

\usepackage{mathtools}
\DeclarePairedDelimiterX{\infdivx}[2]{(}{)}{%
  #1\;\delimsize|\delimsize|\;#2%
}

\newcommand{\captiona}{{(a)}}
\newcommand{\captionb}{{(b)}}
\newcommand{\captionc}{{(c)}}

\newcommand{\textcaptiona}{{a}}
\newcommand{\textcaptionb}{{b}}
\newcommand{\textcaptionc}{{c}}

\def\gD{{\mathcal{D}}}

\def\gO{{\mathcal{O}}}

\def\gQ{{\mathcal{Q}}}

\def\gX{{\mathcal{X}}}

\def\gZ{{\mathcal{Z}}}

\definecolor{tab10_2wash}{rgb}{0.655,0.906,0.655}
\definecolor{tab10_3wash}{rgb}{0.969,0.82,0.82}

\usepackage{pifont}
\newcommand{\cmark}{\ding{51}}%
\newcommand{\xmark}{\ding{55}}%

\newcommand{\sublambda}[2]{%
    \mathchoice%
        {#1_{\raisebox{#2}{\scalebox{0.7}{$\displaystyle\bflambda$}}}}%
        {#1_{\raisebox{#2}{\scalebox{0.7}{$\textstyle\bflambda$}}}}%
        {#1_{\raisebox{#2}{\scalebox{0.7}{$\scriptstyle\bflambda$}}}}%
        {#1_{\raisebox{#2}{\scalebox{0.7}{$\scriptscriptstyle\bflambda$}}}}%
}
\newcommand{\stdlam}{\sublambda{\mathbf{s}}{-1.5pt}}
\newcommand{\mulam}{\sublambda{\boldsymbol{\mu}}{-0.8pt}}
\newcommand{\qlam}{\sublambda{q}{-2pt}}

\newcommand{\subother}[3]{%
    \mathchoice%
        {#1_{\raisebox{#2}{\scalebox{0.7}{$\displaystyle#3$}}}}%
        {#1_{\raisebox{#2}{\scalebox{0.7}{$\textstyle#3$}}}}%
        {#1_{\raisebox{#2}{\scalebox{0.7}{$\scriptstyle#3$}}}}%
        {#1_{\raisebox{#2}{\scalebox{0.7}{$\scriptscriptstyle#3$}}}}%
}

\newcommand{\ggamma}{\subother{g}{-0.95pt}{\bfeta}}
\newcommand{\ggammaj}{\subother{g}{-1.3pt}{\bfeta_j}}
\newcommand{\ggammaus}[1]{\subother{g}{-1.3pt}{\bfeta_{#1}}}

\usepackage{admin/jmlr2e}
\usepackage{microtype}
\usepackage[T1]{fontenc}

\usepackage{lastpage}
\jmlrheading{23}{2022}{1-\pageref{LastPage}}{12/20; Revised 12/21}{8/22}{20-1446}{Alex Bird, Christopher K.\ I.\ Williams and Christopher Hawthorne}

\ShortHeadings{Multi-Task Dynamical Systems}{Bird, Williams and Hawthorne}
\firstpageno{1}

\begin{document}

\title{Multi-Task Dynamical Systems}

\author{\name Alex Bird \email mtds@alxbrd.com \\
       \addr School of Informatics, University of Edinburgh,\\
       Edinburgh, EH8 9AB, UK \\
       and The Alan Turing Institute, London, NW1 2DB, UK
       \AND
       \name Christopher K.\ I.\ Williams \email ckiw@inf.ed.ac.uk \\
       \addr School of Informatics, University of Edinburgh,\\
       Edinburgh, EH8 9AB, UK \\
       and The Alan Turing Institute, London, NW1 2DB, UK
       \AND
       \name Christopher Hawthorne \email christopher.hawthorne@glasgow.ac.uk \\
       \addr Institute of Neurological Sciences, \\
       Queen Elizabeth University Hospital, Glasgow, G52 4TF, UK \\
       and Academic Unit of Anaesthesia, \\
       University of Glasgow, Glasgow, G31 2ER, UK}
\vfill

\editor{Samuel Kaski}

\maketitle

\begin{abstract}%
Time series datasets are often composed of a variety of sequences from the same domain, but from different entities, such as individuals, products, or organizations. We are interested in how time series models can
be specialized to individual sequences (capturing the specific
characteristics) while still retaining statistical power by sharing
commonalities across the sequences. This paper describes the multi-task dynamical system (MTDS); a general methodology for extending multi-task learning (MTL) to time series models. Our approach endows dynamical systems with a set of hierarchical latent variables which can modulate \emph{all} model parameters. To our knowledge, this is a novel development of MTL, and applies to time series both with and without control inputs. We apply the MTDS to motion-capture data of people walking in various styles using a multi-task recurrent neural network (RNN), and to patient drug-response data using a multi-task pharmacodynamic model.
\end{abstract}


\section{Introduction} \label{sec:intro}

Perhaps the most important class of time series models today are dynamical systems, which encompass a wide variety of models. Applications may be found in domains as diverse as physical modelling \citep[see e.g.][]{linderman2017bayesian}, drug response \citep[e.g.][]{white2008use},  public transport demand forecasting \citep[e.g.][]{toque2016forecasting}, motion capture \citep[e.g.][]{martinez2017human}, and retail sales data \citep[e.g.][]{rangapuram2018deep} to name but a few. Since such time series data can arise from various sources (such as different people, systems, locations or organizations), each data set often comprises a variety of sequences with different characteristics. For instance, motion capture data might include different styles of walking, and healthcare data might exhibit a variety of personalized responses to the same drug. These characteristic differences often require different dynamics, which a single dynamical system is unable to provide, at least explicitly. In this paper, we describe how dynamical systems can be modified to adapt to this inter-sequence variation via the use of a set of hierarchical latent variables, thus enabling `personalization' or `customization' of the models.

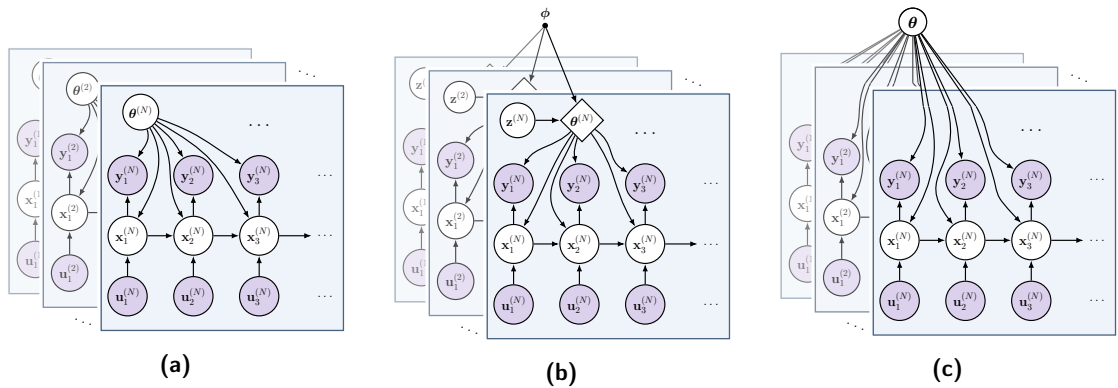
\begin{figure}
    \centering
    \begin{subfigure}[t!]{0.32\textwidth}
        \centering
        \scalebox{0.35}{\begin{tikzpicture}

\def\numseq{3}
\def\numvars{2}  

\pgfmathtruncatemacro{\numseqm}{\numseq - 1}
\pgfmathtruncatemacro{\numvarsm}{\numvars - 1}

\newcommand{\dotsangle}{-20}
\newcommand{\dotsxshift}{.2ex}
\newcommand{\dotsyshift}{1ex}
\newcommand{\rdots}{\hspace{\dotsxshift}%
    \raisebox{\dotsyshift}{\rotatebox{\dotsangle}{$\ddots$}}}

\node [invis] (phi) at (2.5,9.2){};

\specialmergetwolists{/}{1,2,N}{0,1.3,3.5}\ziplist
\foreach \curix/\curx in \ziplist{
    \pgfmathsetmacro\cury{\curx *0.4}
    \fill[white, opacity=0.3] (-1.2, -1.5-\cury) rectangle (2.5*\numseq + 0.7 + \curx, 9.4);
    
    \draw[color=white, fill=RoyalBlue!5!white, line width=1mm] (1*\curx -1.1, -1.3-\cury) rectangle ++(2.5*\numseq + 1.8,9.4);
    \draw[color=RoyalBlue!50!Black, fill=RoyalBlue!5!white, line width=0.5mm] (1*\curx -1.0, -1.3-\cury) rectangle ++(2.5*\numseq + 1.7,9.3);
  \foreach \x in {0,...,\numseq}      
    \foreach \y in {0,...,\numvars} 
       {\pgfmathtruncatemacro{\label}{\x + 1}
       \ifnum \y = 0
            \def\varlbl{\bu}
            \def\tzndstyle{visiblects}
        \else
            \ifnum \y = 1
                \def\varlbl{\mbf x}
                \def\tzndstyle{latentctswhite}
            \else  
                \def\varlbl{\mbf y}
                \def\tzndstyle{visiblects}
            \fi
        \fi
        \def\tzndlbl{\largeptfont $\varlbl_{\label}^{(\curix)}$}
        \def\tzndxpos{2.5*\x}
        
        \ifnum \x = \numseq
            \def\tzndlbl{}
            \def\tzndstyle{invis}
            \def\tzndxpos{2.5*\x - 0.5}
        \fi
        
       \node [\tzndstyle]  (\y\x\curix) at (\tzndxpos + 1*\curx, 2.3*\y -\cury) {\tzndlbl}
       ;} 
    
      \foreach \y  in {0,...,\numvars}
        \node[anchor=center] at (2.5*\numseq + 1*\curx, 2.3*\y-\cury) {\largeptfont $\ldots$};
        
  \foreach \y in {1,...,1}  
    \foreach \x in {0,...,\numseqm}
      \pgfmathtruncatemacro{\xp}{\x + 1}%
      \draw[stdarrow] (\y\x\curix)--(\y\xp\curix)
      ;
      
  \foreach \y in {1,...,\numvars}  
    \foreach \x in {0,...,\numseqm}
      \pgfmathtruncatemacro{\ym}{\y - 1}%
      \draw[stdarrow] (\ym\x\curix)--(\y\x\curix)
      ;
    
    \node [latentctswhite]  (psi\curix) at (0.5 + 1*\curx, 7.0- \cury){\largeptfont $\bftheta^{(\curix)}$};
    \foreach \y in {1,...,\numvars}
      \foreach \x in {0,...,2}
        \ifthenelse{\x>0}{\def\looseamt{1.2}}{\def\looseamt{1.0}}
        \ifthenelse{\x>0}{\def\inangle{120}}{\def\inangle{60}}
        \ifthenelse{\equal{\x}{2} \AND \equal{\y}{2}}{\def\inangle{148}}{}
        \ifthenelse{\equal{\x}{0}}{\def\outangle{290+\x*20-\y*10+20}}{\def\outangle{290+\x*20+\ym*10}}
        \pgfmathtruncatemacro{\ym}{\y - 1}%
        \draw[stdarrow] (psi\curix) to[out=\outangle, in=\inangle, looseness=\looseamt] (\y\x\curix);
        
    \node [invistrans] (dotspsi) at (5.0+\curx, 6.5-\cury) {\huge $\ldots$};
 }

\node [invistrans, rotate=20] (dotsd1) at (1.8, -2.3) {\huge $\rdots$};
\node [invistrans, rotate=20] (dotsd2) at (2.8+\numseq*2.5, 7.13) {\huge $\rdots$};

\end{tikzpicture}}
        \caption{}
        \label{fig:MTDS-stl}
    \end{subfigure}%
    ~
    \begin{subfigure}[t!]{0.32\textwidth}
        \centering
        \scalebox{0.35}{\begin{tikzpicture}

\def\numseq{3}
\def\numvars{2}  

\pgfmathtruncatemacro{\numseqm}{\numseq - 1}
\pgfmathtruncatemacro{\numvarsm}{\numvars - 1}

\newcommand{\dotsangle}{-20}
\newcommand{\dotsxshift}{.2ex}
\newcommand{\dotsyshift}{1ex}
\newcommand{\rdots}{\hspace{\dotsxshift}%
    \raisebox{\dotsyshift}{\rotatebox{\dotsangle}{$\ddots$}}}

\node [blackdotmed] (phi) at (4.7,9.2){};

\specialmergetwolists{/}{1,2,N}{0,1.3,3.5}\ziplist
\foreach \curix/\curx in \ziplist{
    \pgfmathsetmacro\cury{\curx *0.4}
    \fill[white, opacity=0.3] (-1.2, -1.5-\cury) rectangle (2.5*\numseq + 0.7 + \curx, 9.4);
    
    \draw[color=white, fill=RoyalBlue!5!white, line width=1mm] (1*\curx -1.1, -1.3-\cury) rectangle ++(2.5*\numseq + 1.8,9.4);
    \draw[color=RoyalBlue!50!Black, fill=RoyalBlue!5!white, line width=0.5mm] (1*\curx -1.0, -1.3-\cury) rectangle ++(2.5*\numseq + 1.7,9.3);
  \foreach \x in {0,...,\numseq}      
    \foreach \y in {0,...,\numvars} 
       {\pgfmathtruncatemacro{\label}{\x + 1}
       \ifnum \y = 0
            \def\varlbl{\bu}
            \def\tzndstyle{visiblects}
        \else
            \ifnum \y = 1
                \def\varlbl{\mbf x}
                \def\tzndstyle{latentctswhite}
            \else  
                \def\varlbl{\mbf y}
                \def\tzndstyle{visiblects}
            \fi
        \fi
        \def\tzndlbl{\largeptfont $\varlbl_{\label}^{(\curix)}$}
        \def\tzndxpos{2.5*\x}
        
        \ifnum \x = \numseq
            \def\tzndlbl{}
            \def\tzndstyle{invis}
            \def\tzndxpos{2.5*\x - 0.5}
        \fi
        
       \node [\tzndstyle]  (\y\x\curix) at (\tzndxpos + 1*\curx, 2.3*\y -\cury) {\tzndlbl}
       ;} 
    
      \foreach \y  in {0,...,\numvars}
        \node[anchor=center] at (2.5*\numseq + 1*\curx, 2.3*\y-\cury) {\largeptfont $\ldots$};
        
  \foreach \y in {1,...,1}  
    \foreach \x in {0,...,\numseqm}
      \pgfmathtruncatemacro{\xp}{\x + 1}%
      \draw[stdarrow] (\y\x\curix)--(\y\xp\curix)
      ;
      
  \foreach \y in {1,...,\numvars}  
    \foreach \x in {0,...,\numseqm}
      \pgfmathtruncatemacro{\ym}{\y - 1}%
      \draw[stdarrow] (\ym\x\curix)--(\y\x\curix)
      ;
    
    \node [latentctswhite]  (z\curix) at (0.15 + 1*\curx, 7.0- \cury){\largeptfont $\z^{(\curix)}$};
    \node [latentdeterministic]  (psi\curix) at (2.6 + 1*\curx, 7.0- \cury){\largeptfont $\bftheta^{(\curix)}$};
    \draw[thickerarrow] (z\curix) to (psi\curix);
    \foreach \y in {1,...,\numvars}
      \foreach \x in {0,...,2}
        \ifthenelse{\x>0}{\def\looseamt{1.2}}{\def\looseamt{1.0}}
        \ifthenelse{\equal{\x}{0}}{\def\inangle{60}}{
            \ifthenelse{\equal{\x}{1}}{\def\inangle{130}}{
                \def\inangle{120}}}
        \ifthenelse{\equal{\x}{1} \AND \equal{\y}{2}}{\def\inangle{100}}{}
        \ifthenelse{\equal{\x}{2} \AND \equal{\y}{2}}{\def\inangle{148}}{}
        \ifthenelse{\equal{\x}{0}}{\def\outangle{225+\x*20-\y*10+20}}{
            \ifthenelse{\equal{\x}{1}}{\def\outangle{225+\x*20- 30 + \y*10+20}}{
                \def\outangle{225+\x*40+\ym*20}}}
        \pgfmathtruncatemacro{\ym}{\y - 1}%
        \draw[stdarrow] (psi\curix) to[out=\outangle, in=\inangle, looseness=\looseamt] (\y\x\curix);
        
    \node [invistrans] (dotspsi) at (5.0+\curx, 6.5-\cury) {\huge $\ldots$};
    
    \pgfmathtruncatemacro{\hierangle}{45 + 18*\curx};
    \draw[thickerarrow] (phi) to (psi\curix.\hierangle);
 }

\node [blackdotmed, label=above:{\largeptfont $\bfphi$}] (phisolid) at (4.7,9.2){};
    
\node [invistrans, rotate=20] (dotsd1) at (1.8, -2.3) {\huge $\rdots$};
\node [invistrans, rotate=20] (dotsd2) at (2.8+\numseq*2.5, 7.13) {\huge $\rdots$};

\end{tikzpicture}}
        \caption{}
        \label{fig:MTDS-MTDS}
    \end{subfigure}%
    ~
    \begin{subfigure}[t!]{0.32\textwidth}
        \centering
        \scalebox{0.35}{\begin{tikzpicture}

\def\numseq{3}
\def\numvars{2}  

\pgfmathtruncatemacro{\numseqm}{\numseq - 1}
\pgfmathtruncatemacro{\numvarsm}{\numvars - 1}

\newcommand{\dotsangle}{-20}
\newcommand{\dotsxshift}{.2ex}
\newcommand{\dotsyshift}{1ex}
\newcommand{\rdots}{\hspace{\dotsxshift}%
    \raisebox{\dotsyshift}{\rotatebox{\dotsangle}{$\ddots$}}}

\node [latentcts] (phi) at (4,9.2){\LARGE $\bftheta$};

\specialmergetwolists{/}{1,2,N}{0,1.3,3.5}\ziplist
\foreach \curix/\curx in \ziplist{
    \pgfmathsetmacro\cury{\curx *0.4}
    \fill[white, opacity=0.3] (-1.2, -1.5-\cury) rectangle (2.5*\numseq + 0.7 + \curx, 9.4);
    
    \draw[color=white, fill=RoyalBlue!5!white, line width=1mm] (1*\curx -1.1, -1.3-\cury) rectangle ++(2.5*\numseq + 1.8,9.4);
    \draw[color=RoyalBlue!50!Black, fill=RoyalBlue!5!white, line width=0.5mm] (1*\curx -1.0, -1.3-\cury) rectangle ++(2.5*\numseq + 1.7,9.3);
  \foreach \x in {0,...,\numseq}      
    \foreach \y in {0,...,\numvars} 
       {\pgfmathtruncatemacro{\label}{\x + 1}
       \ifnum \y = 0
            \def\varlbl{\bu}
            \def\tzndstyle{visiblects}
        \else
            \ifnum \y = 1
                \def\varlbl{\mbf x}
                \def\tzndstyle{latentctswhite}
            \else  
                \def\varlbl{\mbf y}
                \def\tzndstyle{visiblects}
            \fi
        \fi
        \def\tzndlbl{\largeptfont $\varlbl_{\label}^{(\curix)}$}
        \def\tzndxpos{2.5*\x}
        
        \ifnum \x = \numseq
            \def\tzndlbl{}
            \def\tzndstyle{invis}
            \def\tzndxpos{2.5*\x - 0.5}
        \fi
        
       \node [\tzndstyle]  (\y\x\curix) at (\tzndxpos + 1*\curx, 2.3*\y -\cury) {\tzndlbl}
       ;} 
    
      \foreach \y  in {0,...,\numvars}
        \node[anchor=center] at (2.5*\numseq + 1*\curx, 2.3*\y-\cury) {\largeptfont $\ldots$};
        
  \foreach \y in {1,...,1}  
    \foreach \x in {0,...,\numseqm}
      \pgfmathtruncatemacro{\xp}{\x + 1}%
      \draw[stdarrow] (\y\x\curix)--(\y\xp\curix)
      ;
      
  \foreach \y in {1,...,\numvars}  
    \foreach \x in {0,...,\numseqm}
      \pgfmathtruncatemacro{\ym}{\y - 1}%
      \draw[stdarrow] (\ym\x\curix)--(\y\x\curix)
      ;
    
  \def\ydelta{0.4}
  \foreach \y in {1,...,\numvars}
    \node [invis]  (psi\y0\curix) at (0.9 + 1*\curx + 0.2*2-0.2*\y, 7.0 - \cury + \ydelta){};
    
  \foreach \y in {1,...,\numvars}
    \node [invis]  (psi\y1\curix) at (1.2 + 1*\curx + 0.4 + 0.2*\y, 7.0- \cury + \ydelta){};
    
  \foreach \y in {1,...,\numvars}
    \node [invis]  (psi\y2\curix) at (1.5 + 1*\curx + 0.8 +0.2 + 0.2*\y, 7.0- \cury + \ydelta){};
        
    \foreach \y in {1,...,\numvars}
      \foreach \x in {0,...,2}
        \ifthenelse{\x>0}{\def\looseamt{1.4}}{\def\looseamt{1.0}}
        \ifthenelse{\x>0}{\def\inangle{120}}{\def\inangle{60}}
        \ifthenelse{\equal{\curix}{N}}{\def\outangle{285}}{\def\outangle{245}} 
        \ifthenelse{\equal{\x}{1} \AND \equal{\y}{1}}{\def\inangle{128}}{}
        \ifthenelse{\equal{\x}{1} \AND \equal{\y}{2}}{\def\inangle{100}}{}
        \ifthenelse{\equal{\x}{2} \AND \equal{\y}{1}}{\def\inangle{148}}{}
        \ifthenelse{\equal{\x}{2} \AND \equal{\y}{1}}{\def\outangle{300}}{}
        \ifthenelse{\equal{\x}{2} \AND \equal{\y}{2}}{\def\inangle{148}}{}
        \ifthenelse{\equal{\x}{2} \AND \equal{\y}{2}}{\def\outangle{300}}{}
        \pgfmathtruncatemacro{\ym}{\y - 1}%
        \ifthenelse{\equal{\y}{1} \AND \equal{\x}{2}}{
            \draw[stdarrow] plot [smooth] coordinates {(psi\y\x\curix)  (7,3.5) (7.3,2.5) (\y\x\curix.120)}
          }{
            \draw[stdarrow] (psi\y\x\curix) to[out=\outangle, in=\inangle, looseness=\looseamt] (\y\x\curix)
        };

    \foreach \y in {1,...,\numvars}
      \foreach \x in {0,...,2}
        \draw[stdline] (phi) to (psi\y\x\curix);

    \node [invistrans] (dotspsi) at (5.0+\curx, 6.5-\cury) {\huge $\ldots$};
 }

\node [latentcts] (phisolid) at (4,9.2){\LARGE $\bftheta$};
    
\node [invistrans, rotate=20] (dotsd1) at (1.8, -2.3) {\huge $\rdots$};
\node [invistrans, rotate=20] (dotsd2) at (2.8+\numseq*2.5, 7.13) {\huge $\rdots$};

\end{tikzpicture}}
        \caption{}
        \label{fig:MTDS-DS}
    \end{subfigure}%
    \caption{Comparison of approaches for modelling multiple sequences using dynamical systems with inputs $\bu$, observations $\y$, hidden states $\x$, task variables  $\z$; and dynamical system parameters $\bftheta$, multi-task parameters $\bfphi$. Panels show \captiona\ a single task approach, where each sequence is learned separately; \captionb\ the multi-task dynamical system; \captionc\ the pooled approach.}
    \label{fig:MTDS}
\end{figure}

There are two common approaches\footnote{We address more sophisticated approaches in the related work in Section \ref{sec:MTDS:related}.} to modelling such inter-sequence variation: the most flexible option is to train an individual model per sequence (as per the graphical model in Figure \ref{fig:MTDS}\textcaptiona). An individual model can in principle capture idiosyncratic features, but suffers from overfitting and fails to exploit the regularities between the sequences in the training set. More commonly, the different sequences are \emph{pooled} together to train a single dynamical system, despite the inter-sequence variation (a one-size-fits-all approach, Figure \ref{fig:MTDS}\textcaptionc). This may fail to capture the idiosyncratic features at all; a simple model such as a linear dynamical system (LDS) will learn only an average effect.

In contrast to these approaches, we aim to learn a \emph{family} of dynamical systems that is consistent with the sequences in the training set. Each training sequence is given a bespoke model from the `sequence family', but this family exploits the regularities observed across all sequences, which can substantially reduce overfitting.
We achieve this via use of a set of hierarchical latent variables, similar to the approach used in multi-task learning for iid data, now with each \emph{sequence} treated as a task. We hence call this approach the \emph{multi-task dynamical system} (MTDS). The MTDS learns a low dimensional manifold in parameter space, indexed by a latent code $\z \in \gZ$, which corresponds to the specialization of each sequence model. See Figure \ref{fig:MTDS}\textcaptionb\ for a graphical model. 

The choice of a low dimensional manifold enables the MTDS to determine directions in parameter space with respect to which the parameters $\bftheta \in \Theta$ \emph{need not change}, sharing strength across sequences. This avoids considering directions in $\Theta$ which result in unlikely or uncharacteristic predictions, as well as ignoring so-called `sloppy directions' \citep{transtrum2011geometry}, which can add great complexity to inference, with little benefit in terms of fit or generalization. But it can also find the key direction(s) of variability in $\Theta$ between training sequences. As a simple example, consider sequences generated from the family of $d$th order linear ODEs. Such sequences may vary in terms of oscillation frequencies, magnitude, half life to an input impulse, etc. A good approximation of these sequences is possible via a $d$-dimensional LDS \citep[see e.g.][\S2.2, \S5.3]{astrom2010feedback}, but while the ODE has $d{+}1$ degrees of freedom, the LDS has $\gO(d^2)$ in general. An MTDS approach can \emph{learn} the relevant $d{+}1$ degrees of freedom of the LDS in $\Theta$ simply by training on example sequences. For an example of this, see \S4, \citet{bird21thesis}.

Unlike previous efforts to customize time series models, our MTDS construction allows application to general classes of dynamical systems. This is available due to our choice of learning an \emph{arbitrary manifold} over \emph{all} the model parameters (both system and emission), rather than being constrained to pre-determined parameter sets (see the related work in Section \ref{sec:MTDS:related}). This allows the MTDS to model variability in observation space (such as magnitude or offset), as well as different responses to inputs, different sensitivity to initial conditions, and differences in dynamic evolution.
The MTDS can thus be applied to classical ARMA models, state space models, and modern recurrent neural networks (RNNs). 

The MTDS provides user visibility of the task specialization, as well as the ability to control it directly if desired, via use of the hierarchical latent variable $\z$. This stands in contrast to a RNN approach, which also has the flexibility to model inter-sequence variation, but does so in an implicit and opaque `black-box' manner. This prohibits interpretability and end-user control, but can also suffer from mode drift where the personalization erroneously changes over time. For example, in \citet{ghosh2017learning} a RNN generates a motion capture (mocap) sequence which performs an unprompted transition from walking to drinking. Our contributions with respect to such end-user control will be explored further in the empirical work, especially in \secref{sec:mocap}.

In this paper we propose the MTDS, which endows dynamical systems with the ability to adapt to inter-sequence variation. Our contributions go beyond existing work by: (i) allowing the full adaptation of all parameters of general dynamical systems via use of a learned nonlinear manifold; (ii) describing efficient general purpose forms of learning and inference; (iii) performing in-depth experimental studies using data from human locomotion data and medical time series; and (iv) investigating the advantages of our proposal, which---compared to standard single-task and pooled approaches---include substantial improvements in data efficiency, robustness to dataset-shift, user control (which may be used for style transfer, for example), and online customization of interpretable models.\medskip

In what follows, Section \ref{sec:MTDS} introduces the MTDS, including a discussion of learning and inference, and \secref{sec:MTDS:related} provides an overview of related work. We provide two in-depth case studies: a multi-task RNN in \secref{sec:mocap} for a mocap data application, and a multi-task pharmacodynamic model in \secref{sec:mtpd} for personalized drug response modelling.

\section{The Multi-Task Dynamical System} \label{sec:MTDS}

In this section, we define the multi-task dynamical system (\secref{sec:MTDS:def}) and provide methods for learning (\secref{sec:MTDS:learning}) and inference (\secref{sec:MTDS:inference}).

\subsection{Definition} \label{sec:MTDS:def}

Consider a collection of $N$ input-output sequences $\Dtrain = \big\{\mkern2mu(\mkern1mu\bu_{1:T_i}^\ii,\; \y_{1:T_i}^\ii\mkern1mu)\mkern2mu\big\}_{i=1}^N$ consisting of inputs and outputs respectively, where $T_i$ is the length of sequence $i$.  Each sequence $i$ is described by a different dynamical system with state variables $\x_t^\ii \in \gX$, $t=1,\ldots,T_i$ whose parameter $\bftheta^\ii \in \Theta$ depends on the hierarchical latent variable $\z^\ii \in \gZ$. See Figure \ref{fig:MTDS-MTDS} for a graphical model.\footnote{The dimensions of each variable are denoted: $\dim(\bu)=n_u$, $\dim(\x) = n_x$ and $\dim(\y)=n_y$.} The MTDS is defined by the equations:
\begin{align}
   \bftheta^\ii &\;=\; \hphi(\z^\ii), \quad  \z^\ii \;\sim\; p(\z) \label{eq:mtds-z}\\
    \x_t^\ii &\;\sim\; p(\x \mid\; \x_{t-1}^\ii,\; \bu_t^\ii,\; \bftheta^\ii), \label{eq:mtds-x} \\ 
    \y_t^\ii &\;\sim\; p(\y \mid\; \x_t^\ii,\; \bu_t^{(i)},\; \bftheta^\ii), \label{eq:mtds-y}
\end{align}
for $t=1,\ldots,T_i$ for each $i=1,\ldots,N$, where eq.\ (\ref{eq:mtds-x}) represents the dynamic evolution and is sometimes referred to as the \emph{system equation}, and \eqref{eq:mtds-y} is the \emph{emission equation}.
In this paper we assume $\x_0 \defeq \mbf 0$, and $\gZ = \R^k$ which the vector-valued function $\hphi(\cdot)$ transforms to conformable model parameters $\bftheta \in \R^d$. By restricting the parameter manifold to $k \ll d$ dimensions, eqs.\ (\ref{eq:mtds-z}-\ref{eq:mtds-y}) result in a multi-task model rather than simply a hierarchical model. 

The MTDS model thus requires the specification of three key quantities:
\begin{enumerate}  
    \item The base model (eqs.\ \ref{eq:mtds-x}-\ref{eq:mtds-y}), such as a LDS or RNN.
    \item The dimensions of $\bftheta$ that depend on the latent variable $\z$ (for example, one might choose the emissions (eq.\ \ref{eq:mtds-y}) to be constant wrt.\ $\z$, or to modulate only the offset/bias terms of eqs.\ (\ref{eq:mtds-x}) and (\ref{eq:mtds-y}) wrt.\ $\z$).
    \item The choice of prior $p(\bftheta)$, that is, the choice of distribution $p(\z)$ and transformation $\hphi$.
\end{enumerate}

\noindent
We restrict our focus to deterministic state dynamical systems which simplifies the exposition and is motivated by our focus on longer term prediction. In our experience we have found deterministic models generally outperform stochastic state models for long-term prediction, see also (for example) \citet{bengio2015scheduled, chiappa2017recurrent}, and additional comments in Appendix \ref{appdx:mtds:deterministic}. Some example choices of base model are elaborated upon in Sections \ref{sec:mocap:expmt-setup:models:mtds} and \ref{sec:mtpd:model:singletask}. Note that where the dynamics of \eqref{eq:mtds-x} are linear, some care must be given to ensuring the stability of the system \citep[see][\S3.2]{bird21thesis}.

The choice of prior $p(\bftheta)$ considered in this paper is a (nonlinear) factor analysis model, described by:
\begin{align}
    \bftheta \,=\, \hphi(\z), \qquad \qquad p(\z) \,=\, \Normal{0, I},
\end{align}
where $\hphi$ is some deterministic function. An affine $\hphi$ may be useful where interpretability is important, or base models are highly flexible. Non-affine functions, such as multilayer perceptrons \citep[cf.][]{kingma2014vae, rezende2014stochastic} have proved to be important when using relatively inflexible base models such as LDSs. See Appendix \ref{appdx:mtds:prior} for some additional considerations.

\subsection{Learning} \label{sec:MTDS:learning}
The parameters $\bfphi$ of an MTDS can be learned from a data set $\Dtrain \defeq \{Y^\ii, U^\ii\}_{i=1}^N$, defining $Y \defeq \y_{1:T},\; U \defeq \bu_{1:T}$ to reduce the notational burden. We can learn the parameters via maximum marginal likelihood:
\begin{align}
\bfphi^* \;&=\; \arg\max_\bfphi\, \sum_{i=1}^N \log p(Y^\ii\, |\, U^\ii, \bfphi),
\intertext{where}
\log p(Y \, |\, U, \bfphi) \;&=\; \log \int_\gZ p(Y \,|\, U, \hphi(\z)) \,p(\z) \dif \z. \label{eq:MLE}
\intertext{The first term in this integrand,}
    p(Y \,|\, U, \hphi(\z)) \;&=\; \int_{\gX^{T}} p(Y \,|\, X, U, \hphi(\z))\, p(X | U, \hphi(\z)) \dif X \label{eq:mtds:state-integral}
\end{align}
is generally intractable for stochastic dynamics (with notable exceptions of discrete and linear-Gaussian models). In this paper we restrict our attention to deterministic state models (see above, and discussion in Appendix \ref{appdx:mtds:deterministic}), but \eqref{eq:mtds:state-integral} may be approached via existing methods such as variational methods  \citep[e.g.][]{goyal2017zforce, miladinovic2019dssm} or a \emph{Monte Carlo objective} (MCO) e.g.\ \citet{maddison2017filtering,anh2018autoencoding,naesseth2018variational}.

We then turn to the integral in equation (\ref{eq:MLE}). Generally, this integral is also unavailable in closed form, and must be approximated. We make use of a variational approach, optimizing the \emph{Evidence Lower Bound} (ELBO), $\Ls(Y, U; \bfphi, \bflambda) \,\le\, \log p(Y \, |\, U, \bfphi)$, defined by:
\begin{align}
    \Ls(Y, U; \bfphi, \bflambda) &\,=\, \E_{\qlam(\z \,|\, Y, U)} \left[ \log p(Y \,|\, U, \hphi(\z))\right] - \KLDiv{\,\qlam(\z\,|\,Y, U)\,}{\,p(\z)\,}, \label{eq:mtds:elbo}
\end{align}
where $\textrm{KL}$ is the Kullback-Leibler divergence and $\qlam(\z \,|\, Y, U)$ is an approximate posterior for $\z$. The lower bound $\sum_{i=1}^N\Ls(Y^\ii, U^\ii; \bfphi, \bflambda)$ may then be optimized via reparameterization \citep{kingma2014vae,rezende2014stochastic}, with minibatches of size $N_{\textrm{batch}} < N$. We will generally consider $\qlam(\z \,|\, Y, U) =\,\Normal{\mulam(Y, U),\,  \stdlam(Y, U)}$, where $\mulam$, $\stdlam$ are either inference networks \citep[as e.g.][]{fabius2015variational} or optimized directly as parameters. These variational parameters may of direct interest (e.g.\ for visualization), but may alternatively be an auxiliary artifact to be discarded after optimization. In all our experiments, we chose the $\stdlam(Y, U)$ to be diagonal matrices. 

We can optimize the ELBO in \eqref{eq:mtds:elbo} via access to $\nabla_\bfphi\log p(Y \,|\, U, \hphi(\z))$ which we assume to be available (typically via use of an automatic differentiation framework). For our implementation, see Appendix \ref{appdx:mtds:code}. A tighter lower bound can be achieved using related ideas, for example, using the bound introduced by the importance weighted autoencoder (IWAE) of \citet{burda2016iwae}. Where $p(Y \,|\, U, \hphi(\z))$ is a powerful base model such as an RNN, the ELBO is well-known to exhibit latent collapse: a higher value of the ELBO can be achieved if the model is able to avoid using the latent $\z$ \citep[e.g.][]{chen2016variational}. In our experiments we used a form of KL annealing \citep{bowman2016generating} to avoid this.

\subsection{Inference of $\z$} \label{sec:MTDS:inference}
A key quantity for predicting future observations is the posterior predictive distribution. For an unseen test sequence $\{Y^\prime, U^\prime\}$, the posterior predictive distribution is
\begin{align}
p(\y_{t+1:T}^\prime\,|\, \y_{1:t}^\prime,\, \bu_{1:T}^\prime) \,=\, \int_\gZ p(\y_{t+1:T}^\prime\,|\, \bu_{1:T}^\prime,\, \z)\, p(\z\,|\, \y_{1:t}^\prime,\, \bu_{1:t}^\prime) \dif \z,  \label{eq:inference-pred-post}
\end{align}
usually estimated via Monte Carlo. We must therefore have access to the posterior $p(\z\,|\, \y_{1:t}^\prime,\, \bu_{1:t}^\prime)$. This may sometimes be available as an artifact of the variational optimization, but in general we cannot assume that the variational distributions $\qlam$ generalize to $t > T$ or to out-of-sample sequences. In this section we consider a more general method of inference.

Inferring $\z$ is possible via a wide variety of variational or Monte Carlo approaches. However, given the sequential nature of the model, it is natural to consider exploiting the posterior at time $t$ for calculating the posterior at time $t{+}1$. Bayes' rule implies an update of the following form:
\begin{align}
    p(\z\,|\,\y_{1:t+1}^\prime, \bu_{1:t+1}^\prime) \;\propto\; p(\y_{t+1}^\prime \,|\,\y_{1:t}^\prime, \bu_{1:t+1}^\prime, \hphi(\z))\, p(\z\,|\,\y_{1:t}^\prime, \bu_{1:t}^\prime), \label{eq:inference-update}
\end{align}
following the conditional independence assumptions of the MTDS. This update (in principle) incorporates the information learned at time $t$ in an optimal way. We are interested in inferential methods which can exploit this prior information efficiently. Below we discuss existing work using both Monte Carlo (MC) and variational inference, before discussing our preferred approach in Sec.\ \ref{sec:MTDS:inference:our} using adaptive importance sampling.

\subsubsection{Monte Carlo Inference} 
A gold standard of inference over $\z$ may be the No U-Turn Sampler (NUTS) of \citet{hoffman2014no} (a form of Hamiltonian Monte Carlo), provided $k$ is not too large and efficiency is not a concern. However, eq.\ (\ref{eq:inference-update}) casts doubt on the use of Markov Chain Monte Carlo (MCMC) methods, since it is not obvious how to incorporate at time $t{+}1$ the samples of $\z$ obtained at time $t$. Perhaps a more relevant approach is Sequential Monte Carlo \citep[SMC, e.g.][]{doucet2000onsmc} which is designed for use in a sequential context. Unfortunately, na\"ive use of SMC (particle filtering) will result in severe particle depletion over time. To see this, let the posterior after time $t$ be approximated as $p(\z\,|\,\y_{1:t}^\prime, \bu_{1:t}^\prime) = \frac{1}{M} \sum_{m=1}^M w_m \delta(\z - \z_m)$. Then the  updated posterior at time $t+1$ will be:
\begin{align}
    p(\z\,|\,\y_{1:t+1}^\prime, \bu_{1:t+1}^\prime) &\;\propto\; \frac{1}{M} \sum_{m=1}^M w_m p(\y_{t+1}^\prime \,|\,\y_{1:t}^\prime, \bu_{t+1}^\prime, \hphi(\z)) \delta(\z - \z_m),  \nonumber \\
    \Rightarrow p(\z\,|\,\y_{1:t+1}^\prime, \bu_{1:t+1}^\prime) &\;=\; \frac{1}{M} \sum_{m=1}^M \tilde{w}_m \delta(\z - \z_m), \nonumber
\end{align}
where $\tilde{w}_m = \frac{w_m p(\y_{t+1}^\prime \,|\,\y_{1:t}^\prime, \bu_{t+1}^\prime, \hphi(\z_m))}{\sum_{j=1}^M w_j p(\y_{t+1}^\prime \,|\,\y_{1:t}^\prime,\bu_{t+1}^\prime, \hphi(\z_j))}$, simply a \emph{re-weighting} of existing particles. The number of particles with significant weights $w_m$ will reduce quickly over time \citep[see e.g.][sec.\ II.C]{doucet2000onsmc}. But since the model is static with respect to $\z$ \citep[see][]{chopin2002sequential}, there is no dynamic process to `jitter' the $\{\z_m\}$ as in a typical particle filter, and hence a resampling step cannot improve diversity.

\citet{chopin2002sequential} discusses two related solutions: firstly using `rejuvenation steps' \citep[cf.][]{gilks2001following} which applies a Markov transition kernel to each particle. The downside to this approach is the requirement to run until convergence; and the  diagnosis thereof, which may take a long time. The second alternative given is to sample from a `fixed' (or global) proposal distribution (accepting a move with the usual Metropolis-Hastings probability) for which convergence is more easily monitored. This introduces a further difficulty, however, of appropriately tuning the proposal distribution. Neither option appears practical as an efficient inner step for iterations of \eqref{eq:inference-update}. 

\subsubsection{Variational Inference} %
Variational inference (VI) considers a \emph{parametric} approximation to the posterior $p(\z\,|\,\y_{1:t},\,\bu_{1:t})$; variational approaches may not be statistically consistent, but they can generally obtain an approximation more quickly than MC methods. A well-known approach to problems with the structure of eq.\ (\ref{eq:inference-update}) is assumed density filtering \citep[ADF, see e.g.][]{opper1998bayesian}. For each $t$, ADF performs the Bayesian update and then projects the posterior into a parametric family $\gQ$. The projection is done with respect to the reverse KL Divergence, i.e.\ $q_{t+1} = \arg\min_{q \in \gQ} \KLDiv{p(\z\,|\,\y_{1:t+1}^\prime, \bu_{1:t+1}^\prime)\,}{\, q}$. Intuitively, the projection finds an `outer approximation' of the true posterior, avoiding the `mode seeking' behaviour of the usual forward KL, which is particularly problematic if it attaches to the wrong mode. 

Clearly the performance of ADF depends crucially on the choice of $\gQ$. Unfortunately, where $\gQ$ is expressive enough to capture a good approximation, the optimization problem will usually be challenging, typically requiring a stochastic gradient approach, resulting in an expensive inner loop. Furthermore, when the changes from $q_t$ to $q_{t+1}$ are relatively small, the gradient signal will be weak, which may result in misdiagnosed convergence and a consequent accumulation of error over increasing $t$. Some improvements are possible, such as re-use of stale gradients \citep{tomasetti2019updating} or standard variance reduction techniques. Nevertheless, given the possible inefficiencies of stochastic gradient approaches, compounded errors, and inaccuracies derived from $\gQ$, ADF may be considered unreliable for general use.

\subsubsection{An Adaptive Importance Sampling Approach}  \label{sec:MTDS:inference:our}
\newcommand{\qprop}{q_{\textrm{prop}}}
\newcommand{\ptarget}{p_{*}}

Having discussed an overview of possible options, we now introduce our proposed method: a sequential application of adaptive importance sampling (IS). This blends the advantages of both MC and VI methods by using a parametric posterior approximation $\qprop$ which is fitted via Monte Carlo methods, specifically, adaptive IS (see below). The parametric posterior generates the required diversity in samples without resorting to MCMC moves, and the use of IS accelerates convergence and avoids the compounded errors of ADF. The key quantity for IS is the proposal distribution $\qprop$: IS will not perform well unless $\qprop$ is well-matched to the target distribution. Our observation is that the natural annealing properties of the filtering distributions (eq.\ \ref{eq:inference-update}) allow a slow and reliable adaptation of the proposal distribution; it has proved highly effective in our applied work.


\newcommand{\AdaIS}{\scriptscriptstyle \textrm{AdaIS}}
The target distribution (the posterior) can be multimodal and highly non-Gaussian, but in practice can usually be well approximated by a Gaussian mixture model (GMM). We have found this to be a good choice for $\qprop$  in our experiments. At each time $t$, the proposal distribution $\qprop$ is improved over $N_{\AdaIS}$ iterations using adaptive importance sampling (AdaIS), described for mixture models in \citet{cappe2008adaptive}. We briefly review the methodology for a single target distribution $\ptarget$. Let the AdaIS procedure at the $n$th iteration fit the proposal:
\begin{align}
    \qprop^n(\z) \;\defeq\; \sum_{j=1}^J \alpha_j^n \Normal{\z \,\mid\,\mbf \mu_j^n,\, \Sigma_j^n},
\end{align}
with $\alpha_j^n \in \R_+$ such that $\sum_{j=1}^J \alpha_j^n = 1$ (for all $n$). For each iteration $n$, perform:
\begin{alignat}{5}
    \z_m \,&\sim\, \qprop^{n-1}, \quad &&m=1,\dotsc, M \qquad\quad && (\textrm{sample}) \nonumber & \\
    \tilde{w}_m \,&\propto\, \ptarget(\z_m) / \qprop^{n-1}(\z_m),  \quad &&m=1,\dotsc, M \qquad\quad && (\textrm{calculate weights}) \nonumber &
\end{alignat}
(where $\qprop^{0}$ is the prior). The $n$th proposal distribution $\qprop^{n}$ is then fitted to the resulting empirical distribution $\sum_{m=1}^M \tilde{w}_m \delta(\z-\z_m)$, estimating $\{\alpha_j^{n}, \mbf \mu_j^{n}, \Sigma_j^{n}\}_{j=1}^J$ via weighted Expectation Maximization \citep[see][for details]{cappe2008adaptive}. We can  monitor the effective sample size \citep[ESS, see ch.\ 9,][]{owen2013mcbook} every iteration to understand the quality of $\qprop$, and stop once the ESS has reached a certain threshold $M_\textrm{ess}$, or when $n=N_{\AdaIS}$; see Algorithm \ref{algm:mtds:inference-AMIS}.

\begin{algorithm}[t]
\SetAlgoLined
\DontPrintSemicolon
\KwResult{Approximate posteriors $\{q_t\}_{t=1}^T$}
\textbf{Inputs}: $\y_{1:T}^\prime,\, \bu_{1:T}^\prime,\, \bfphi,\, M,\, M_\textrm{ess},\, N_{\AdaIS},\, J$\;
$q_0 \leftarrow  p(\z)$\;
 \For{$t = 1:T$}{
    $\textrm{ess} \leftarrow 0$\;
    $\qprop^0 \leftarrow q_{t-1}$\;
    \For{$n = 1:\textrm{\upshape $N_{\AdaIS}$}$}{
        \For{m = 1:M}{
            $\z_m \,\sim\, \qprop^{n-1}$\;
            $w_m \,\leftarrow\, \frac{p(\y_{1:t}^\prime \,|\,\bu_{1:t}^\prime, \hphi(\z_m)) p(\z)}{\qprop^{n-1}(\z_m)}$\;
        }
        $\tilde{w}_m \,\leftarrow\, \frac{w_m}{\sum_{\ell=1}^M w_\ell},\enskip m = 1,\dotsc, M$\;
        $\qprop^n \leftarrow \textrm{WeightedEM}\left(\{\z_m\}_{m=1}^M, \{\tilde{w}_m\}_{m=1}^M, J;\,\, {\footnotesize \textrm{init}}=\qprop^{n-1} \right)$\;
        $\textrm{ess} \,\leftarrow\, \textrm{EffectiveSampleSize}\left(\{\tilde{w}_m\}_{m=1}^M\right)$\;
        \If{$\textrm{\upshape ess} > \textrm{\upshape $M_\textrm{ess}$}$}{break\;}
    }
    $q_t \leftarrow \qprop^n$
 }
 \caption{Filtered inference via Iterated AdaIS.}
 \label{algm:mtds:inference-AMIS}
\end{algorithm}

In our sequential setting, we fit a sequence of proposal distributions $q_1, q_2, \ldots, q_T$, with each proposal $q_t$ being tuned via AdaIS (using up to $N_{\AdaIS}$ adaptations), and $q_{t-1}$ forming the prior for $q_t$. (We can define $q_0 \defeq p(\z)$ for the MTDS.) The method is thus able to make use of the previous posterior without suffering from accumulated errors, and the AdaIS updates benefit from the fast initial convergence of the EM algorithm \citep[see e.g.][]{xu1996convergence}. The method is more challenging in the case where $p_*$ is intractable, but it can be achieved by integrating ideas from \citet{chopin2013smc2}, for example. Typically only a small number of AdaIS iterations are required (usually $N_{\AdaIS} \le 5$ for our problems), rendering the procedure substantially faster than MCMC moves---and by avoiding stochastic gradients---substantially faster than variational approaches. Finally, we have found the use of low-discrepancy MC samples such as Sobol sequences helpful in reducing sampling variance and further speeding convergence when drawing samples from each $\qprop^{n}$.

Application of adaptive importance sampling to sequential problems with static latent variables such as \eqref{eq:inference-update} is a novel development as far as we are aware, and it has proved superior in our experiments (in terms of speed, accuracy and robustness) to the alternative approaches discussed above. The viability of this approach is perhaps questionable for larger values of $k$ (we use $k \le 10$), but it is unclear whether larger latent spaces will be commonplace. Recall that $k$ refers to the degrees of freedom of the \emph{model family} rather than the models themselves, which can have many orders of magnitude more parameters. For relatively small deterministic LDS models, and $k=4$, inference has taken $\gO(\textrm{100ms})$ on a standard laptop per posterior $q_t$, and one may choose to thin the sequence of posteriors to multiples of $\upsilon$ steps ($t=\upsilon, 2\upsilon, 3\upsilon, \ldots$) rather than every step. An empirical study evaluating the MTDS on synthetic data using the AdaIS algorithm can be found in Chapter 4 of \citet{bird21thesis}.

\section{Related Work} \label{sec:MTDS:related}


Multi-task learning itself is a large and varied area of machine learning. The goal of MTL is to share statistical strength between models, which is achieved in classic work by sharing parameter sets \citep{caruana1998multitask}, clustering tasks \citep{bakker2003task} or learning a low-dimensional subspace prior over parameters \citep{ando2005framework}. Our approach is most similar to \citet{ando2005framework}, although we use a more flexible prior, and crucially apply the hierarchical prior to time-structured problems via dynamical systems. There is relatively little work directly extending MTL approaches to dynamical systems, but `families' of time series data are common and hence have attracted a variety of approaches in different fields including machine learning and statistics. We provide a review of some relevant work below.


\paragraph{Supervised Adaptation:} Probably the simplest form of MTL for time series involves the case where side information, $\mbf \zeta$, is available, perhaps via demographics, or other task descriptions. For RNNs, this is commonly exploited in two ways: $\mbf \zeta$ can be used to customize the $\x_0$ (`initial state customization') or appended to the inputs for each $t$ (`bias customization'), see \citet[][\S10.2.4]{goodfellow2016deep}. In the latter case we use the phrase `bias customization`, since by appending the task variable to each input $\tilde{\bu}_t \leftarrow [\bu_t\Tr, \mbf \zeta\Tr]\Tr$ for each $t$, a RNN cell takes the following form:
\begin{align}
    \x_t \,&=\, \tanh\left(A \x_{t-1} + B \tilde{\bu}_t + \mbf b\right) \,=\, \tanh\left(A \x_{t-1} + [B_1\, B_2] \begin{bmatrix}\bu_t \\ \mbf \zeta \end{bmatrix} + \mbf b\right) \nonumber \\
    &=\, \tanh\left(A \x_{t-1} + B_1 \bu_t + (\mbf b + B_2 \mbf \zeta) \right), \nonumber
\end{align}
where $B_1, B_2$ are vertical blocks of the matrix $B$. This is identical to an RNN cell with the original inputs $\{\bu_t\}$, but now with a customized cell bias $\mbf b + B_2 \mbf \zeta$. A similar argument applies to linear dynamical systems, Long Short-Term Memory networks \citep[LSTMs,][]{hochreiter1997long} and Gated Recurrent Units \citep[GRUs,][]{cho2014gru}.

For initial state customization, we may expect that the customization will be lost for large $t$, whereas bias customization ensures at least that the task information is not forgotten.  
In contrast to both approaches, our work is more expressive and moreover makes no assumption that the necessary task information is available in $\mbf \zeta$; in fact both our applications (\secref{sec:mocap}, \ref{sec:mtpd}) find a richer and more predictive task description than is suggested by known $\mbf \zeta$ metadata.  \citet{salinas2017deepar, rangapuram2018deep} go beyond bias customization by using an RNN to predict time-varying parameters of linear state space models from $\mbf \zeta$ and $\{\bu_t\}$. However, they still assume that all task information is contained in the inputs / metadata, and restrict their work to linear systems.

\paragraph{Unsupervised Bias Customization:} A number of dynamical models following an \emph{unsupervised} `bias customization' approach have been proposed recently. \citet{miladinovic2019dssm} and \citet{hsu2017disentangledseq} propose models where the biases of a Long Short-Term Memory cell (LSTM) depend on a (hierarchical) latent variable. \citet{yingzhen2018disentangled} propose a dynamical system where the latent dynamics are concatenated with a time-constant latent variable prior to the emission. In contrast, our MTDS model performs \emph{full} customization of the dynamics \emph{and} emission distributions. Note in particular that bias-customization is a highly inflexible strategy for simpler models such as the LDS.

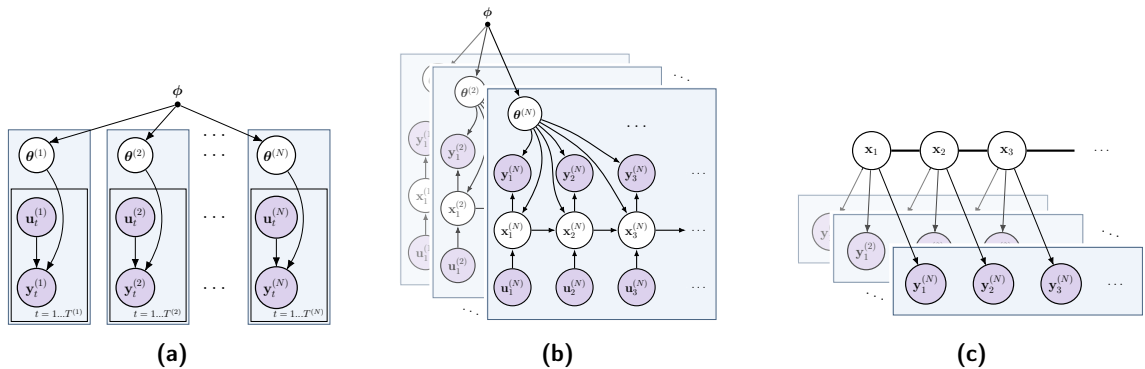
\begin{figure}
    \centering
    \begin{subfigure}[b]{0.3\textwidth}
        \centering
        \scalebox{0.75}{\scalebox{0.5}{
\begin{tikzpicture}


\specialmergetwolists{/}{1,2,N}{0,3.5, 8.5}\ziplist
\foreach \curix/\curx in \ziplist{
    \draw[color=RoyalBlue!50!Black, fill=RoyalBlue!5!white] (\curx -1.0, -1.3) rectangle ++(2.9,6.9);
    \draw[draw=black] (\curx -0.9, -1.2) rectangle ++(2.7,4.7);
    
    \node [latentctswhite]  (psi\curix) at (\curx, 4.7){\largeptfont $\bftheta^{(\curix)}$};
    \node [visiblects]  (x\curix) at (\curx, 2.5){\largeptfont $\bu_t^{(\curix)}$};
    \node [visiblects]  (y\curix) at (\curx, 0){\largeptfont $\y_t^{(\curix)}$};
    \draw[-{Latex[length=3mm]}, black] (x\curix)--(y\curix);
    
    \draw[-{Latex[length=3mm]}, black] (psi\curix) to[out=305, in=60, looseness=1] (y\curix);
    \node [invistrans] (ts\curix) at (\curx + 0.8, -0.9) {\small $t=1...T^{(\curix)}$};
 }

\node [invistrans] (dots1) at (6.3, 4.7) {\huge $\ldots$};
\node [invistrans] (dots2) at (6.3, 2.5) {\huge $\ldots$};
\node [invistrans] (dots2) at (6.3, 0) {\huge $\ldots$};

\node [blackdotmed, label=above:{\largeptfont $\bfphi$}] (phi) at (5.0,6.5){};
\draw[-{Latex[length=3mm]}, black] (phi) to (psi1.45);
\draw[-{Latex[length=3mm]}, black] (phi) to (psi2.55);
\draw[-{Latex[length=3mm]}, black] (phi) to (psiN.135);
\node [invistrans] (dots2) at (6.3, 5.4) {\huge $\ldots$};


\end{tikzpicture}
}}
        \caption{}
        \label{fig:MTDS:related:mtl-iid}
    \end{subfigure}%
    ~
    \begin{subfigure}[b]{0.35\textwidth}
        \centering
        \scalebox{0.33}{\begin{tikzpicture}

\def\numseq{3}
\def\numvars{2}  

\pgfmathtruncatemacro{\numseqm}{\numseq - 1}
\pgfmathtruncatemacro{\numvarsm}{\numvars - 1}

\newcommand{\dotsangle}{-20}
\newcommand{\dotsxshift}{.2ex}
\newcommand{\dotsyshift}{1ex}
\newcommand{\rdots}{\hspace{\dotsxshift}%
    \raisebox{\dotsyshift}{\rotatebox{\dotsangle}{$\ddots$}}}

\node [blackdotmed] (phi) at (2.5,9.2){};

\specialmergetwolists{/}{1,2,N}{0,1.3,3.5}\ziplist
\foreach \curix/\curx in \ziplist{
    \pgfmathsetmacro\cury{\curx *0.4}
    \fill[white, opacity=0.3] (-1.2, -1.5-\cury) rectangle (2.5*\numseq + 0.7 + \curx, 9.4);
    
    \draw[color=white, fill=RoyalBlue!5!white, line width=1mm] (1*\curx -1.1, -1.3-\cury) rectangle ++(2.5*\numseq + 1.8,9.4);
    \draw[color=RoyalBlue!50!Black, fill=RoyalBlue!5!white, line width=0.2mm] (1*\curx -1.0, -1.3-\cury) rectangle ++(2.5*\numseq + 1.7,9.3);
  \foreach \x in {0,...,\numseq}      
    \foreach \y in {0,...,\numvars} 
       {\pgfmathtruncatemacro{\label}{\x + 1}
       \ifnum \y = 0
            \def\varlbl{\bu}
            \def\tzndstyle{visiblects}
        \else
            \ifnum \y = 1
                \def\varlbl{\mbf x}
                \def\tzndstyle{latentctswhite}
            \else  
                \def\varlbl{\mbf y}
                \def\tzndstyle{visiblects}
            \fi
        \fi
        \def\tzndlbl{\largeptfont $\varlbl_{\label}^{(\curix)}$}
        \def\tzndxpos{2.5*\x}
        
        \ifnum \x = \numseq
            \def\tzndlbl{}
            \def\tzndstyle{invis}
            \def\tzndxpos{2.5*\x - 0.5}
        \fi
        
       \node [\tzndstyle]  (\y\x\curix) at (\tzndxpos + 1*\curx, 2.3*\y -\cury) {\tzndlbl}
       ;} 
    
      \foreach \y  in {0,...,\numvars}
        \node[anchor=center] at (2.5*\numseq + 1*\curx, 2.3*\y-\cury) {\largeptfont $\ldots$};
        
  \foreach \y in {1,...,1}  
    \foreach \x in {0,...,\numseqm}
      \pgfmathtruncatemacro{\xp}{\x + 1}%
      \draw[stdarrow] (\y\x\curix)--(\y\xp\curix)
      ;
      
  \foreach \y in {1,...,\numvars}  
    \foreach \x in {0,...,\numseqm}
      \pgfmathtruncatemacro{\ym}{\y - 1}%
      \draw[stdarrow] (\ym\x\curix)--(\y\x\curix)
      ;
    
    \node [latentctswhite]  (psi\curix) at (0.5 + 1*\curx, 7.0- \cury){\largeptfont $\bftheta^{(\curix)}$};
    \foreach \y in {1,...,\numvars}
      \foreach \x in {0,...,2}
        \ifthenelse{\x>0}{\def\looseamt{1.2}}{\def\looseamt{1.0}}
        \ifthenelse{\x>0}{\def\inangle{120}}{\def\inangle{60}}
        \ifthenelse{\equal{\x}{2} \AND \equal{\y}{2}}{\def\inangle{148}}{}
        \ifthenelse{\equal{\x}{0}}{\def\outangle{290+\x*20-\y*10+20}}{\def\outangle{290+\x*20+\ym*10}}
        \pgfmathtruncatemacro{\ym}{\y - 1}%
        \draw[stdarrow] (psi\curix) to[out=\outangle, in=\inangle, looseness=\looseamt] (\y\x\curix);
        
    \node [invistrans] (dotspsi) at (5.0+\curx, 6.5-\cury) {\huge $\ldots$};
    
    \pgfmathtruncatemacro{\hierangle}{45 + 18*\curx};
    \draw[thickerarrow] (phi) to (psi\curix.\hierangle);
 }

\node [blackdotmed, label=above:{\largeptfont $\bfphi$}] (phisolid) at (2.5,9.2){};
    
\node [invistrans, rotate=20] (dotsd1) at (1.8, -2.3) {\huge $\rdots$};
\node [invistrans, rotate=20] (dotsd2) at (2.8+\numseq*2.5, 7.13) {\huge $\rdots$};

\end{tikzpicture}}
        \caption{}
        \label{fig:MTDS:related:mtds}
    \end{subfigure}%
    ~
    \begin{subfigure}[b]{0.35\textwidth}
        \centering
        \scalebox{0.36}{\begin{tikzpicture}

\def\numseq{3}
\def\numvars{2}  

\pgfmathtruncatemacro{\numseqm}{\numseq - 1}
\pgfmathtruncatemacro{\numvarsm}{\numvars - 1}

\newcommand{\dotsangle}{-20}
\newcommand{\dotsxshift}{.2ex}
\newcommand{\dotsyshift}{1ex}
\newcommand{\rdots}{\hspace{\dotsxshift}%
    \raisebox{\dotsyshift}{\rotatebox{\dotsangle}{$\ddots$}}}

\newcommand{\gpchainxoffset}{1.5}
\foreach \x in {0,...,\numseqm}
   {
    \pgfmathtruncatemacro{\label}{\x + 1}
    \node [latentcts, minimum size=40pt]  (L\x) at (2.5*\x + \gpchainxoffset, 3.0) {\largeptfont $\mathbf{x}_{\label}$};
   } 
\node [invis]  (L\numseq) at (2.5*\numseq + \gpchainxoffset, 3.0) {};
   
\foreach \x in {0,...,\numseqm}
  \pgfmathtruncatemacro{\xp}{\x + 1}%
  \draw[line width=0.8mm, black] (L\x)--(L\xp);

\node[anchor=center] at (2.5*\numseq + \gpchainxoffset + 1.0 , 3.0) {\largeptfont $\ldots$};
      
\specialmergetwolists{/}{1,2,N}{0,1.3,3.5}\ziplist
\foreach \curix/\curx in \ziplist{
    \pgfmathsetmacro\cury{\curx *0.55}
    \fill[white, opacity=0.3] (-1.4, -1.5-\cury) rectangle (2.5*\numseq + 0.7 + \curx, 2.28);
    
    \draw[color=white, fill=RoyalBlue!5!white, line width=1mm] (1*\curx -1.3, -1.15-\cury) rectangle ++(2.5*\numseq + 1.8,2.6);
    \draw[color=RoyalBlue!50!Black, fill=RoyalBlue!5!white, line width=0.2mm] (1*\curx -1.2, -1.15-\cury) rectangle ++(2.5*\numseq + 1.7,2.5);
  \foreach \x in {0,...,\numseq}
       {\pgfmathtruncatemacro{\label}{\x + 1}
        \def\tzndlbl{\largeptfont $\y_{\label}^{(\curix)}$}
        \def\tzndxpos{2.5*\x}
        
        \def\tzndstyle{visiblects}
        \ifnum \x = \numseq
            \def\tzndlbl{}
            \def\tzndstyle{invis}
            \def\tzndxpos{2.5*\x - 0.5}
        \fi
        
       \node [\tzndstyle]  (0\x\curix) at (\tzndxpos + 1*\curx, -\cury) {\tzndlbl}
       ;} 
    
      \node[anchor=center] at (2.5*\numseq + 1*\curx -0.5, -\cury) {\largeptfont $\ldots$};
        
      
    \foreach \x in {0,...,\numseqm}
      \draw[-{Latex[length=2.5mm]}, black] (L\x)--(0\x\curix);
 }

 
\node [invistrans, rotate=20] (dotsd1) at (1.6, -2.3) {\huge $\rdots$};
\node [invistrans, rotate=20] (dotsd2) at (2.8+\numseq*2.5, 0) {\huge $\rdots$};

\end{tikzpicture}}
        \caption{}
        \label{fig:multi-task-gp}
    \end{subfigure}%
    \caption{Comparison of MTL approaches: \captiona\ MT iid model;  \captionb\ MT dynamical system; \captionc\ MT Gaussian process. A thick horizontal bar represents a set of fully connected nodes.}
    \label{fig:MTDS:related}
\end{figure}

\paragraph{Adjusting Transition Weights:} In the context of autoregressive or `gated' Boltzmann machines, \citet{memisevic2007unsupervised} propose dynamics based on the sum of transition matrices, weighted by a (binary-valued vector) latent variable. This idea is extended in \citet{memisevic2010learning} to be a sum of \emph{low rank} transition matrices, which substantially lowers the parameter count. \citet{spieckermann2015exploiting} use a similar idea to adjust the transition matrix of an RNN, although the latent variable $\z$ is optimized as a parameter. The MTDS goes beyond these models by modulating \emph{all} weights of \emph{general} dynamical systems via nonlinear manifolds in parameter space via a probabilistic objective.

\paragraph{Clustering via Time Series Dynamics:} A related idea to the MTDS is clustering a collection of time series via their dynamics. This is useful in a variety of circumstances including scientific discovery, exploratory analysis and model building. Approaches include using different projections of the same multivariate latent process \citep[e.g.][]{inoue2007cluster, chiappa2007output}, or independent dynamical systems which merely share parameters, similar to the MTDS \citep[e.g.][]{lin2019clustering}. The MTDS uses a relaxation of the cluster assumption to a real-valued latent $\z$. 

\paragraph{Multi-task GPs (MTGPs):} Multi-task GPs \citep{bonilla2008multi} are commonly used for sequence prediction. Examples include those in \citet{osborne2008towards, titsias2011spike, alvarez2012kernels, roberts2013gpts}. MTGPs however are only \emph{linear} combinations of latent functions (see Figure \ref{fig:MTDS:related}(c)). Since the sharing of information is mediated via  the same latent process, an MTGP can only provide \emph{different projections of the same underlying phenomena}. In contrast, an MTDS can also gain strength over dynamic properties such as differing oscillation frequencies and damping factors, modulations of which cannot be expressed as finite linear combinations of a set of training task observations.

Configurations beyond that shown in Figure 2(c) are possible. For example Linear latent Force Models (LLFMs, \citealt{alvarez2013linear}) have inputs which are best thought of in our notation as $\x$ variables, although they are referred to as $\bu$'s in the LLFM paper. ODE structure on the $\y$'s in LLFMs gives rise to a process convolution of the input processes $\x$ with respect to the Green's function associated with each ODE. The shared set of input processes $\x_q(t),\; q = 1, \ldots, Q$ gives rise to correlated outputs amongst the $\y$'s. Thus, despite the process convolutions, the structure on the $\y$'s is similar to the work on multi-task GPs cited above, in that they linearly combine the $\x$'s. In contrast, an MTDS can allow entirely different realizations of the latent process, as well as make use of inputs $\{ \bu_t \}$.

\paragraph{Application to Video Data:} Models for video data have considered the related problem of scene decomposition. Methods include \citet{denton2017video, villegas2017decomposing, tulyakov2018mocogan} which decompose time-varying and static features of a video using adversarial ideas or architectural constraints. \citet{hsieh2018learning} extends these approaches by learning a parts-based scene decomposition, and applying such a latent decomposition to each part. These methods have a broad similarity to the unsupervised bias-customization approaches; in particular, the  dynamic evolution cannot be easily customized.


\paragraph{Mixed Effects and Hierarchical Models:} Mixed effects (ME) models are a statistical approach to account for inter-individual differences in data, and have been applied to time series data. In the Bayesian context, these models can be interpreted as hierarchical models, which bear some similarity to the MTDS. However, the philosophy is very different: ME models usually consider the random parameters as `nuisance variation' to be \emph{integrated out}; the MTDS uses low-dimensional assumptions to gain strength for learning \emph{bespoke parameters}. ME models for time series models tend to be relatively simple \citep[e.g.][]{tsimikas1997mixed}, or include bespoke and complex learning algorithms \citep[e.g.][]{zhou2013nonlinear} which do not generalize easily. In contrast, our MTDS methods are more widely applicable, and can be used to facilitate improved predictions as more data are observed.

\section{Application to Mocap Data} \label{sec:mocap}

\begin{figure}
    \centering
    \includegraphics[width=0.99\textwidth]{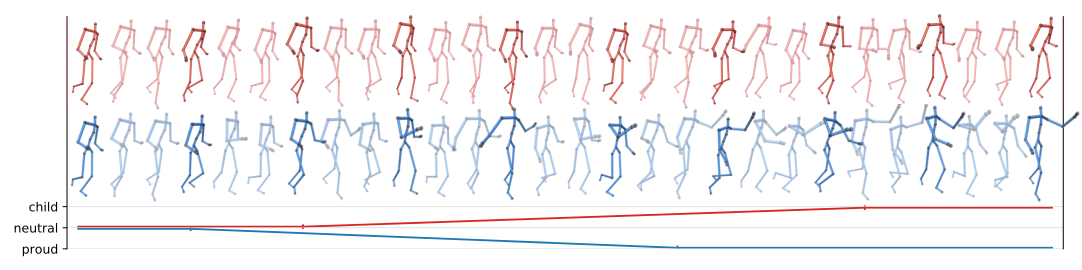}
    \caption{Morphing the style of a human locomotion sequence over time: (red) from neutral to childlike; (blue) from neutral to proud; (bottom) interpolation schedule.}
    \label{fig:mocap:intro:morphing}
\end{figure}

We now turn to real-world applications of the multi-task dynamical system. This section details the first of two such applications: modelling human locomotion. This is a clear example of a sequence family: humans can walk in a variety of different styles depending on their current activity or mood, but for each person there is a unique character present in all these styles. More generally, despite inter-individual differences \citep[see e.g.][]{lee2002gait} there is an intrinsic similarity to all human locomotion. We might therefore expect the majority of such differences to be captured with relatively few degrees of freedom. 

To this end, we train an MTDS on motion capture (mocap) data,  which is introduced in \secref{sec:mocap:data}. The MTDS can help to address several outstanding problems in this context: training dynamical models on limited data, robustness to dataset shift, and style transfer for generative models. We give our experimental results for each of these categories in \secref{sec:mocap:results}, with the experimental setup given in  \secref{sec:mocap:expmt-setup}. A result of particular interest is that the MTDS allows interpolation or morphing of style over time (see Figure \ref{fig:mocap:intro:morphing}), which, to the best of our knowledge, is an entirely novel contribution.

\subsection{Data and Pre-Processing} \label{sec:mocap:data}
This section provides an introduction to the mocap data that was used (\secref{sec:mocap:data:data}), together with the representation constructed for the observations (\secref{sec:mocap:data:rep}).

\subsubsection{Mocap Data} \label{sec:mocap:data:data}

\begin{figure}
    \centering
    \begin{subfigure}[b]{0.5\textwidth}
        \centering
        \includegraphics[width=0.9\textwidth]{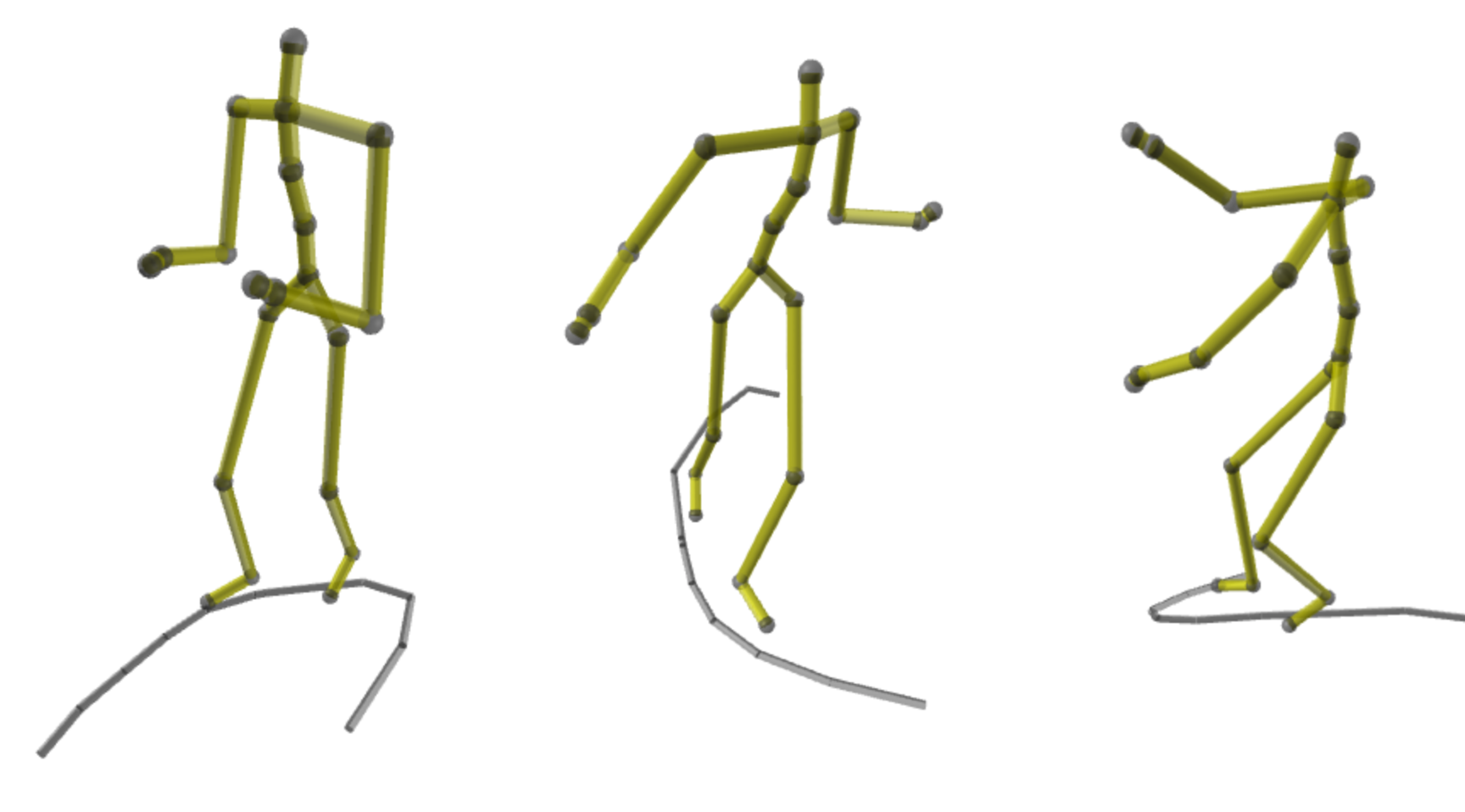}
        \caption{}
        \label{fig:mocap:intro:example-datapoints:eg}
    \end{subfigure} %
    \qquad\quad %
    \begin{subfigure}[b]{0.3\textwidth}
        \includegraphics[width=0.9\textwidth]{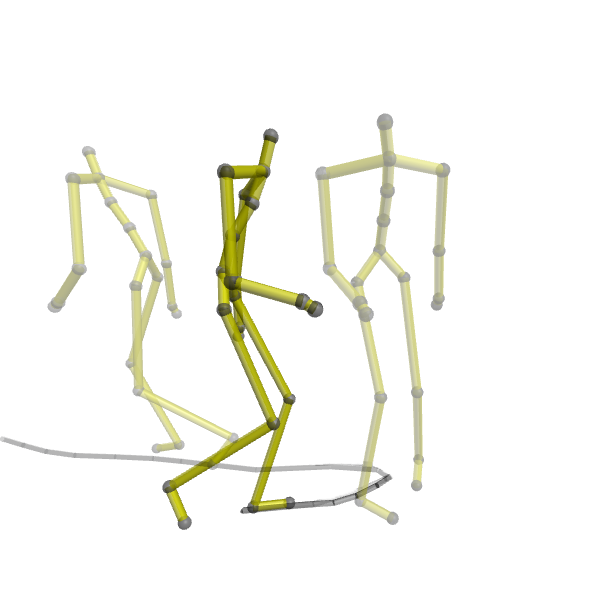}
        \caption{}
        \label{fig:mocap:intro:reverse}
    \end{subfigure} 
    \caption{(a) Example frames from mocap data set, together with $\pm 1$ second ground trajectory shown in grey. (b) Example subsequence where the skeleton rotates in the opposite direction to the corner.}
    \label{fig:mocap:intro:example-datapoints}
\end{figure}

The data are obtained from \citet{mason2018few} which consists of planar walking and running motions in 8 styles recorded with a motion capture (mocap) suit by a single actor. These styles are: `angry', `childlike', `depressed', `neutral', `old', `proud', `sexy', `strutting'. The path (or `trajectory') along which the skeleton is walking is provided as an input, in order to avoid modelling the random/unpredictable choices of the actor. See Figure \ref{fig:mocap:intro:example-datapoints} for examples. In this case the sequence family corresponds to the possible walking styles humans exhibit while following a pre-determined path.

The original data is recorded at 120 fps; we downsample to 30 fps as per \citet{martinez2017human, pavllo2018quaternet}. Each style has 3 to 5 individual sequences of varying length, totalling ca.\ $2000$ frames per style. The data are mapped to a 21-joint skeleton: this is a subset of the CMU skeleton \citep{cmu2009guide} used by \citet{holden2016deep}, see Figure \ref{fig:mocap:representation:skeleton}. Unlike \citet{mason2018few}, we do not perform any data augmentation such as mirroring. Our interests are primarily in the contribution of the MTDS in modelling style, rather than in producing high fidelity computer graphics. For inputs to the model, we provide the ground trajectory, the gait cycle (a value in $[0, 2\pi)$ corresponding to the phase of the leg motion), and a Boolean indicator for the rotational direction while traversing a corner (see Appendix \ref{sec:mocap:data:inputs} for further details). 

\subsubsection{Data Representation} \label{sec:mocap:data:rep}

\begin{figure}
    \centering
    \begin{subfigure}[b]{0.16\textwidth}
        \includegraphics[trim={170 40 160 50}, clip, width=0.9\textwidth]{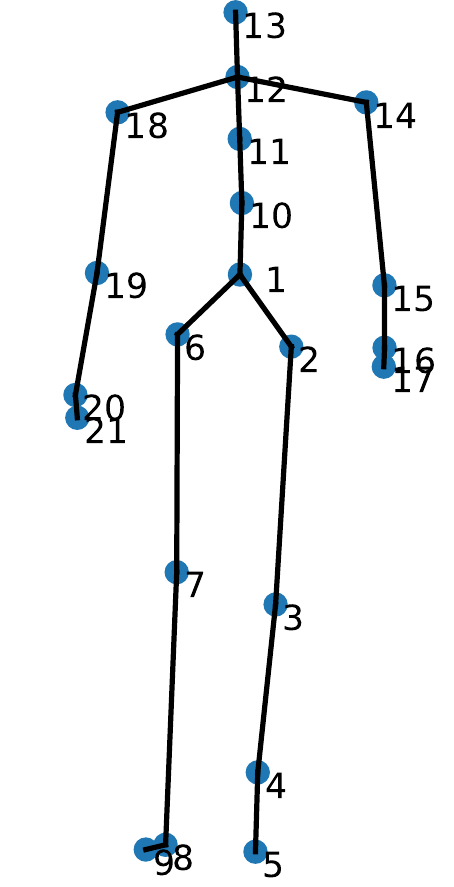}
        \caption{}
        \label{fig:mocap:representation:skeleton}
    \end{subfigure} 
    \begin{subfigure}[b]{0.24\textwidth}
        \includegraphics[trim={0 0 0 40}, clip, width=0.9\textwidth]{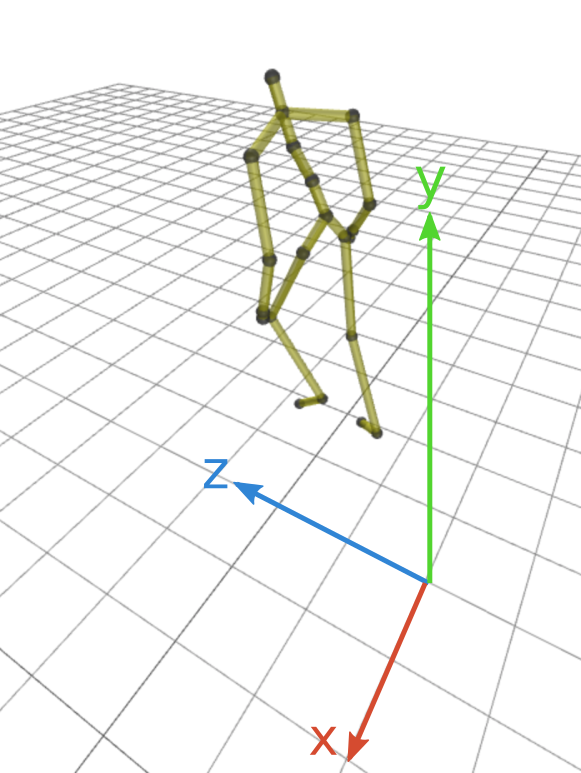}
        \caption{}
        \label{fig:mocap:representation:euler}
    \end{subfigure} 
    \begin{subfigure}[b]{0.24\textwidth}
        \includegraphics[trim={0 0 0 40}, clip, width=0.9\textwidth]{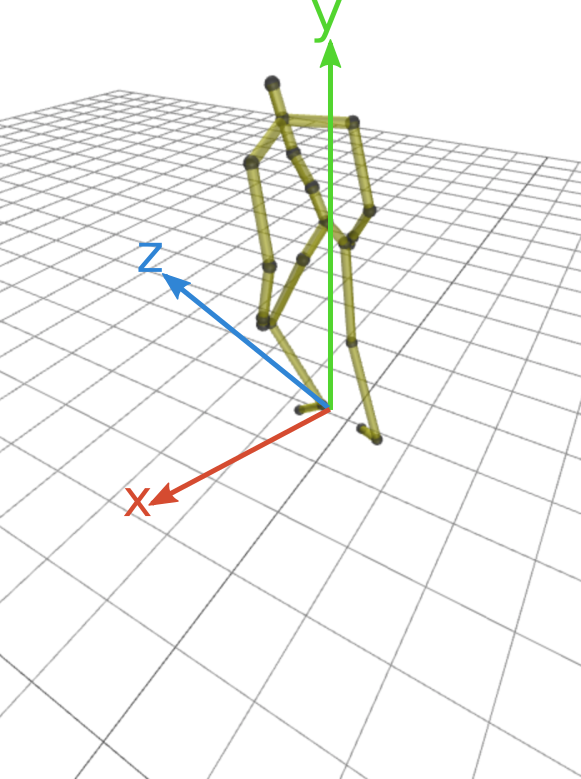}
        \caption{}
        \label{fig:mocap:representation:lagrange}
    \end{subfigure} 
    \caption{\captiona\ The 21-joint skeleton. \captionb\ Eulerian representation. \captionc\ Lagrangian representation.}
    \label{fig:mocap:representation}
\end{figure}

We choose a Lagrangian representation (Figure \ref{fig:mocap:representation:lagrange}) where the coordinate frame is centered at the \emph{root joint} of the skeleton (joint 1 in Fig.\ \ref{fig:mocap:representation:skeleton}, the pelvis), projected onto the ground. The frame is rotated such that the $z$-axis points in the `forward' direction, approximately normal to the plane described by the shoulder blades and pelvis. This is in contrast to the Eulerian frame, which has a fixed position for all $t$ (Figure \ref{fig:mocap:representation:euler}). In the Lagrangian frame, the joint positions are always relative to the root joint, which avoids confusing their motion with the trajectory and rotation of the entire skeleton. The relative joint positions are encoded by position rather than by the angle made with their parent joint. This may result in the model violating bone length constraints, but reduces the sensitivity of internal joints, as discussed for example in \S2.1, \citet{pavllo2018quaternet}. 

Finally, we choose to encode the root joint via its velocity (i.e.\ via differencing), which allows an animator to more easily amend the trajectory of the model output. We do not represent joints 2 to 21 (relative joint motion) in this way to avoid accumulated errors over time. Our per-frame representation hence consists of the `velocity' $\dot{x}, \dot{z}, \dot{\omega}$ of the co-ordinate frame, the vertical position of the root joint, and 3-d position of the remaining 20 joints, resulting in $\y_t \in R^{64}$. The $\{\y_t\}$ are standardized to have zero mean and unit variance per observation channel. 

\subsection{MTDS Model} \label{sec:mocap:expmt-setup:models:mtds}

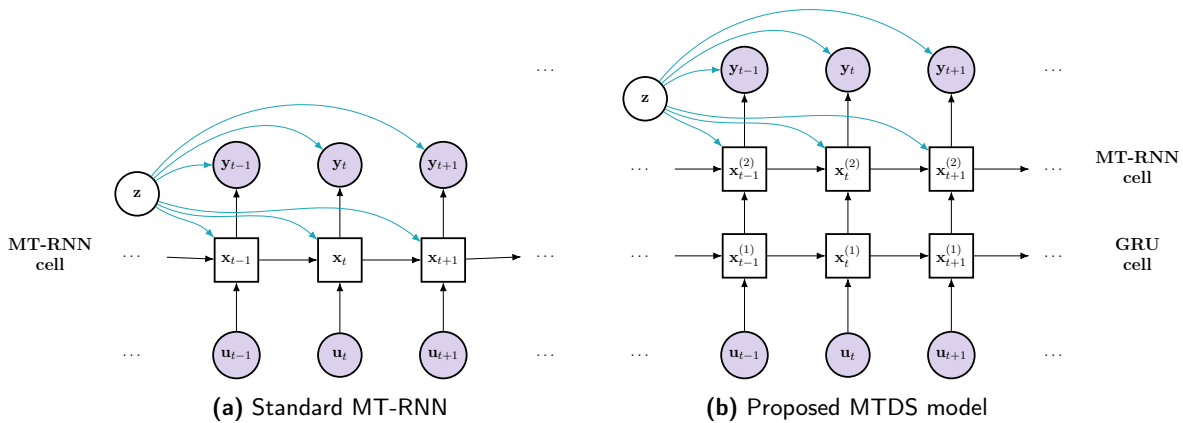
\begin{figure}
\begin{adjustwidth}{-0.5cm}{-0.5cm}
\centering
\begin{subfigure}[t]{0.49\linewidth}
    \scalebox{0.55}{
        \begin{tikzpicture}

\def\numseq{2}   
\def\xlabels{{"t-1","t","t+1"}}

\def\vsepnodes{2.3}  
\def\hsepnodes{2.5}
\def\vrnnadj{0.1}

\foreach \x in {0,...,\numseq}
{
    \def\tzxpos{\hsepnodes*\x}
    \pgfmathsetmacro{\label}{\xlabels[\x]}
    \node[visiblects](y\label) at (\tzxpos,\vsepnodes) {$\mbf y_{\label}$};
    \node[latentdiscrete](x1\label) at (\tzxpos,0) {$\mbf x_{\label}$};
    \node[visiblects](u\label) at (\tzxpos,-\vsepnodes) {$\mbf u_{\label}$};
}

\pgfmathtruncatemacro{\numseqm}{\numseq - 1}
\pgfmathtruncatemacro{\numseqp}{\numseq + 1}
\pgfmathtruncatemacro{\numseqpp}{\numseq + 2}
\pgfmathtruncatemacro{\sinklabelm}{\numseq + 1}
\pgfmathtruncatemacro{\sinklabel}{\numseq + 2}
\def\tzsinkpos{\hsepnodes*\sinklabelm}
\node[invis](u\sinklabel) at (\tzsinkpos, -\vsepnodes) {\ldots};
\node[invis](x1\sinklabel) at (\tzsinkpos, \vrnnadj) {\ldots};
\node[invis](y\sinklabel) at (\tzsinkpos, 2*\vsepnodes) {\ldots};
\def\sourcelabel{source}
\node[invis](u\sourcelabel) at (-\hsepnodes, -\vsepnodes) {\ldots};
\node[invis](x1\sourcelabel) at (-\hsepnodes, +\vrnnadj) {\ldots};

\node[invis](x1label) at (-2*\hsepnodes+0.5, +\vrnnadj) {\bfseries\large\begin{tabular}{c} MT-RNN \\ cell\end{tabular}};

\node[latentcts](z) at (-\hsepnodes+0.1, \vsepnodes-0.7) {$\z$};

\foreach \x in {0,...,\numseq}
{
    \pgfmathsetmacro{\label}{\xlabels[\x]}
    \draw[blackarrow] (u\label) -- (x1\label);
    \draw[blackarrow] (x1\label) -- (y\label);
}
\foreach \x in {0,...,\numseqm}
{
    \pgfmathsetmacro{\label}{\xlabels[\x]}
    \pgfmathsetmacro{\labelnext}{\xlabels[\x+1]}
    \draw[blackarrow] (x1\label) -- (x1\labelnext);
}
\foreach \x in {0,...,\numseq}
{
    \pgfmathsetmacro{\label}{\xlabels[\x]}
    \pgfmathtruncatemacro{\outangle}{-35+8*\x}
    \pgfmathtruncatemacro{\inangle}{135+3*\x}
    \draw[bluearrow] (z) to[out=\outangle,in=\inangle] (x1\label);
    \pgfmathtruncatemacro{\outangle}{35+8*\x}
    \ifthenelse{\equal{\x}{0}}{
        \pgfmathtruncatemacro{\inangle}{180}
        }{
        \pgfmathtruncatemacro{\inangle}{135+3*\x}
    }
    \draw[bluearrow] (z) to[out=\outangle,in=\inangle] (y\label);
}

\pgfmathsetmacro{\firstlabel}{\xlabels[0]}
\draw[blackarrow] ( $ (x1\sourcelabel)!.33!(x1\firstlabel) $ ) -- (x1\firstlabel);

\pgfmathsetmacro{\lastlabel}{\xlabels[\numseq]}
\draw[blackarrow] (x1\lastlabel) -- ( $ (x1\lastlabel)!.75!(x1\numseqpp) $ );

\end{tikzpicture}
    }
    \subcaption*{\protect\phantom{XXXX} \textbf{(a)}  Standard MT-RNN}
\end{subfigure}%
\hfill%
\begin{subfigure}[t]{0.49\linewidth}
    \scalebox{0.55}{
        \begin{tikzpicture}

\def\numseq{2}   
\def\xlabels{{"t-1","t","t+1"}}

\def\vsepnodes{2.3}  
\def\hsepnodes{2.5}
\def\vrnnadj{0.1}

\foreach \x in {0,...,\numseq}
{
    \def\tzxpos{\hsepnodes*\x}
    \pgfmathsetmacro{\label}{\xlabels[\x]}
    \node[visiblects](y\label) at (\tzxpos,2*\vsepnodes) {$\mbf y_{\label}$};
    \node[latentdiscrete](x2\label) at (\tzxpos,\vsepnodes-\vrnnadj) {$\mbf x_{\label}^{(2)}$};
    \node[latentdiscrete](x1\label) at (\tzxpos,+\vrnnadj) {$\mbf x_{\label}^{(1)}$};
    \node[visiblects](u\label) at (\tzxpos,-\vsepnodes) {$\mbf u_{\label}$};
}

\pgfmathtruncatemacro{\numseqm}{\numseq - 1}
\pgfmathtruncatemacro{\numseqp}{\numseq + 1}
\pgfmathtruncatemacro{\numseqpp}{\numseq + 2}
\pgfmathtruncatemacro{\sinklabelm}{\numseq + 1}
\pgfmathtruncatemacro{\sinklabel}{\numseq + 2}
\def\tzsinkpos{\hsepnodes*\sinklabelm}
\node[invis](u\sinklabel) at (\tzsinkpos, -\vsepnodes) {\ldots};
\node[invis](x1\sinklabel) at (\tzsinkpos, \vrnnadj) {\ldots};
\node[invis](x2\sinklabel) at (\tzsinkpos, \vsepnodes-\vrnnadj) {\ldots};
\node[invis](y\sinklabel) at (\tzsinkpos, 2*\vsepnodes) {\ldots};
\def\sourcelabel{source}
\node[invis](u\sourcelabel) at (-\hsepnodes, -\vsepnodes) {\ldots};
\node[invis](x1\sourcelabel) at (-\hsepnodes, +\vrnnadj) {\ldots};
\node[invis](x2\sourcelabel) at (-\hsepnodes, \vsepnodes-\vrnnadj) {\ldots};

\node[invis](x1label) at (\hsepnodes*\sinklabel-0.5, +\vrnnadj) {\bfseries\large\begin{tabular}{c} GRU \\ cell\end{tabular}};
\node[invis](x2label) at (\hsepnodes*\sinklabel-0.5, \vsepnodes-\vrnnadj) {\bfseries\large\begin{tabular}{c} MT-RNN \\ cell\end{tabular}};

\node[latentcts](z) at (-\hsepnodes+0.1, 2*\vsepnodes-0.7) {$\z$};

\foreach \x in {0,...,\numseq}
{
    \pgfmathsetmacro{\label}{\xlabels[\x]}
    \draw[blackarrow] (u\label) -- (x1\label);
    \draw[blackarrow] (x1\label) -- (x2\label);
    \draw[blackarrow] (x2\label) -- (y\label);
}
\foreach \x in {0,...,\numseqm}
{
    \pgfmathsetmacro{\label}{\xlabels[\x]}
    \pgfmathsetmacro{\labelnext}{\xlabels[\x+1]}
    \draw[blackarrow] (x1\label) -- (x1\labelnext);
    \draw[blackarrow] (x2\label) -- (x2\labelnext);
}
\foreach \x in {0,...,\numseq}
{
    \pgfmathsetmacro{\label}{\xlabels[\x]}
    \pgfmathtruncatemacro{\outangle}{-35+8*\x}
    \pgfmathtruncatemacro{\inangle}{135+3*\x}
    \draw[bluearrow] (z) to[out=\outangle,in=\inangle] (x2\label);
    \pgfmathtruncatemacro{\outangle}{35+8*\x}
    \ifthenelse{\equal{\x}{0}}{
        \pgfmathtruncatemacro{\inangle}{180}
        }{
        \pgfmathtruncatemacro{\inangle}{135+3*\x}
    }
    \draw[bluearrow] (z) to[out=\outangle,in=\inangle] (y\label);
}

\pgfmathsetmacro{\firstlabel}{\xlabels[0]}
\draw[blackarrow] ( $ (x1\sourcelabel)!.33!(x1\firstlabel) $ ) -- (x1\firstlabel);
\draw[blackarrow] ( $ (x2\sourcelabel)!.33!(x2\firstlabel) $ ) -- (x2\firstlabel);

\pgfmathsetmacro{\lastlabel}{\xlabels[\numseq]}
\draw[blackarrow] (x1\lastlabel) -- ( $ (x1\lastlabel)!.75!(x1\numseqpp) $ );
\draw[blackarrow] (x2\lastlabel) -- ( $ (x2\lastlabel)!.75!(x2\numseqpp) $ );

\end{tikzpicture}
    }
    \subcaption*{ \textbf{(b)} Proposed MTDS model \protect\phantom{XXXXXXXX}}
\end{subfigure}
\end{adjustwidth}
\caption{Graphical model of a standard MT-RNN and the proposed MTDS, for a single style $i$. For clarity, the superscript $i$ is omitted, and the dependence on $\z$ is shown in a different colour.}
\label{fig:mocap:mtrnn-gm}
\end{figure}

A natural choice of base model for the MTDS is a RNN (or gated variant such as LSTM or GRU) due to the widespread use of such models in the mocap literature. A graphical model of a multi-task RNN (MT-RNN) is shown in Figure \ref{fig:mocap:mtrnn-gm}\captiona. Preliminary experiments using MT-RNNs showed strong performance on the training data, but on changing the value of $\z$, predictions either did not change, or became implausible. This appears to be due to some form of information leakage about style (despite substantial efforts to the contrary---see Appendix \secref{sec:mocap:data:inputs}). The RNN was hence performing some of the style inference instead of relying on the latent $\z$. Furthermore, since certain features such as trajectory corner types, or extreme values of speed, or gait frequency are only found within a single style (or group of styles), the response to such unique inputs was only learned for certain values of $\z$.

In order to capture the global input-output relationships across all styles, we propose a 2-layer RNN structure, where the first layer is \emph{not} multi-task and is able to learn a shared representation across all tasks. The second layer is a MT-RNN which performs  specialization to the style using the latent variable $\z$. The separation of responsibility between layers is assisted by three important refinements. The first layer passes information to the second layer via a bottleneck of $\ell$ units, encouraging the first layer to focus on the globally most important features. Secondly, we omit a skip connection from the first layer, which would undermine the bottleneck. Finally, the MT-RNN has a relatively higher learning rate, resulting in a preference to model more variation than the first layer. Since its only inputs are the low-dimensional shared representation of the first layer, $\z$ is forced to explain more of the variance.

In detail, our proposed MTDS model is a 2 hidden-layer recurrent model where the first hidden layer is a 1024 unit GRU and the second hidden layer is a 128 unit MT-RNN,  followed by a linear decoding layer. Explicitly, omitting index $i$, the model for a style represented by the task variable $\z$ is:
\newcommand{\smallminus}{\raisebox{-0.15ex}{\scalebox{0.5}[1.0]{\( - \)}}}
\begin{align}
    [\bfpsi_2(\z), C(\z), \mbf d(\z)] &= \hphi(\z), \label{eq:mocap-theta}\\
    \x_{1,t} &= \textrm{\small GRUCell}_{1024}\left(\x_{1, t\smallminus 1},\, \bu_t;\,\,\, \bfpsi_1\right), \label{eq:mocap-gru}\\
    \x_{2,t} &= \textrm{\small RNNCell}_{128}\left(\x_{2, t\smallminus 1},\,H \x_{1, t\smallminus 1};\, \bfpsi_2(\z)\right), \label{eq:mocap-mtrnn}\\
    \hat{\y}_t &= C(\z) \x_{2,t} + \mbf d(\z), \label{eq:mocap-output}
\end{align}
for $t = 1, \ldots, T$; $\textrm{\small GRUCell}$, $\textrm{\small RNNCell}$ are defined in Appendix \ref{appdx:mocap:recurrentcells}. See Figure \ref{fig:mocap:mtrnn-gm}b for a graphical model. The parameters which depend on $\z$ are $\bftheta = \{\bfpsi_2, C, \mbf d\} \in \R^d$ and the learnable parameters are thus $\{\bfpsi_1, H, \bfphi\}$.\footnote{The parameters which depend on $\z$ have the following dimensions; $\bfpsi_2$: $128\times128$; $C_2$: $128\times64$; and $\mbf d$: 64. This results in a total of $d=24,640$ dimensions for the base parameters of the MT-RNN.} In our experiments, we use $\ell=24$ units for the bottleneck matrix $H$.\footnote{The choice of $\ell$ was not highly optimized---in fact this was the first value we tried, and it appeared to be adequate.} The first layer GRU uses 1024 units primarily for qualitative reasons, since it was observed to produce smoother animations than smaller networks. Using a MT-GRU instead of a MT-RNN in the second layer produced worse style transfer; changing the value of $\z$ in this case often made little difference to the style. We have observed in other work that GRUs can more easily perform style inference via use of their gates \citep[see][\S4.5]{bird21thesis}. 

The choice of $\hphi$ in this instance appears to be relatively unimportant. In practice we used an affine function (i.e.\ $\hphi(\z) = W \z + \mbf b$ for $W \in \R^{d \times k}$ and $\mbf b \in \R^{d}$). Preliminary experiments suggested that use of a multi-layer perceptron (MLP) for $\hphi$ resulted in similar performance, but initialization was more difficult, and inference of $\z$ was slower.

\begin{figure}
    \centering
    \includegraphics[height=200px]{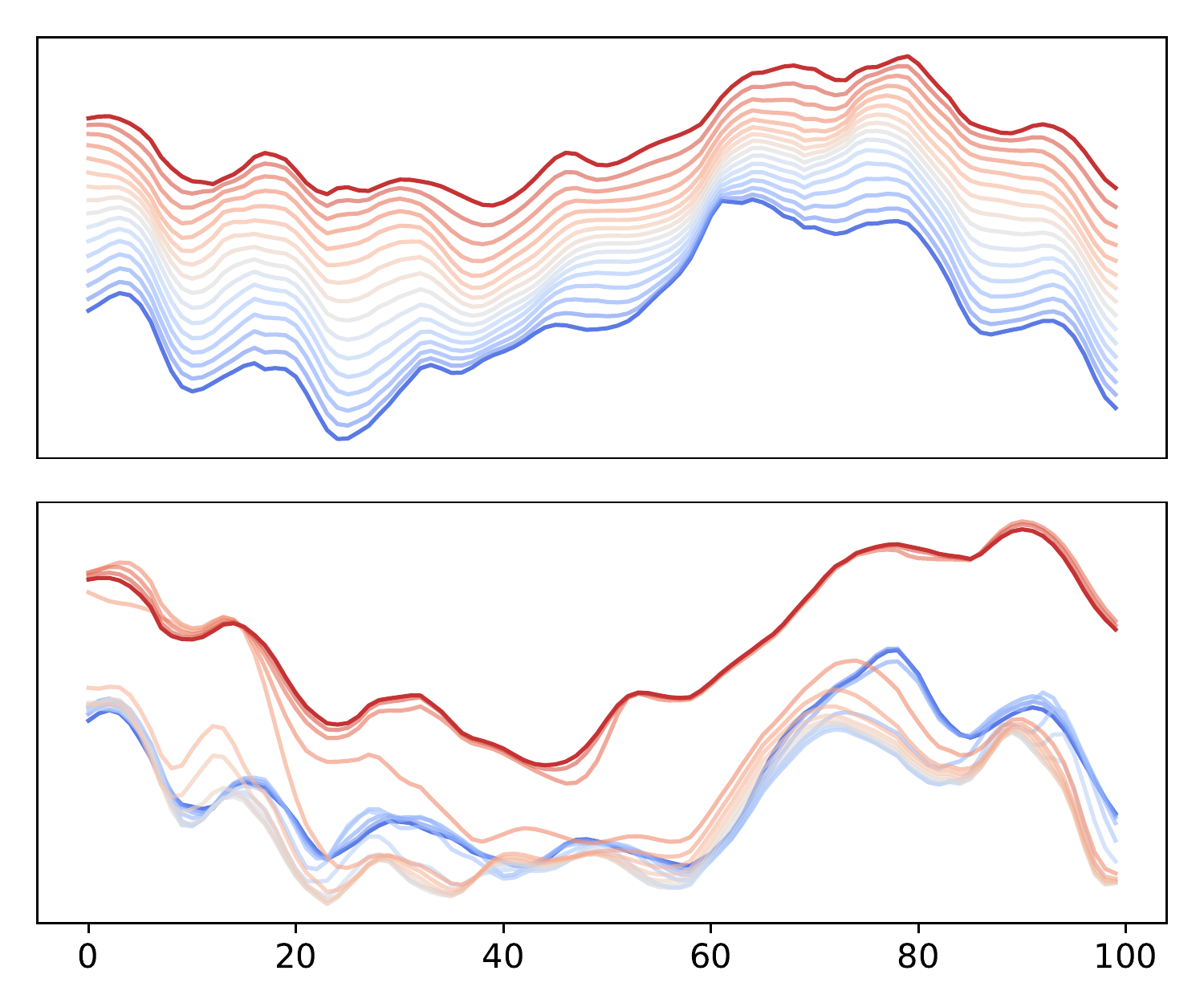}
    \caption{Sequence interpolation for model of joint 11 ($x$ direction) over time $t=1,\ldots,100$ with latent $\z$ varied from `strutting' (blue) to `angry' (red) style. \emph{(top)} MTDS model; \emph{(bottom)} MT-Bias model.}
    \label{fig:mocap:interp_MTRNN_vs_MTBias}
\end{figure}%

In order to ensure smooth variation of the dynamics wrt.\ $\z$, it proved important to fix the dynamical bias of the MT-RNN layer to a single point estimate (i.e.\ no dependence on $\z$). Smooth variation of this parameter otherwise resulted in `jumps' when interpolating between sequences styles. A particularly stark example is given in Figure \ref{fig:mocap:interp_MTRNN_vs_MTBias} which visualizes various sequence predictions of a given joint while interpolating the latent $\z$ from a `strutting' (blue)  to `angry' (red) style. The top figure shows the model output for the MT-RNN model where the bias is fixed wrt.\ $\z$, the bottom shows the output for the model where \emph{only} the bias depends on $\z$. We speculate that modulating the bias induces bifurcations in the state space, whereas adapting the transition matrix allows for smooth interpolation. This is consistent with the results of \citet[][]{sussillo2013opening}.

\subsubsection{Related Work in the Mocap Literature} \label{sec:mocap:related}

Generative models for mocap data include those proposed by \citet{wang2008gaussian}, \citet{taylor2010dynamical},  \citet{holden2016deep}. Competitive RNN-based approaches are introduced in \citet{fragkiadaki2015recurrent}, and \citet{martinez2017human} introduce the idea of sequence-to-sequence or `sampling-based' training \citep[cf.][]{bengio2015scheduled} in order to avoid predictions converging to a mean pose. We are not aware of any work which learns style-specific representations within the same model, except for \citet{mason2018few} who use residual adapters (and note that no quantitative results are given in their paper); much less of any models which permit style \emph{interpolation}. The ideas of \citet{tulyakov2018mocogan, hsieh2018learning} (discussed in Sec.\ \ref{sec:MTDS:related}) may perhaps be applied, but such architectures modulate the style simply via changing the bias or input to the recurrent generation network \citep[similar to][]{miladinovic2019dssm}. While these papers propose additional modifications to the architecture or objective function, we wish to isolate the contribution of our latent $\z$,  motivating our decision to consider bias-customized variants as our primary point of comparison.


\subsection{Experimental Setup}  \label{sec:mocap:expmt-setup}
In order to understand the benefits of the MTDS approach, we consider a number of different experiments which include: performance under limited training data, predictions on novel styles, and style transfer. We provide further details of these in the results section (\secref{sec:mocap:results}). In the current section, we discuss the competitor models  (\secref{sec:mocap:expmt-setup:models:benchmark}) and provide an overview of learning and inference (\secref{sec:mocap:expmt-setup:models:learninf}).

\subsubsection{Benchmark Models} \label{sec:mocap:expmt-setup:models:benchmark}

For comparison, we implement a standard 1024-unit GRU pooled over all styles, which serves both as a competitor model \citep{martinez2017human} and an ablation test. This pooled model can perform some `multi-task'-style customization, but in an implicit and black-box manner. Style inference follows the approach of \citet{martinez2017human}, where an initial seed sequence $\y_{1:\Tenc}$ is provided to the network before prediction.  In our experiments we use $\Tenc=64$ frames. This is essentially a  sequence-to-sequence (seq2seq) learning architecture with shared weights between the encoder and decoder.

Secondly, we provide a restricted form of our MTDS model, where only the bias/offset of the RNN is a function of $\z$, providing a comparison to existing work (as discussed above; Section \ref{sec:mocap:related}). We denote this simpler approach as the \emph{MT-Bias} approach in contrast to the full \emph{MT} approach proposed in this paper.  We also provide constant baseline predictions of (i) the training set mean and (ii) the last observed frame of the seed sequence (`zero-velocity' prediction). 


\subsubsection{Learning and Inference} \label{sec:mocap:expmt-setup:models:learninf}

The standard benchmark models were trained by encoding a 64-step seed sequence, and predicting a 64-step forecast \emph{without} access to the true $\{\y_t\}$, in contrast to `teacher forcing'. We call this an `open-loop' criterion, as the model receives no feedback during a prediction cycle when training.\footnote{We borrow the terms `open-' and `closed-loop' from control theory, e.g.\ \citet{astrom2010feedback}, ch.\ 1.} This forces the model to learn to recover from its mistakes and was first introduced in this context in \citet{martinez2017human}. In case the advantage of this is not obvious to the reader, we also provide results using a standard closed-loop criterion (i.e.\ via teacher forcing) too. For the MT and MT-Bias models, we use a seed sequence to infer the latent $\z$, but not to infer the dynamic state $\x_0$ (as in a seq2seq model) to avoid information leakage about the style. Both of these models are learned using the variational procedure described in \secref{sec:MTDS:learning} and hyperparameters for all models were chosen with respect to a 12.5\% validation set. We optimized the posterior distribution of each $\z^{(i)}$ directly for the sake of simplicity, but one can use a recurrent inference network instead such as that proposed in \citet{fabius2015variational}.  We report results for a variety of latent dimensions to provide the reader with an intuition of this hyperparameter's importance. Further details regarding learning can be found in Appendix \ref{appdx:mtds:expmt-details:learn-inf}. The predictive posterior of \eqref{eq:inference-pred-post} is used for predictions for MT and MT-Bias models at test time, but due to use of the length-64 seed sequence we found that the posterior was sufficiently concentrated that a MAP estimate of $\z$ performed similarly to our sequential AdaIS method.

\subsection{Results} \label{sec:mocap:results}

\begin{figure*}
    \centering
    \scalebox{0.6}{\input{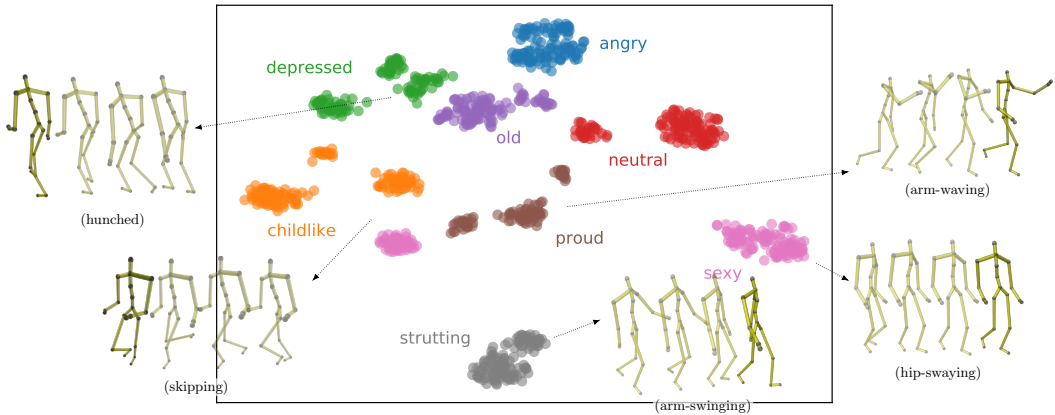}}
    \caption{$k=8$ mean embedding of each sequence segment, visualized via t-SNE, coloured by its (unseen) task label.}
    \label{fig:mocap:latent-viz}
\end{figure*}

Figure \ref{fig:mocap:latent-viz} shows a t-SNE plot \citep{van2008visualizing} of the $k=8$ mean embedding of $\z$ for each of the 64-frame segments in the data set, coloured by the true style label. (We remind the reader that this label is unavailable during training.) Our MTDS model can successfully disambiguate the styles without supervision, and in fact provides a more fine-grained representation than the original labels. Many walking styles have at least two sub-styles (e.g.\ `childlike' comprises both skipping and juggling motions). This visualization suggests that our MTDS model can indeed learn a useful manifold over sequence styles; further investigation in \secref{sec:mocap:expmt4} validates the intermediate points on this manifold, and demonstrates the potential of style interpolation.

This granular description of sequence style, and customization of predictions is unavailable from any of our competitor models. In the case of standard GRUs, style customization is occurring, but it is entangled in the various hidden units, and unavailable to an end user. Moreover, it cannot be controlled should a different style be desired. MT-Bias models appear to achieve a poorer latent representation than the full MT approach (see Appendix \ref{appdx:mocap:results:qual} where the MT-Bias shows substantial conflation of styles). This results in significantly less control over style than our MTDS model, as is demonstrated in the style transfer experiments in \secref{sec:mocap:expmt3}, as well as poorer inter-style interpolation (as e.g.\ in Figure \ref{fig:mocap:interp_MTRNN_vs_MTBias}).

To demonstrate the benefit of the MTDS approach quantitatively, we provide results from three experiments, concerning: data efficiency (Sec.\ \ref{sec:mocap:expmt1}), performance on unseen walking styles (Sec.\ \ref{sec:mocap:expmt2}), and style transfer (Sec.\ \ref{sec:mocap:expmt3}). We conclude in Section \ref{sec:mocap:conclusion}.

\subsubsection{Data Efficiency} \label{sec:mocap:expmt1}

\begin{table}
\centering
\footnotesize
\begin{tabular}{l c ccc c ccc}
\toprule
 && \multicolumn{6}{c}{\bf MSE}  \\ \cmidrule{3-8}
  && \multicolumn{6}{c}{Training set size} \\
 \textbf{Model} && 3\% &  7\% & 13\% & 27\% & 53\% &  97\% \\
 \cmidrule{1-1} \cmidrule{3-8}
Training mean &&  0.76 & 0.76 & 0.72 & 0.73 & 0.73 & 0.73\\
Zero-velocity &&  1.23 & 1.23 & 1.23 & 1.23 & 1.23 & 1.23 \\
\addlinespace[0.15cm]
STL GRU (open loop) && 1.11 & 0.88 & 0.40 & 0.33 & \bf 0.18 & \bf 0.18\\
\addlinespace[0.15cm]
Pooled GRU (closed loop) && 0.79 & 0.61 & 0.82 & 0.87 & 0.76 & 1.21\\
Pooled GRU (open loop) && 0.69 & 0.52 & 0.36 & 0.29 & \bf 0.16 & \bf 0.16\\ \addlinespace[0.15cm]
MT Bias ($k=3$) && 0.93 & 0.44 & 0.30 & \bf 0.21 & \bf 0.14 & \bf 0.16 \\
MT Bias ($k=5$) && 0.98 & 0.44 & 0.30 & \bf 0.20 & \bf 0.14 & \bf 0.16 \\
MT Bias ($k=7$) && 0.94 & 0.49 & 0.30 & \bf 0.21 & \bf 0.15 & \bf 0.16 \\
\addlinespace[0.15cm]
MTDS ($k=3$) && 0.62 & 0.34 & 0.35 & \bf 0.21 & 0.21 & 0.19 \\
MTDS ($k=5$) && \bf 0.53 & \bf 0.29 & \bf 0.22 & \bf 0.19 & \bf 0.15 & \bf 0.16 \\
MTDS ($k=7$) && \bf 0.51 & \bf 0.27 & \bf 0.24 & \bf 0.20 & \bf 0.16 & \bf 0.18 \\
\bottomrule
\end{tabular}
\caption{Experiment 1 (data efficiency): MSE for length-64 predictions where training set size is expressed as a fraction of the original data set. The best performances for each training set size (up to $\alpha=0.05$ significance---see text) are shown in bold.}
\label{tbl:mocap:results:mtl}
\end{table}

We test the conventional advantage of MTL by considering reduced subsets of the original data set. We perform six experiments which use between $2^8$ to $2^{13}$ frames per style (logarithmically spaced) for training, with sampling stratified carefully across major variations of all styles. A `single-task' 1024-unit GRU benchmark is also included for comparison, which is trained and tested on a single style. In all cases, the test performance (MSE) is calculated from 4 held-out sequences from each style (64 frames each), and averaged over all styles. 

\begin{figure}
    \centering
    \hspace*{-20pt}
    \begin{subfigure}[b]{.45\linewidth}
        \centering
        \includegraphics[trim={24 10 11 10}, clip, width=0.99\textwidth]{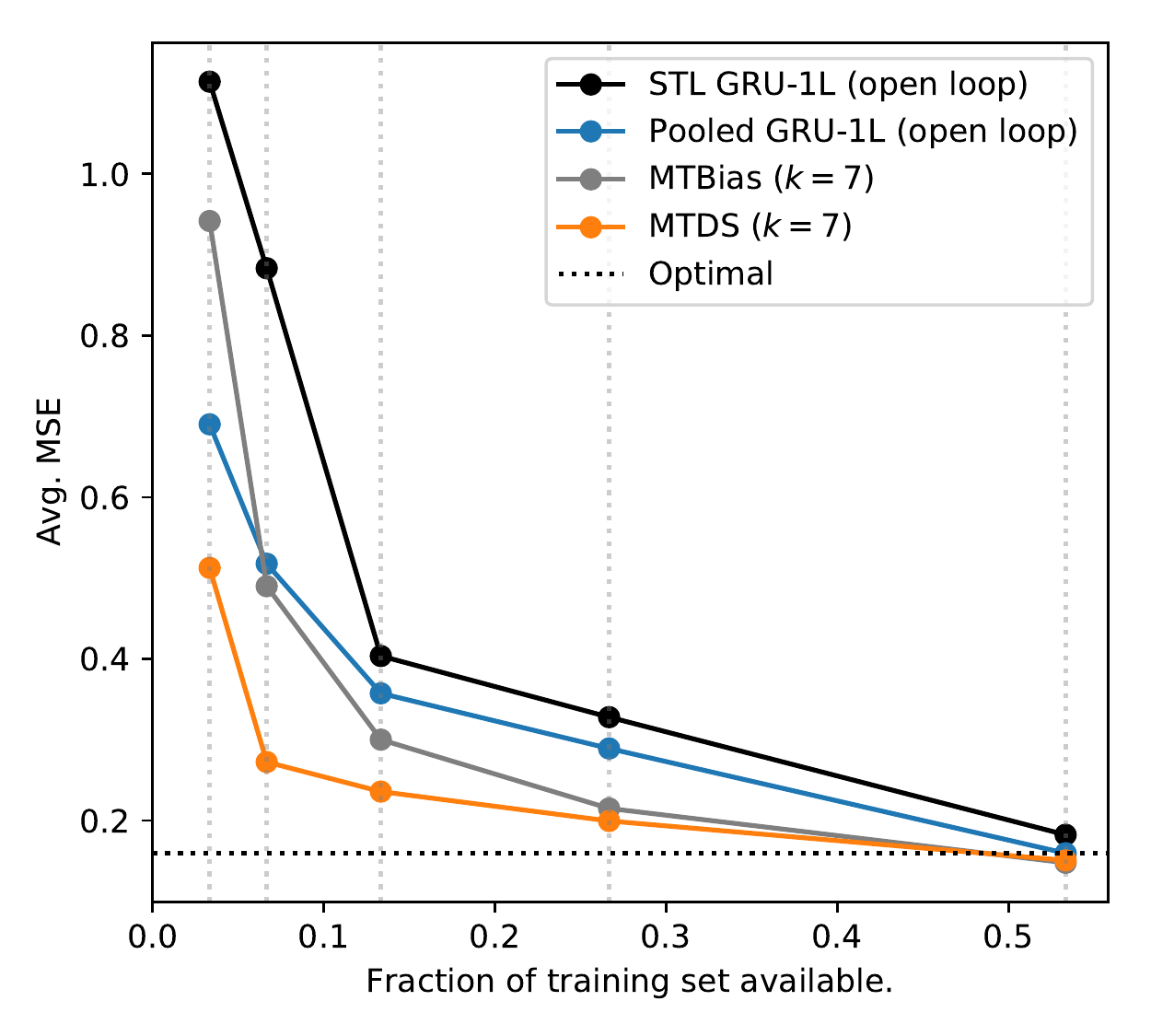}
        \caption{}
        \label{fig:mocap:results:mtl}
    \end{subfigure}~
    \begin{subfigure}[b]{0.45\linewidth}
        \centering
        \includegraphics[trim={24 10 11 10}, clip, width=0.99\textwidth]{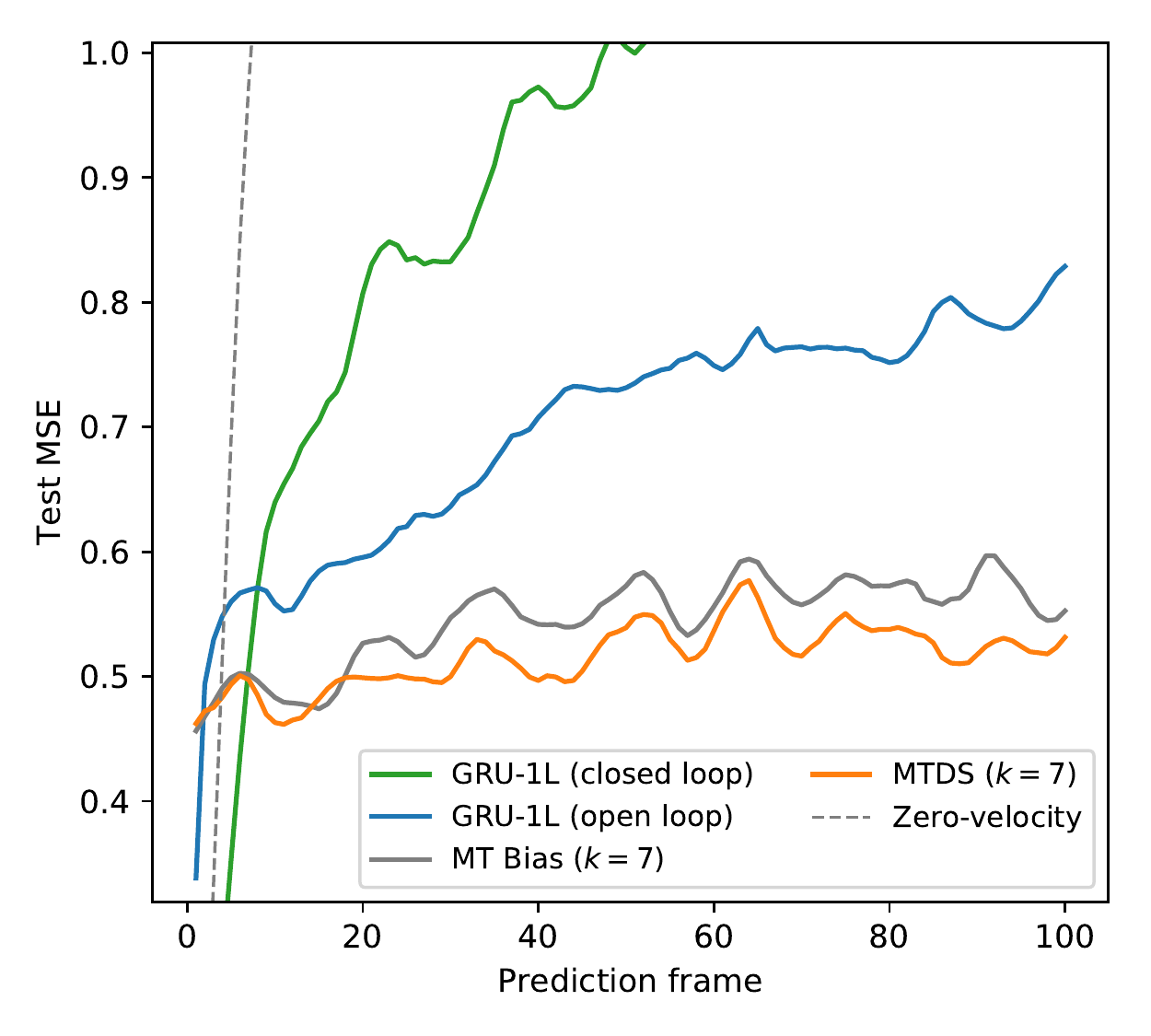}
        \caption{}
        \label{fig:mocap:results:tl}
    \end{subfigure}
    \caption{\captiona\ Experiment 1 (data efficiency): out-of-sample MSE by \% of training set seen. The performance achieved by the GRU models for the entire training set is shown as the `Optimal' dashed line. \captionb\ Experiment 2 (novel test data): MSE performance (avg over folds).}
\end{figure}

The results for the six data set sizes are shown in Table \ref{tbl:mocap:results:mtl}.\footnote{Some example animations can also be found from the linked video in \secref{sec:mocap:expmt4}.} Results in each column are compared to the best performing model using a paired $t$-test; the results which have comparable performance to the best model (i.e.\ are not significantly different at an $\alpha=0.05$ level) for the held out sequences are shown in bold. The open-loop GRU \citep{martinez2017human} performs far better than the standard closed-loop variant, and we will omit the latter from further discussion. The results show strong performances for the MTDS approach, which tends to perform best for $k>3$ (results for $k=5$ and $k=7$ are not significantly different). For small data sets (< 50\% of the data set), we observe advantages from all multi-task models (including the Pooled GRU) over a STL approach. However, the MTDS demonstrates far superior performance than the other MT approaches for smaller data sets; achieving a MSE of 0.27 after only 7\% of the data set. The MT-Bias model requires more than twice this amount of data to obtain the same performance, and the Pooled GRU requires more than four times this amount. These results are plotted graphically in Figure \ref{fig:mocap:results:mtl}. For data set sizes up to (and including) 13\% of the original, the MTDS is equal or better across \emph{all} styles and data set sizes, except for the  \{`angry',  `sexy'\} styles for the smallest data set (data not shown).


\subsubsection{Novel Test Data} \label{sec:mocap:expmt2}

Our second experiment investigates how well the MTDS can generalize to novel sequence styles. This is similar to a domain adaptation or zero-shot learning task. We consider a leave-one-out (LOO) procedure with eight folds, where each fold has a training set comprising 7 styles, and a test set comprising the held-out style. We consider the deterioration of predictive MSE over a large time window ($\tau \le 200$ frames, ca.\ 7 seconds). It can be relatively easy to predict $\tau$-steps ahead for $\tau \le 10$ \citep[see][]{martinez2017human} even for novel sequences, but the error usually deteriorates with increasing $\tau$. Our results report the predictive MSE for each $\tau$ averaged over the 8 folds, and over 32 different starting locations within each fold. The competitor models are as above (excluding the STL model), but we also include a 2-layer GRU (with 1024 units in each layer) as the training set is larger than in the previous experiment and can reliably learn such a model.

A summary of results is shown in Figure \ref{fig:mocap:results:tl}, where the axes are truncated for clarity. The standard Pooled GRUs work well for small values of $\tau$ but degrade very quickly. The closed-loop variants perform the best for $\tau \le 5$ but degrade even more rapidly than the open-loop approach. The deteriorating results for these models are consistent with the inputs moving the state into an area where the dynamics have not been trained. In contrast, the MTDS and MT-Bias models find a better customization which evidences very little worsening over the predictive interval. Importantly, these models are able to `remember' their customization over long intervals via the latent $\z$. The $k=7$ MTDS shows equal or better performance to the pooled GRU on all styles for $\tau \ge 10$, and its initial performance may perhaps be improved via interpolation from the seed sequence given the performance of the `zero-velocity' baseline for $\tau \le 5$. The results of the 2-layer competitors are shown in  Table \ref{tbl:mocap:results:tl}, but they achieve similar performance to the 1-layer models on aggregate (a similar result is suggested in \citeay{martinez2017human}).

\begin{table}
\centering
\footnotesize
\begin{tabular}{l c ccc c ccc}
\toprule
 && \multicolumn{6}{c}{\bf MSE}  \\ \cmidrule{3-8}
 \textbf{Model} && $\tau=5$  & $\tau=10$ & $\tau=20$ & $\tau=50$  & $\tau=100$ & $\tau=200$ \\ \cmidrule{1-1} \cmidrule{3-8}
Training mean &&  1.04 & 1.04 & 1.05 & 1.04 & 1.06 & 1.07 \\
Zero-velocity && 0.69 & 1.20 & 1.37 & 1.21 & 1.35 & 1.48 \\ \addlinespace[0.15cm]
Pooled 1-layer GRU (closed loop) && \bf 0.35 & 0.64 & 0.81 & 1.00 & 1.45 & 7.28 \\
Pooled 2-layer GRU (closed loop) && \bf 0.34 & 0.61 & 0.79 & 0.97 & 1.41 & 6.34 \\ \addlinespace[0.15cm]
Pooled 1-layer GRU (open loop) && 0.56 & 0.56 & 0.60 & 0.73 & 0.83 & 0.92 \\
Pooled 2-layer GRU (open loop) && 0.53 & 0.55 & 0.59 & 0.73 & 0.85 & 0.94 \\ \addlinespace[0.15cm]
MT Bias ($k=3$) && 0.60 & 0.60 & 0.58 & 0.59 & 0.64 & 0.63 \\
MT Bias ($k=7$) && 0.50 & \bf 0.48 & 0.53 & 0.57 & \bf 0.55 & \bf 0.63 \\
MTDS ($k=3$) && 0.61 & 0.62 & 0.59 & 0.61 & 0.63 & 0.63 \\
MTDS ($k=7$) && 0.49 & \bf 0.46 & \bf 0.50 & \bf 0.54 & \bf 0.53 & \bf 0.61 \\
\bottomrule
\end{tabular}
\caption{Experiment 2 (novel test data): average predictive MSE at $\tau=5, 10, 20, 50, 100, 200$. The best performing model(s) (up to $\alpha=0.05$ significance) for each $t$ is highlighted in bold.}
\label{tbl:mocap:results:tl}
\end{table}

These experiments demonstrate that it can be crucial to retain control over the task inference for novel data, rather than delegating it to a black-box procedure; the implicit inference of standard GRU networks can perform very poorly when presented with unexpected inputs. For this experiment, while the full MTDS consistently outperforms the MT-Bias approach, the difference is not large. In practice, perhaps either could be used. We note that all models struggle to capture the arm movements of the unseen styles since these are often entirely novel. Customization to the legs and trunk is easier since less extrapolation is required (see animation videos linked in \secref{sec:mocap:expmt4}).

\subsubsection{Style Transfer} \label{sec:mocap:expmt3}

Finally, we investigate how much user control is available via the latent code, $\z$. We can hold the input trajectory $\{\bu_t\}$ constant, and vary the latent $\z$ from its inferred position. In theory, this should result in style transfer: the same trajectory of locomotion but performed in a different style. This is unavailable from any existing GRU approaches (see related work), and hence in this section we can only compare the full MTDS with the restricted MT-Bias model. For each pair of styles $(s_1, s_2)$ we investigate style transfer from a source sequence of style $s_1$ to a target style $s_2$. Due to the \emph{within}-style variation, we use four different source sequences for each pair $(s_1, s_2)$.  We learn the embedding using a $k=8$ model, and choose a single latent value $\z^{(s_2)}$ for each of the eight target styles $s_2=1,\ldots,8$ to perform the style transfer; see Appendix \ref{appdx:mocap:results:style-tf} for more details. Evaluation is performed via use of a classifier, for which we use a 512-unit GRU to encode the sequence followed by a 300-unit hidden layer MLP with multinomial emissions. The classifier is trained on the complete data using the (previously unused) labels from \citet{mason2018few}.  Qualitative results are  available via the videos linked in \secref{sec:mocap:expmt4}, and further experimental details are given in Appendix \ref{appdx:mocap:results:style-tf}.

\begin{table}
    \footnotesize
    \centering
    \begin{tabular}{ccccccccc}
        
        \toprule
        Target & Angry & Child & Depr.\ & Neut.\ & Old & Proud & Sexy & Strut \\ \midrule
        MT Bias & 0.78 & 0.59 & 0.65 & 0.79 & 0.55 & 0.71 & 0.55 & 0.71 \\
        MTDS    & \bf 0.86 & \bf 0.95 & \bf 0.81 & \bf 0.93 & \bf 0.86 & \bf 0.82 & \bf 0.90 & \bf 0.92 \\
        \bottomrule
    \end{tabular}
    \caption{Experiment 3 (style transfer): classifier probability of target style  averaged over all input styles.}\label{tbl:mocap:expmt3}
\end{table} 

The results are summarized in Table \ref{tbl:mocap:expmt3}, which provides the average `probability' assigned by the classifier for each \emph{target} style $s_2$, averaged over all the input sequences where the source $s_1 \ne s_2$. (The best performing model for each style is highlighted in bold.) Successful style transfer should result in the classifier assigning a high probability to the target style. The results suggest that the style can generally be well controlled by $\z$ in the case of the full MTDS, but the MT-Bias model exhibits reduced control for some (source, target) pairs. Style transfer appears to be easier between more similar styles; the lowest scores tend to be for transfer \emph{from} the `childlike' and `angry' styles (which have unusually fast trajectories) or the `old' style (which has unusually slow ones). Appendix \ref{appdx:mocap:results:style-tf} provides a more detailed comparison of this  performance. 

These results demonstrate that an MTDS approach can provide end-user control of task-level attributes, in this case resulting in style transfer. We have also seen that our full MT parameterization provides far greater control than the limited MT-Bias approach. Some insight is available from the respective latent representations (Appendix \ref{appdx:mocap:results:qual}), for which the MT-Bias model has apparently conflated a variety of styles. In this case, such styles can only be disambiguated via the inputs, and changing the latent $\z$ alone may result in unrecognizable changes. Further improvements to the MTDS may be possible via use of domain knowledge or adversarial objectives; we leave this to future work.

\subsubsection{Qualitative Results} \label{sec:mocap:expmt4}

Finally, we discuss the qualitative results of our MTDS model, making use of the latent representation (as in Figure \ref{fig:mocap:latent-viz}, page \pageref{fig:mocap:latent-viz}) and animations of the predictions. 

The latent representation of the MTDS appears sensible; similar motions are located close together, and differing ones are further apart. For instance, the neighboring `old' and `depressed' styles in Figure \ref{fig:mocap:latent-viz} both involve leaning over, and the neighboring `childlike' and `sexy' clusters both comprise `skipping'-type motions. As such, the learned embeddings appear to capture more information than the original labels. Smooth interpolation between styles is also available from the full MTDS model as suggested by Figure \ref{fig:mocap:intro:morphing} (page \pageref{fig:mocap:intro:morphing}), and Figure \ref{fig:mocap:interp_MTRNN_vs_MTBias} (top, page  \pageref{fig:mocap:interp_MTRNN_vs_MTBias}); this can be verified in the animations linked below. Interpolation of the style manifold of the MT-Bias model tends to result in `jumps', as suggested by Figure \ref{fig:mocap:interp_MTRNN_vs_MTBias} (bottom).

The animations for all experiments have been  collected into a project webpage.\footnote{\url{https://sites.google.com/view/mtds-customized-predictions/home}} They form a crucial part of the model evaluation, which cannot be adequately summarized in static figures. These animations include:
\begin{enumerate}[topsep=5pt,itemsep=0ex,partopsep=1.5ex,parsep=1.5ex]
    \item \textbf{In-sample predictions}: demonstrating the best possible performance of the models.
    \item \textbf{MTL examples} from \secref{sec:mocap:expmt1} comparing the quality of animations and fit to the ground truth for two limited training set sizes (6.7\% and 13.3\% of the full data).
    \item \textbf{Novel test examples} from \secref{sec:mocap:expmt2} showing the adaptions obtained by each model to novel sequences.
    \item \textbf{Style morphing}. This animation demonstrates the effect of interpolating $\z$ over time by \emph{morphing} between all eight styles, which goes beyond the style transfer of \secref{sec:mocap:expmt3}.
\end{enumerate}
%

\subsection{Conclusion} \label{sec:mocap:conclusion}

We have shown that the MTDS framework can be applied to RNN-type models and can capture the inter-sequence differences of a training set $\gD$ using the latent variable $\z$. This is not limited to simple low-dimensional sequences, but can be applied to complex sequences with highly nonlinear relationships, such as mocap data. Our experiments have suggested a number of advantages over existing approaches. Firstly, the MTDS can result in substantial improvements in performance in small data settings. Secondly, the same model can avoid performance deterioration under dataset shift, and thirdly, be used to perform highly effective style transfer. Finally, the resulting sequence family admits interpolation between its members, which for this application produces smooth morphing between walking styles.



\section{Application to Drug Response Data} \label{sec:mtpd}

Our second application of the multi-task dynamical system is in the medical domain. In this context (unlike many high-profile applications of machine learning) sample sizes are small, and consequences of mistakes can be severe. For this reason, models tend to be simple, inflexible and predict average effects. Personalization of these models is 
often thought to require larger samples and more covariates (e.g.\ genomic and proteomic data), and thus the necessary data sets may be unavailable for many years to come, and carry increased requirements for secure infrastructure and privacy protection. Our notion of a sequence family, modelled by a multi-task dynamical system (MTDS), allows us to take a step towards personalization without this additional data. 

In this section, we consider the example of predicting patient response to the anaesthetic agent propofol. We learn a family of possible responses, using a MTDS, and provide an increasingly personalized prediction as more observations are seen. Our goal is not to provide the best possible model, but to show how the MTDS can personalize the \emph{existing model} that is in use by current practitioners. We may hence improve predictions while maintaining trust. If an alternative model becomes acceptable to clinicians in future, then the MTDS can equally be applied to this too.

This section is structured as follows: \secref{sec:mtpd:bground} introduces the modelling background, \secref{sec:mtpd:model} describes our proposed base model and the MTDS variant.
\secref{sec:mtpd:experimental} introduces the experiments, and the results are  provided in \secref{sec:mtpd:results}. We conclude in \secref{sec:mtpd:conclusion}.

\subsection{Background} \label{sec:mtpd:bground}

In this section, we introduce the modelling task (\secref{sec:mtpd:bground:task}), discuss existing approaches (PK/PD models, \secref{sec:mtpd:bground:pkpd}), and provide the necessary background of our target model (\secref{sec:mtpd:bground:pd}).

\subsubsection{Introduction} \label{sec:mtpd:bground:task}

In order to sedate a patient, an anaesthetist initially targets a certain blood concentration of an anaesthetic agent. In the case of propofol, this may be between 0.5 - 5.0 $\mu$g/ml. The drug is administered via use of an intravenous infusion pump, which uses an internal model to provide the desired concentration. The response of the patient is quantified via vital signs, providing an important feedback loop to anaesthetists; in our case the vital signs are systolic and diastolic blood pressure and BIS,\footnote{The Bispectral Index (BIS) of \citet{myles2004bispectral} is a proprietary scalar-valued transformation of EEG signals which attempts to quantify the level of consiousness. BIS incorporates time-domain, frequency-domain, and bispectral analysis of the EEG to obtain a scalar between 0 (deep anaesthesia) and 100 (awake).} a measure of consciousness.

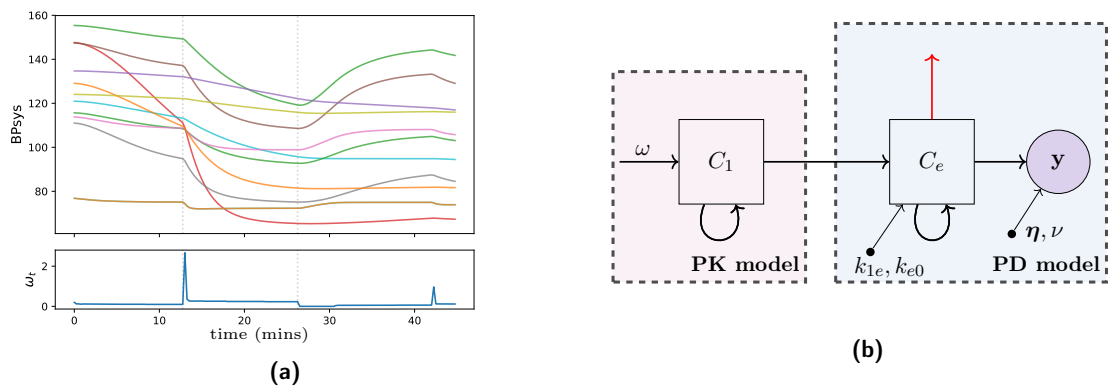
\begin{figure}
    \centering
    \advance\leftskip-0.25cm
    \advance\rightskip-0.25cm
    \begin{subfigure}{0.5\textwidth}
        \begin{tikzpicture}
            \node[anchor=south west,inner sep=0] (image) at (0,0) {\includegraphics[height=4.5cm]{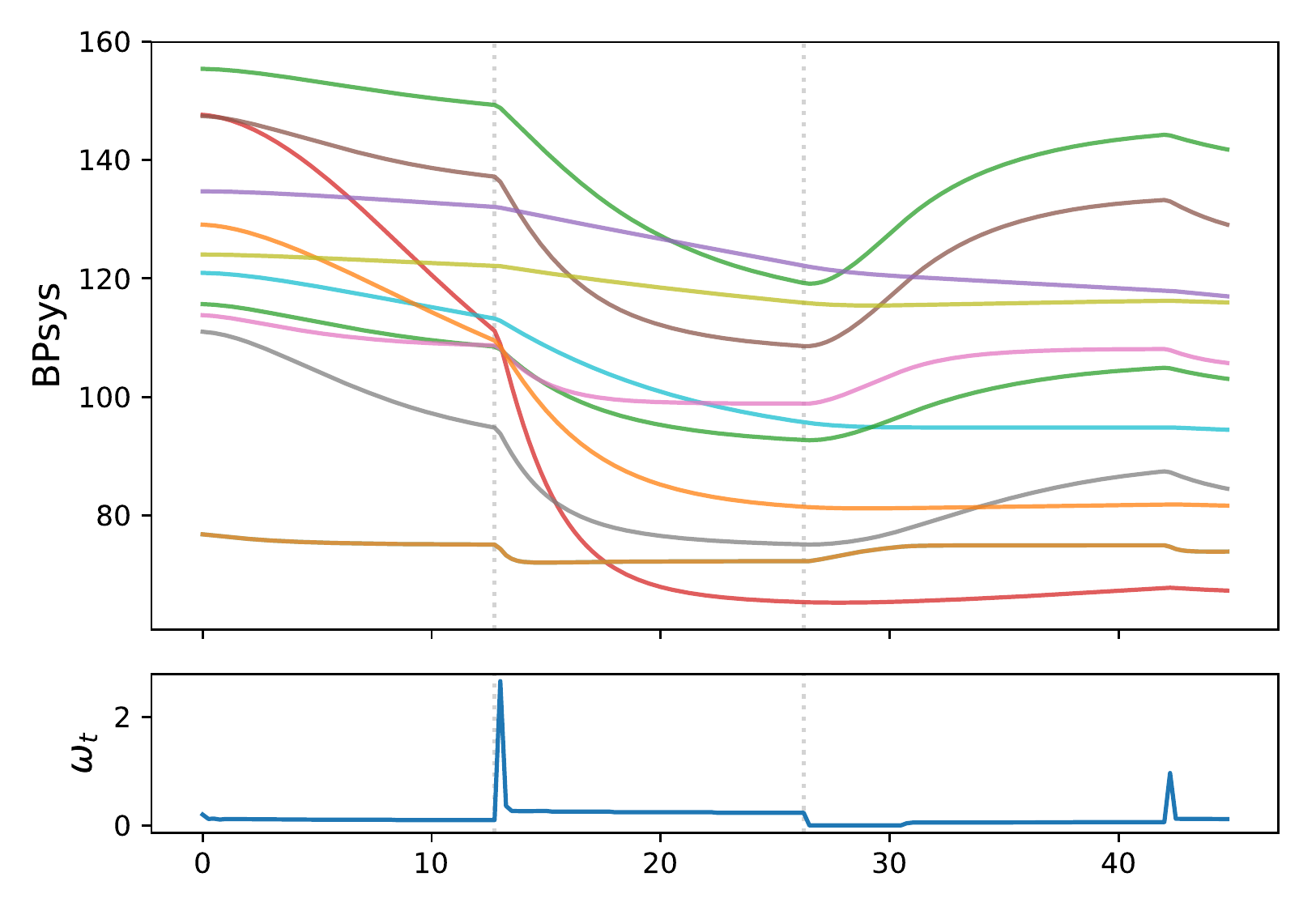}};
            \begin{scope}[x={(image.south east)},y={(image.north west)}]
            \node[anchor=south west,inner sep=0] (xlbl) at (0.43,-0.015)      {\tiny time (mins)};
            \end{scope}
        \end{tikzpicture}
        \caption{}
        \label{fig:mtpd:possible-responses}
    \end{subfigure}
    \begin{subfigure}{0.5\textwidth}
        \centering
        \scalebox{0.8}{
\iftrue
\begin{tikzpicture}[customsq/.style={minimum size=40pt,draw},
                    latentdiscrete/.style={minimum size=30pt,draw},
                    visiblects/.style={circle,minimum size=30pt,fill=purplevis!50, draw},
                    visiblediscrete/.style={minimum size=30pt,fill=purplevis!50, draw},
                    invis/.style={minimum size=0pt,fill=white, inner sep=0pt},
                    customsqgrey/.style={minimum size=40pt, color=lightgray, text=lightgray, draw}]
\definecolor{purplevis}{RGB}{185,164,219}

\draw[color=black!70!white, dashed, fill=RedViolet!5!white, line width=0.5mm] (-2.8,-1) rectangle ++(3.2,3.5);
\draw[color=black!70!white, dashed, fill=RoyalBlue!5!white, line width=0.5mm] (0.9,-1) rectangle ++(4.5,4.3);

\node [customsq]  (b1) at (-1, 1) {$C_1$};
\node [customsq]  (be) at (2.5, 1) {$C_e$};
\node [visiblects]  (by) at (4.6, 1) {$\y$};
\node [invis]     (b02) at (2.5, 2.8) {};

\draw[thick, ->] (b1) to (be);
\draw[thick, ->] (be) to (by);

\draw[thick, ->, color=red] (be) to (b02);

\node [invis]     (control_in) at (-2.7,1) {};
\draw[thick, ->] (control_in) to (b1);
\node [invis]     (control_in) at (-2.7,1) {};
\node [invistrans] (control_text) at (-2.3, 1.2) {$\omega$};





\node [blackdot] (park1e) at (1.4, -0.5) [right] {};
\node [invistrans] (control_text) at (1.2, -0.75) [right] {$k_{1e}, k_{e0}$};

\draw[->] (park1e) to (be);

\draw[thick, ->] (be) .. controls (2.0,-0.5) and (3.0,-0.5) ..  (be);
\draw[thick, ->] (b1) .. controls (-1.6,-0.5) and (-0.5,-0.5) ..  (b1);

\node [blackdot] (etanu) at (3.75, -0.2) [right] {};
\node [invistrans] (control_text) at (4.05, -0.2) [right] {$\mbf\eta, \nu$};

\draw[->] (etanu) to (by);

\draw[thick, ->] (be) .. controls (2.0,-0.5) and (3.0,-0.5) ..  (be);
\draw[thick, ->] (b1) .. controls (-1.6,-0.5) and (-0.5,-0.5) ..  (b1);

\node [invistrans, align=right] (control_text) at (-0.6, -0.7) {{\small \bf PK model}};

\node [invistrans, align=right] (control_text) at (4.4, -0.7) {{\small \bf PD model}};

\end{tikzpicture}
\fi}
        \vspace*{0.5cm}
        \caption{}
        \label{fig:mtpd:bground:pd}
    \end{subfigure}
    \caption{(a) Differing systolic blood pressure responses across patients to a given drug infusion shown in the bottom panel. The responses follow individual PD models trained offline. (b) A standard PK/PD model. The effect compartment $C_e$ of the PD model attaches to the central compartment, $C_1$ of the PK model. The concentration in $C_e$ is assumed to be directly related to the pharmacodynamic effect, $\y$, e.g.\ BPsys as in the opposite figure.}
\end{figure}

The patient response to the drug infusion depends on their physiology, resulting in substantial inter-patient variation. Some examples of modelled responses to the same infusion sequence are shown in Figure \ref{fig:mtpd:possible-responses} for systolic blood pressure; real data exhibit similar inter-patient differences. The initial dosage targets a BIS value in the range 40-60, but the vital signs must be monitored on a continual basis to ensure the patient stays within the therapeutic window. This task is made substantially harder due to the lag between dose and response. Further complications are introduced in practice (e.g.\ due to surgical stimuli or multiple drugs); we limit the scope of this work to predict the response to a single drug (propofol) in a stationary environment. This is an important first step towards a control system for steady state anaesthetic maintenance, with the potential to free up considerable time from practicing anaesthetists.

\subsubsection{PK/PD Models}  \label{sec:mtpd:bground:pkpd}

Drug response is typically modelled via pharmacokinetic/pharmacodynamic (PK/PD) models. See \citet{bailey2005drug} for an introduction. The PK component models the distribution of drug concentration throughout the body.  There is substantial existing work for personalizing PK models (important examples include \citealt{marsh1991pharmacokinetic, schnider1998influence, white2008use, eleveld2018pharmacokinetic}). Various studies \citep[see e.g.][]{masui2010performance,glen2014comparison, huppe2019eleveldretrospective} have compared the predictive performance of propofol PK models currently used in clinical practice. These studies have confirmed a degree of bias and inaccuracy of the models but overall their performance is considered by most clinicians to be adequate for clinical use (at least within the populations in which they were developed). 

A PD model maps the drug distribution estimated by the PK model to the physiological effect. In practice, it is difficult to provide analytic models of PD processes, and most proposals instead take an empirical approach \citep[][]{bailey2005drug}. However, despite the relatively simple models proposed, there is comparatively little work on their personalization. Many commercially available implementations of the Marsh and Schnider models use fixed population-level parameters, but it is widely accepted by practicing anaesthetists that there exists a significant amount of inter-individual variability in PD response to propofol, a point recently demonstrated in \citet{van2020comparison}. \citet{eleveld2018pharmacokinetic} adjust the PD parameters based on age, but the available improvements are relatively small. Since there is more scope to improve this component, we focus our efforts on providing a personalized PD model, and use the PK model of \citet{white2008use} as-is.


\subsubsection{PD Models} \label{sec:mtpd:bground:pd}

Most PD models propose that the physiological effect can be determined directly (up to random noise) from the drug concentration at some \emph{effect site} ($C_e$).\footnote{For a wider variety of PD models see the review in \citet{mager2003diversity}.} This is a notional physiological site which contains the receptors bound by the drug compound of interest. The concentration of the drug at $C_e$ is affected by the concentration in the blood plasma, modelled in the PK model as the \emph{central compartment}, $C_1$. We will denote the concentration of the two sites as $x(t)$ and $c_1(t)$ respectively. A lag between these two concentrations is usually observed, and may be caused by multiple factors including distribution, receptor binding time, and the effects of intermediate substances. For more details see e.g.\  \citet{holford2018pharmacodynamic}.


A standard choice in the PD literature is to model the effect site concentration $x(t)$ by the following differential equation:
\begin{align}
    \frac{\dif x}{\dif t} = k_{1e} c_1(t) - k_{e0} x(t), \label{eq:mtpd:PD-ODE}
\end{align} 
where the rate of in-flow to the effect site is denoted $k_{1e}$ and  the elimination rate is denoted $k_{e0}$. We assume that the central compartment concentration $c_1(t)$ is available (via application of a PK model to the raw drug infusion sequence $\omega(t)$). A schematic is shown in Figure \ref{fig:mtpd:bground:pd}. 
Where multiple effects are observed simultaneously (e.g.\ BPsys, BPdia, BIS), it is common to use one effect site per observation channel, resulting in effect site concentrations $x_j(t)$ for $j= 1,\ldots, n_y$.
The relationship of $\x(t)$ to the observations $\y(t)$ is usually modelled by some nonlinear transformation $\ggamma$ plus white (Gaussian) noise, i.e.\ for a given time $t$, 
\begin{align}
    y_{j}(t) \,\sim\, \Normal{\ggammaj(x_j(t)), \mkern4mu\nu_j^{-1}} \label{eq:mtpd:PD-ggamma},
\end{align}
for $j=1,\ldots,n_y$ with parameters $\bfeta, \mbf \nu$. Most common choices of $\ggamma(\cdot)$ are sigmoidal in nature and include the Hill function \citep{wagner1968kinetics} and the generalized logistic sigmoid \citep{georgatzis2016ionlds}.



\subsection{Proposed Model} \label{sec:mtpd:model}

In this section, we consider learning a \emph{sequence family} of PD responses using an MTDS, conditioned on a propofol infusion sequence. At test time we can choose the most probable future response from the family by comparison to the patient's current observations. The parameterization of the base PD model is described in \secref{sec:mtpd:model:singletask} followed by the MTDS version in \secref{sec:mtpd:model:multitask} which enables online personalization. We compare this to related work in \ref{sec:mtpd:related}.


\subsubsection{Base PD Model} \label{sec:mtpd:model:singletask}

Our proposed base model is a relaxation of the PD model described above in discrete time. We assume access to the PK model prediction applied to each drug  infusion sequence. Specifically, let the inputs $\{u_t\}$ be the modelled central compartment concentration discretized on the unit grid $t=1,\ldots,T$, using the parameters of \citet{white2008use}. Denoting the effect site concentration at time $t$ as $x_{t}$, \eqref{eq:mtpd:PD-ODE} may be discretized as:
\begin{align}
    x_{t} &= \beta_{1} x_{t-1} + \beta_{2} u_{t-1},  \label{eq:mtpd:vanilla-pd-recurrence}
\end{align}
for some $\beta_{1}$, $\beta_{2}$, with no loss of generality if $c_1(t)$ is constant in each interval $(t-1, t]$. These coefficients are related to the rate constants as:
\begin{alignat}{2}
    \beta_{1} \,&=\, e^{-k_{e0}} \qquad && \Rightarrow \beta_1 \, \in \, (0,1)  \nonumber \\
    \beta_{2} \,&=\, \frac{k_{1e}}{k_{e0}} \left(1 - e^{-k_{e0}}\right) \qquad && \Rightarrow \beta_2 \, >\, 0,  \nonumber
\end{alignat}
since the rate constants are positive. We derive these relationships via use of Laplace transforms and the convolution theorem (see Appendix \ref{sec:MTPD-supp:model:wlog-pwise-const} for further details). Since $\beta_1 \in (0,1)$, the ARX(1) process in \eqref{eq:mtpd:vanilla-pd-recurrence} is stable and non-oscillating.

The nonlinear emission is modelled via a function $\ggamma(\cdot)$ with parameter vector $\bfeta$. We have found the choices in previous work (generalized sigmoid, Hill function) to be numerically unstable for gradient-based optimization or insufficiently flexible. Instead we use a basis of logistic sigmoid ($\sigma$) functions and express:
\begin{align}
    \ggamma(x) \,=\, \sum_{r=1}^L \eta_r\, \sigma(\,a_r (x - b_r)), \label{eq:mtpd:ggamma-def}
\end{align}
with constants $a_r < 0$ and coefficients $\eta_r \ge 0$ for all $r$. These constraints enforce the desired monotonicity that as concentration increases, the observations (blood pressure etc.) are non-increasing. We fitted an 8-dimensional basis with pre-selected constants $\{a_r, b_r\}_{r=1}^8$ chosen by optimising the fit to the learned generalized sigmoid functions used by \citet{georgatzis2016ionlds}. 


To complete the model, we introduce additional parameters $\mbf \beta_3$ and $\mbf \alpha$ which provide personalized offsets to the values of the effect site dynamics and the emission respectively. These are degrees of freedom one might expect in a dynamical system, but are not present in the usual PD formulation. The full model for a given patient can be written with $\x_t \in \R^{n_x}$ and $\y_t \in \R^{n_y}$ as:
\begin{subequations}
\label{eq:mtpd:model:singletask}
\begin{align}
    \x_t \,&=\,
    \bfbeta_1 \odot  \x_{t-1} \,+\, \bfbeta_2 u_{t}  \label{eq:mtpd:model:singletask:x} \\
     y_{tj} \,&=\, \ggammaus{j}(x_{tj} + \beta_{3j})
\, + \, \alpha_j \,+\, \epsilon_{tj}, \label{eq:mtpd:model:singletask:y}
\end{align}
\end{subequations}
$\epsilon_{tj} \sim \Normal{0,\, \nu_j^{-1}}$ for $j=1,\ldots,n_y$ and $t = 1, \ldots, T$. Here $\mbf \beta_j \in \R^{n_y}, j=1,\ldots,3$, with each dimension corresponding to the parameters of each channel's dynamics, $\mbf \eta_j \in \R^{L}, j=1,\ldots,n_y$ are the per-channel nonlinear coefficients (see eq.\ \ref{eq:mtpd:ggamma-def}) and  $\odot$ denotes elementwise multiplication. 

This results in a nonlinear deterministic-state dynamical system 
where each dimension is independent. A stochastic state might be considered as an extension to the standard PD approach, but preliminary investigation showed superior predictions with the deterministic approach. The parameters of the model are $\mbf \nu$ and $\bftheta = \{\mbf \alpha,$ $\mbf \beta_1, \mbf \beta_2, \mbf \beta_{3},$ $\bfeta_1, \ldots, \bfeta_{n_y}\} \in \R^d$, $d=36$, while $\mbf a, \mbf b$ are constants. In principle, $\mbf \alpha$ may be estimated prior to anaesthetic induction since it relates to pre-infusion patient-specific vitals levels.

\subsubsection{MTDS Model} \label{sec:mtpd:model:multitask}

We now discuss a version of this PD model which can achieve increasing personalization over time. Unlike the MT-RNN in \secref{sec:mocap}, it is not entirely impractical to place an  uninformative prior over $\bftheta$ and perform Bayesian inference online. But this `single task' (ST) approach fails to take advantage of the inductive bias from the training data, resulting in poor performance for patients with limited data, and poorly conditioned and expensive inference. An MTDS approach avoids these limitations; we describe its application below.

\begin{figure}
    \centering
    \advance\leftskip-0.25cm
    \advance\rightskip-0.25cm
    \begin{subfigure}{0.47\linewidth}
        \centering
        \scalebox{0.62}{
            \begin{tikzpicture}

\def\numseq{2}   
\def\xlabels{{"t-1","t","t+1"}}

\def\vsepnodes{2.3}  
\def\hsepnodes{2.5}

\foreach \x in {0,...,\numseq}
{
    \def\tzxpos{\hsepnodes*\x}
    \pgfmathsetmacro{\label}{\xlabels[\x]}
    \node[visiblects](y\label) at (\tzxpos,\vsepnodes) {$\mbf y_{\label}$};
    \node[latentdiscrete](x1\label) at (\tzxpos,0) {$\mbf x_{\label}$};
    \node[visiblects](u\label) at (\tzxpos,-\vsepnodes) {$\mbf u_{\label}$};
}

\pgfmathtruncatemacro{\numseqm}{\numseq - 1}
\pgfmathtruncatemacro{\numseqp}{\numseq + 1}
\pgfmathtruncatemacro{\numseqpp}{\numseq + 2}
\pgfmathtruncatemacro{\sinklabelm}{\numseq + 1}
\pgfmathtruncatemacro{\sinklabel}{\numseq + 2}
\def\tzsinkpos{\hsepnodes*\sinklabelm}
\node[invis](u\sinklabel) at (\tzsinkpos, -\vsepnodes) {\ldots};
\node[invis](x1\sinklabel) at (\tzsinkpos, 0) {\ldots}; {\ldots};
\node[invis](y\sinklabel) at (\tzsinkpos, \vsepnodes) {\ldots};
\def\sourcelabel{source}
\node[invis](u\sourcelabel) at (-\hsepnodes, -\vsepnodes) {\ldots};
\node[invis](x1\sourcelabel) at (-\hsepnodes, 0) {\ldots}; {\ldots};

\node[latentcts](z) at (-\hsepnodes+0.1, \vsepnodes-0.5) {$\z$};

\foreach \x in {0,...,\numseq}
{
    \pgfmathsetmacro{\label}{\xlabels[\x]}
    \draw[blackarrow] (u\label) -- (x1\label);
    \draw[blackarrow] (x1\label) -- (y\label);
}
\foreach \x in {0,...,\numseqm}
{
    \pgfmathsetmacro{\label}{\xlabels[\x]}
    \pgfmathsetmacro{\labelnext}{\xlabels[\x+1]}
    \draw[blackarrow] (x1\label) -- (x1\labelnext);
}
\foreach \x in {0,...,\numseq}
{
    \pgfmathsetmacro{\label}{\xlabels[\x]}
    \pgfmathtruncatemacro{\outangle}{-35+8*\x}
    \pgfmathtruncatemacro{\inangle}{135+3*\x}
    \draw[bluearrow] (z) to[out=\outangle,in=\inangle] (x1\label);
    \pgfmathtruncatemacro{\outangle}{35+8*\x}
    \ifthenelse{\equal{\x}{0}}{
        \pgfmathtruncatemacro{\inangle}{180}
        }{
        \pgfmathtruncatemacro{\inangle}{135+3*\x}
    }
    \draw[bluearrow] (z) to[out=\outangle,in=\inangle] (y\label);
}

\pgfmathsetmacro{\firstlabel}{\xlabels[0]}
\draw[blackarrow] ( $ (x1\sourcelabel)!.33!(x1\firstlabel) $ ) -- (x1\firstlabel);

\pgfmathsetmacro{\lastlabel}{\xlabels[\numseq]}
\draw[blackarrow] (x1\lastlabel) -- ( $ (x1\lastlabel)!.75!(x1\numseqpp) $ );

\end{tikzpicture}
        }
        \vspace*{1cm}
        \caption{}
        \label{fig:mtpd:gm}
    \end{subfigure}
    \begin{subfigure}{0.55\linewidth}
        \includegraphics[trim={0 15 40 50}, clip, width=\linewidth]{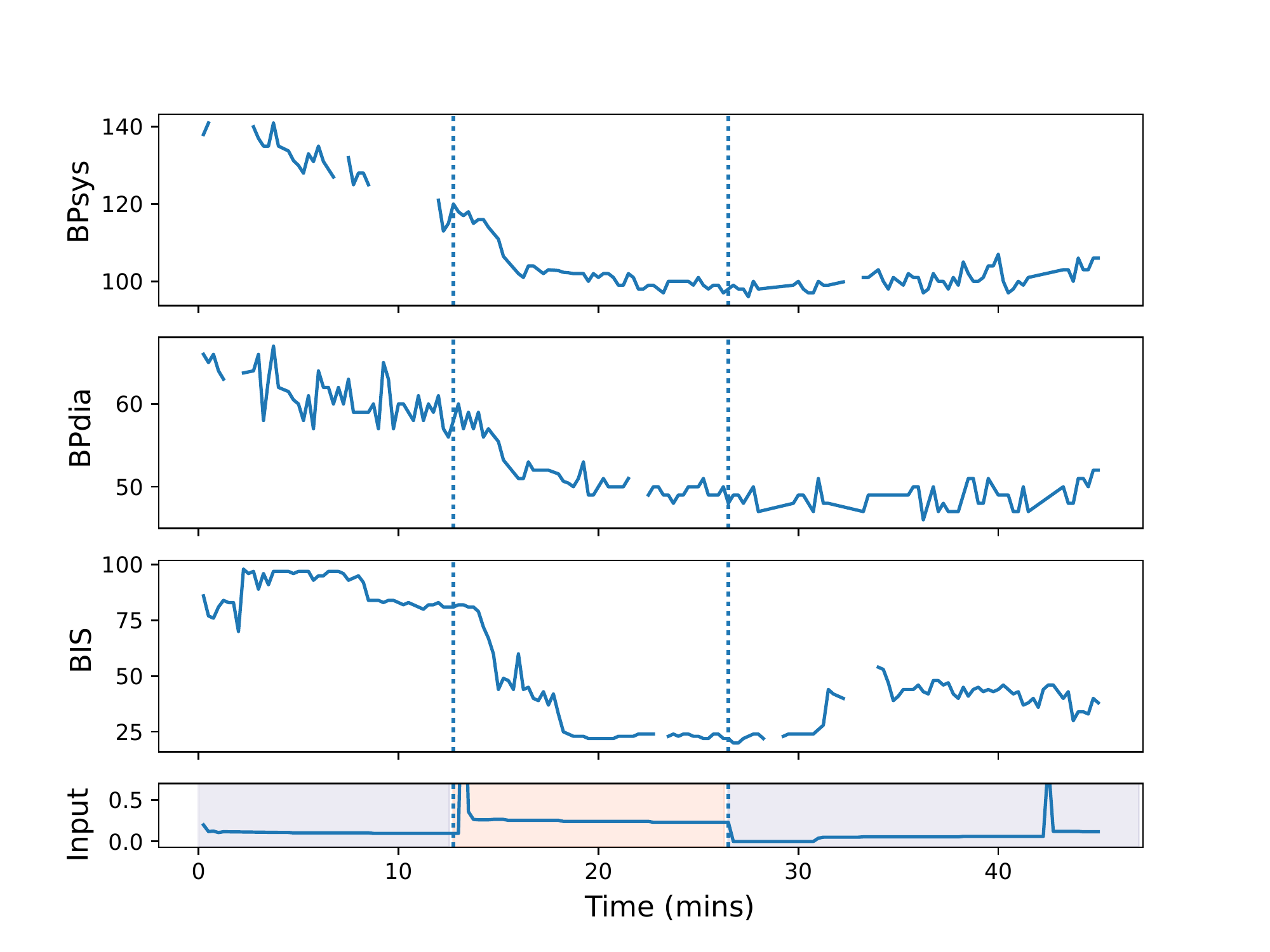}
        \caption{}
        \label{fig:mtpd:example_vitals}
    \end{subfigure}
    \caption{(a) MT PD model for a single patient; the superscript $i$ is omitted for clarity. (b) Vital signs ($\y_t$) and drug infusion ($u_t$) for an example patient.}
\end{figure}
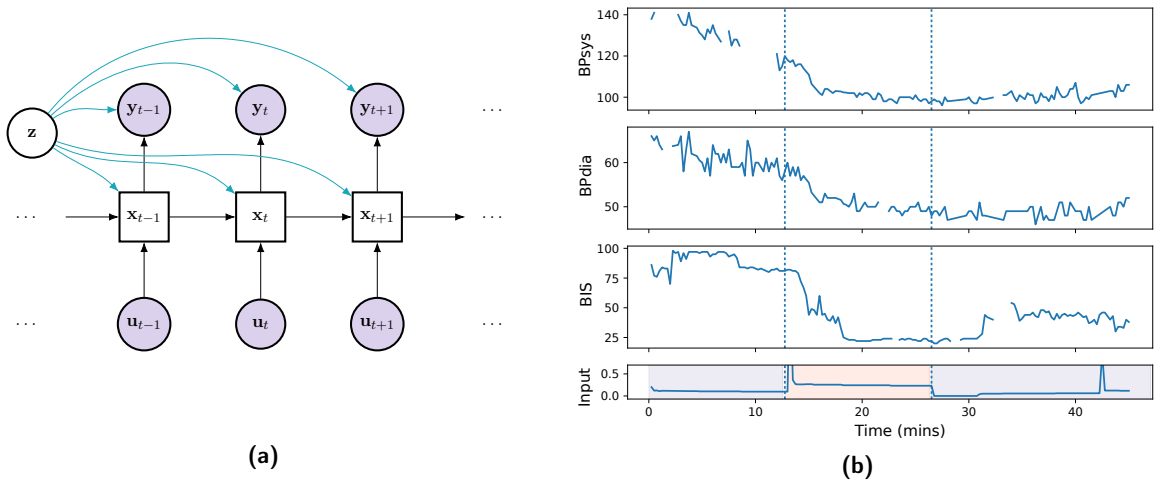

We assume that each patient $i$ can be described by the above PD model with parameter $\bftheta^\ii$. The parameters for patient $i$ (denoted with the associated superscript) will be modelled by use of a latent code $\z^\ii \in \R^k$ with prior $\z^\ii \,\sim\, \Normal{0, I_k}$ which relates to the parameters as:
\begin{align}
    \bftheta^\ii \,=\, [\mbf\alpha^\ii, \bfbeta_1^\ii, \bfbeta_2^\ii, &\,\bfbeta_3^\ii, \bfeta_1^\ii, \ldots, \bfeta_{n_y}^\ii] \,=\, \hphi(\z^\ii). \label{eq:mtpd:model:multitask:z}
\end{align}
See Figure \ref{fig:mtpd:gm} for a graphical model. Here we choose $\hphi(\z) \,=\, \mbf f(\Phi\mkern1.5mu \z + \mbf c)$ for some `loading matrix' $\Phi$, offset $\mbf c$ and elementwise transformation function $\mbf f$ (see below). This construction extends the basic PD model in  \eqref{eq:mtpd:model:singletask} to a family of PD models, where each patient has a personalized parameter vector $\bftheta^\ii$.
The improvements over the ST approach are facilitated by the rank-$k$ $\Phi$, and the `default' parameter $\bftheta^{(0)} \defeq \E_{p(\z)}[\hphi(\z)]$ learned from the training data. We call this an `MTPD' model.

%


The function $\mbf f$ consists of elementwise univariate transformations which ensure that each parameter satisfies the required constraints. For example, the unit interval constraints for $\bfbeta_1^\ii$ is enforced via a logistic sigmoid, and the non-negativity constraints for $\bfbeta_2^\ii$ by $\text{softplus}(x) = \log(1 + e^x)$ etc. If parameters are unconstrained, no nonlinearity is applied. The use of an (elementwise constrained) affine $\hphi$ can result in an interpretable latent code for clinical practice; the meaning of each element of $\z$ can easily be obtained via inspection of the matrix $\Phi \in \R^{d\times k}$.

We have formulated this model for an unknown $\mbf z$ which is inferred over time. However, some information may be gleaned from covariates $\mbf \zeta$ such as age, height, weight etc. To the extent that these covariates `describe' the differences between patient responses, we can set $\z \leftarrow \mbf \zeta$ which we call a \emph{task-descriptor} approach. In this case, test time inference is not required, but the model cannot adapt to the response. A hybrid approach is also possible, which performs inference only on a subset of the dimensions of $\z$.

\subsubsection{Related Work}  \label{sec:mtpd:related}

We pause here to briefly review related work in the machine learning literature
concerning personalized treatment modelling.
Deep learning methods are not considered, due to the small sample sizes encountered in clinical trials. \citet{georgatzis2016ionlds} demonstrate that relaxing the PK/PD model class to a general state space model can result in improved model fit and in-sample prediction, but such models cannot be applied to new patients at test time. Multi-task GPs (MTGPs) have been used for condition monitoring, for example in \citet{durichen2015multitask}, but for gaining strength over multiple observation channels. The approach cannot easily gain strength over different patients (see MTGP discussion in \secref{sec:MTDS:related}) and further, cannot integrate control inputs. \citet{alaa2018personalized} permit gaining strength over patients via use of a mixture of GPs. But in their work, personalization is achieved via use of covariates, restricted to a fixed set of subtypes, and is still unable to integrate control inputs.

\citet{schulam2015framework} propose a form of generalized linear mixed effects model, which assumes an \emph{additive} decomposition of population, individual and (GP-based) noise components. Further extensions are proposed in \citet{xu2016bayesian, futoma2016predicting},  but these approaches only adjust linear coefficients, and cannot customize dynamical parameters. These approaches are extended further by \citet{soleimani2017treatment, cheng2020patient}, who include first order ODEs of control inputs. Nevertheless, these still relate \emph{additively} to the observations (exploiting the linearity of GPs);  extensions to nonlinear dynamical systems (e.g.\ PK/PD models) are not straight-forward. Furthermore, no methods for online inference are presented, and no multi-task ideas are used; adaptivity is restricted to simple mixed effects.

\subsection{Experimental Setup}  \label{sec:mtpd:experimental}

This section describes the experimental setup for the evaluation of our model. Section \ref{sec:mtpd:experimental:data} describes the data, \secref{sec:mtpd:experimental:design} describes the form of evaluation, and the details of the models under comparison are given in \secref{sec:mtpd:experimental:models} -  \secref{sec:mtpd:experimental:models:lstm}.

\subsubsection{Data} \label{sec:mtpd:experimental:data}

The data were obtained from an anaesthesia study carried out at the Golden Jubilee National Hospital in Glasgow, Scotland, as described in \citet{georgatzis2016ionlds}. These consist of $N=40$ time series of Caucasian patients; the median length is 36 minutes (range approx.\ 27 - 50 mins) and the data are subsampled to 15-second intervals. Each patient was assigned to one of two pre-operative infusion schedules of propofol following a high-low-high or low-high-low sequence (see Appendix \ref{sec:MTPD-supp:data}). Each patient has additional covariates of age, gender, height, weight (and body mass index). The observations $\y_t$ at each time $t$ have $n_y=3$ channels comprising systolic and diastolic blood pressure (BPsys, BPdia) and BIS, however 13 patients have no BIS channel, which is considered as missing data.  An example time series is shown in Figure \ref{fig:mtpd:example_vitals}: the observations $\{\y_t\}$ are shown in the top panels, and the raw drug infusion input at the bottom. The infusion schedule is indicated via the vertical dotted lines with the middle section targeting a higher propofol concentration. One can observe many of the discussed features here including lagged and nonlinear responses. The missing values are due to sensor dropouts and other noise which was removed with clinical supervision.

\subsubsection{Evaluation} \label{sec:mtpd:experimental:design}

We evaluate predictions from our MTPD model by Root Mean Squared Error (RMSE) over 20- and 40-step ahead windows (a clinically relevant interval of 5 or 10 mins). The performance is reported at both 12 minutes and 24 minutes to understand if the predictions are improving over time. We follow a leave-one-out (LOO) procedure due to the relatively small number of patients (in machine learning terms); for each of 40 folds, a model is learned on 39 patients and tested on the held-out patient, and the results are averaged. During training, the RMSE is weighted such that each patient has an equal contribution to the objective despite differing sequence lengths, to avoid a bias towards patients with longer sequences.

\begin{table}[]
    \centering
    \begin{tabular}{p{2.35cm} p{4.5cm} p{1.2cm} p{5.5cm}}
        \toprule
        Name & Parameters & \centering Adapt-\ \newline ive $\mbf \alpha$ & Details \\ \midrule
        Pooled & $\bftheta = \bftheta_0$ & \xmark & \small One-size-fits-all model. \\
        Pooled-$\alpha$ & $\bftheta = \bftheta_0$ & \cmark &\small As above, but with adaptive $\mbf \alpha$. \\
        Task-ID & $\bftheta = \hphi(\mbf \zeta)$ & \xmark & \small Customized using patient covariates. \\
        Task-ID-$\alpha$ &  $\bftheta = \hphi(\mbf \zeta)$ & \cmark & \small As above, but with adaptive $\mbf \alpha$. \\
        MTPD-$k$ & $\bftheta = \hphi(\z)$, \hspace*{4pt} $\z\sim\Normal{0, I_k}$ & \cmark & \small  $k=5,7$ chosen using the ELBO. \\
        Single Task & $\bftheta \sim \Normal{0, 100^2 I_{33}}$ & \cmark & \small (Relatively) uninformative Gaussian prior on all dimensions of $\bftheta$. \\
        \bottomrule
    \end{tabular}
    \caption{Versions of the pharmacodynamic model considered in our experiments.}
    \label{tbl:mtpd:models}
\end{table}

\subsubsection{PD model variants} \label{sec:mtpd:experimental:models}

We consider a number of variants of the PD model described in \secref{sec:mtpd:model:singletask}. See Table \ref{tbl:mtpd:models} for an overview. 

\paragraph{Pooled Model and Task-ID Model:} The most basic benchmarks are a one-size-fits-all Pooled model, and a task-descriptor (`Task-ID') version. The Task-ID model adapts $\bftheta$ from known covariates or `task-descriptors' of the patients ($\mbf \zeta$) via $\hphi$; the use of patient covariates resulted in practice in a small improvement on the training set, but regularization of $\bfphi$ was essential to to avoid poor performance on the validation set.\footnote{Regularization hyperparameters were tuned on a validation set of size $N=4$.}  These models use a single set of parameters estimated from the training set, and perform no online adaptation. This provides a proxy for state-of-the-art models such as those in \citet{jeleazcov2015pharmacodynamic, eleveld2018pharmacokinetic}. 

\paragraph{Adaptive-offset Models:} We also provide improvements of these benchmarks which adapt the `offset' or `level' $\mbf \alpha$ online, denoted `Pooled-$\alpha$' and `Task-ID-$\alpha$' respectively. Online inference of the level $\mbf \alpha$ is not usually considered in the literature, but it provides a helpful comparison point due to the inter-patient variance of this parameter in fitting real-world data (see Figure \ref{fig:mtpd:possible-responses}, page \pageref{fig:mtpd:possible-responses}). The prior for the adaptive offsets $\mbf \alpha$ for the Pooled-$\alpha$ and Task-ID-$\alpha$ models was obtained by fitting a Gaussian to the learned offsets $\{\mbf \alpha^\ii\}_{i=1}^N$ in the training set (we fit a per-channel Gaussian). At test time, we use sequential inference via exact Bayesian updating using standard Gaussian formulae.

\begin{figure}
\begin{floatrow}
\ffigbox[\FBwidth]{%
  \includegraphics[scale=0.48]{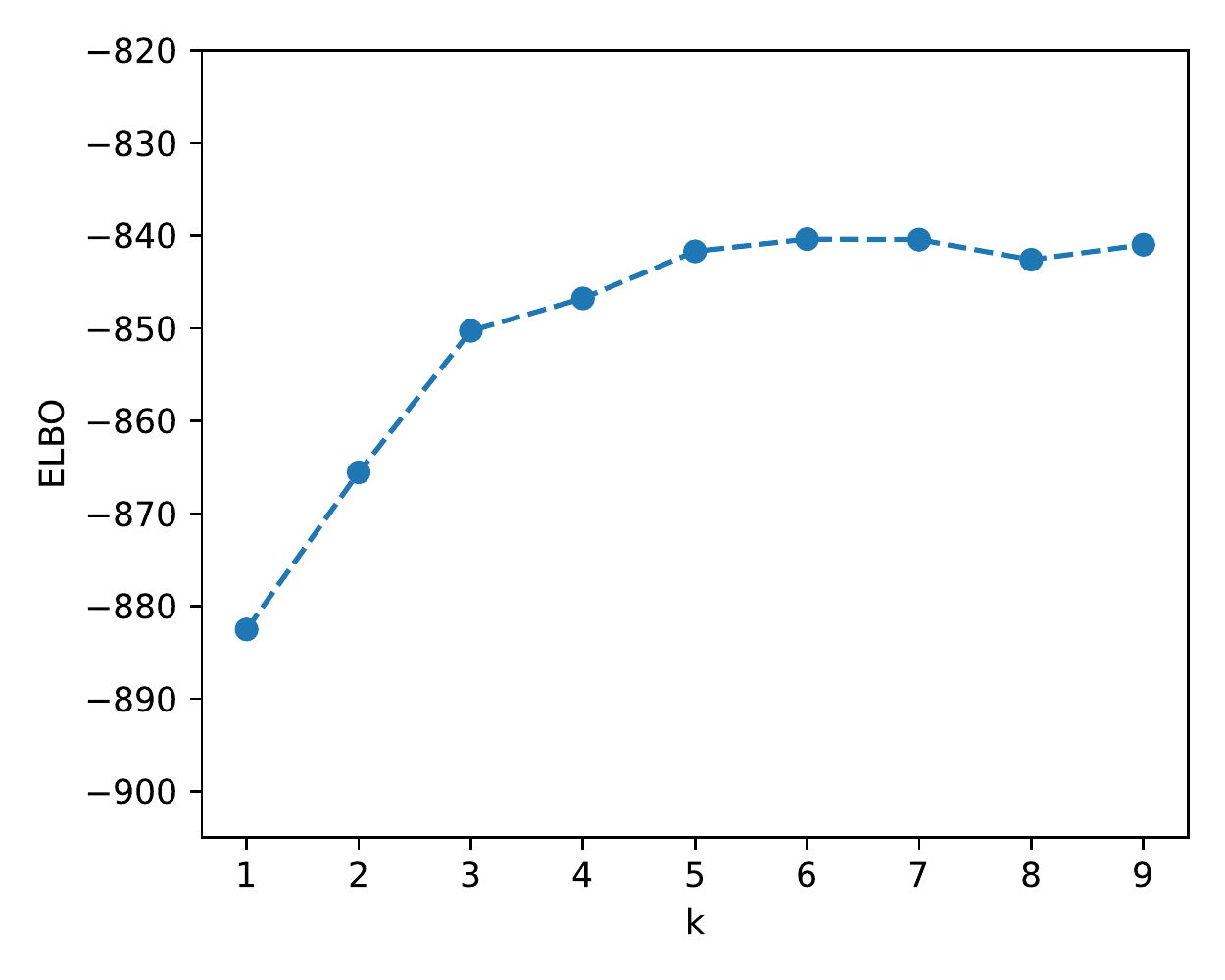}
  \vspace*{-8pt}
}{%
  \caption{Average (per patient) ELBO of MTPD model fitted on entire dataset for varying latent dimension $k$}%
  \label{fig:mtpd:elbo}
} \enskip
\capbtabbox{%
  \scriptsize
  \setlength{\tabcolsep}{3pt}
  \begin{tabular}{lrrrrrr}
\toprule
{} && \multicolumn{2}{c}{$t=12$ m} && \multicolumn{2}{c}{$t=24$ m} \\
\cmidrule(r){3-4}\cmidrule(r){6-7}
  {} && RMSE &  RMSE &&  RMSE &  RMSE \\
  {} && 20-ahead &  40-ahead && 20-ahead & 40-ahead  \\
\cmidrule(r){1-1}\cmidrule(r){3-4}\cmidrule(r){6-7}
BPsys && 5.40 & 5.31 && 5.43 & 5.19 \\
BPdia && 3.63 & 3.79 && 3.55 & 3.75 \\
BIS   && 7.42 & 7.35 && 7.68 & 7.81 \\
\bottomrule
\end{tabular}
  \vspace*{30pt}
}{%
  \caption{Minimum possible predictive RMSE of PD model class. 20-, 40-step errors calculated \emph{in-sample} for \emph{per-patient} models after $t=12,24$.}%
  \label{tbl:mtpd:pd-opt}
}
\end{floatrow}
\end{figure}

\paragraph{MTPD Model:} The MTPD model is implemented according to eqs.\ (\ref{eq:mtpd:model:singletask}) and (\ref{eq:mtpd:model:multitask:z}) in \secref{sec:mtpd:model}. Figure \ref{fig:mtpd:elbo} shows the average per-patient ELBO (see equation \ref{eq:mtds:elbo}) of an MTPD model fitted on the entire dataset for latent dimensions $k=1,\ldots,9$. This motivates the choice of latent dimensions $k=5$ and $k=7$;\footnote{$k=7$ is highest value of the ELBO, and $k=5$ is chosen as a more pragmatic trade-off between dimensionality and the ELBO.} the respective models will be referred to as MTPD-5 and MTPD-7.\footnote{The degrees of freedom of $\mbf \alpha$ are not included in $k$, which is adapted separately from the MT parameters in order to compare all models  like-for-like.} Learning and inference largely follow \citet{bird19mtpd}, although that paper used a MAP approximation to the marginal likelihood (eq.\ \ref{eq:MLE}); we obtain similar results with the variational approach discussed in \secref{sec:MTDS:learning}. Furthermore, we use the AdaIS method (\secref{sec:MTDS:inference:our}, with parameters $J=4, N_{\AdaIS}=5$) to perform inference of the latent $\z$ for each patient, which proved to be two orders of magnitude faster than the Hamiltonian Monte Carlo (HMC) approach of \citet{bird19mtpd}.

\paragraph{Single Task Model:} The single-task version of the PD model is the most flexible variant and requires no offline learning stage. Instead, a relatively uninformative prior is placed on each parameter (see \citeay{bird19mtpd}; parameters are constrained to their support via sigmoidal or softplus transformations where relevant, cf.\ \secref{sec:mtpd:model:multitask}). Due to the high-dimensional and poorly conditioned nature of the posteriors here, we could not use the AdaIS method of \secref{sec:MTDS:inference:our} and instead use HMC as per \citeauthor{bird19mtpd}. Even with the mature library \texttt{Stan} \citep{carpenter2016stan} we had to perform offline work to estimate the mass matrix of the sampler in order to avoid unstable chain dynamics and numerical problems.

\subsubsection{LSTM Benchmark}  \label{sec:mtpd:experimental:models:lstm}

It is unlikely that more complex/`neural' models such as RNNs will be accepted by practicing anaesthetists in the near future for a variety of reasons. The sample complexity of a RNN is poorly matched to the typical sample size of a clinical trial, predictions may perform very poorly under dataset shift (see \secref{sec:mocap:expmt2}), and the model is inscrutable, which precludes both an understanding of the prediction, and the ability to alter it. Nevertheless, it is still useful to provide a `neural' benchmark to help us understand the opportunity cost of using simpler models.  Note that if black box models are permissible, we might also expect improvements to RNNs using an MTDS approach (as in \secref{sec:mocap}). 

For the benchmark, we use a one-layer LSTM, with a hidden layer size of 32 and L2 regularization coefficient $10^{-3}$ chosen by grid search from $\{16, 32, 128\}$ $\times\, \{10^{-5}, 10^{-4},$ $10^{-3}, 10^{-2}, 10^{-1}\}$, and fitted via use of the Adam optimizer. As in \secref{sec:mocap} we train the model in an open-loop (or seq2seq) fashion, encoding 40 timesteps of inputs and outputs prior to a 40-step prediction. In each training iteration, the starting time $t$ is randomized. Observations with missing values required transformation for the `encoder' section of the LSTM: these were handled by zero-imputation, concatenated with a one-hot encoding of the pattern of missingness. 


\subsection{Results}
\label{sec:mtpd:results}

This section provides the experimental results, with an introduction in \secref{sec:mtpd:results:context}, and the leave-one-out results presented in \secref{sec:mtpd:results:all}.

\begin{figure}[t!]
    \centering
    \hspace*{10pt}
    \begin{tikzpicture}
        \node[anchor=south west,inner sep=0] (image) at (0,0) {\includegraphics[trim={0 0 160 0}, clip, width=0.82\textwidth]{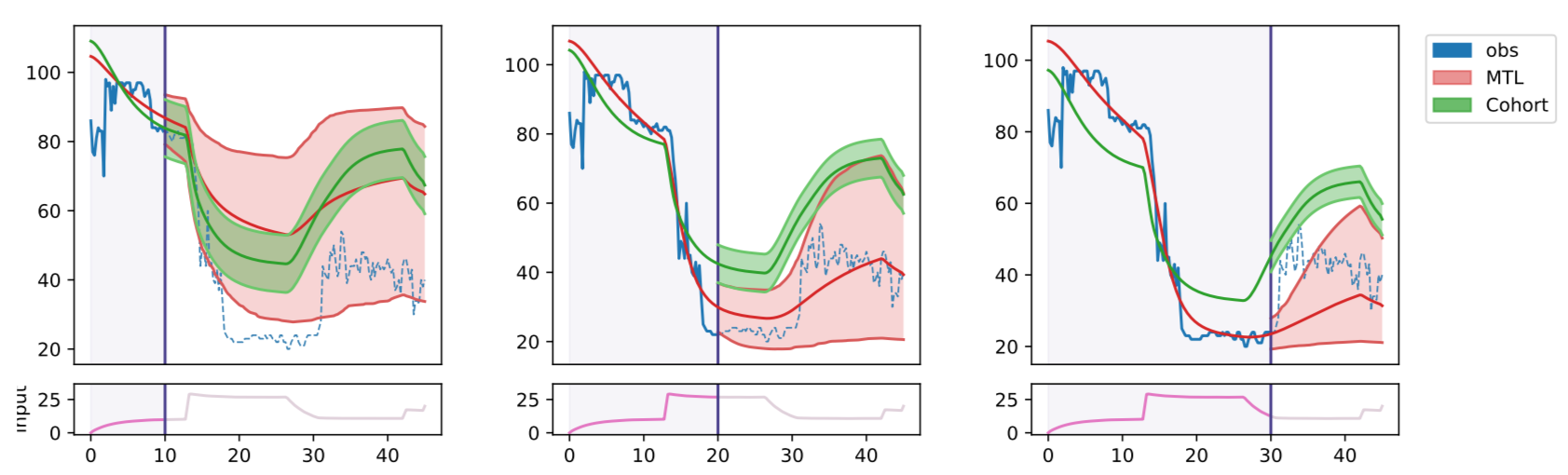}};
        \begin{scope}[x={(image.south east)},y={(image.north west)}]
            \node (leg1) at (1.07,0.82) {\includegraphics[trim={0 0 0 0}, clip, height=36pt]{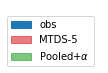}};
        \end{scope}
    \end{tikzpicture}
    \caption{Example predictions (mean and 90\% CI) for BIS channel of a patient at $t=10, 20, 30$ minutes using Pooled-$\alpha$ and MTPD-5 models. The PK central compartment concentration is shown in the bottom panel. Retrospective fits are shown without intervals for clarity.}
    \label{fig:mtpd:BIS_example}
\end{figure}


\subsubsection{Introduction}   \label{sec:mtpd:results:context}
Example predictions from both the MTPD-5 and Pooled-$\alpha$ models can be seen in Figure \ref{fig:mtpd:BIS_example} for the BIS channel. One can see the models adapting over time at $t=10,20,30$ minutes, where the credible intervals show the predictive posterior for the \emph{underlying PD function}. While the adaptive Pooled-$\alpha$ model (green) is fixed in shape and only updates its offset, the MTDS approach permits much greater flexibility, providing continual adaption over increasing $t$.  Further examples can be seen in \citet[][supplementary material]{bird19mtpd}.

Table \ref{tbl:mtpd:pd-opt} provides a lower bound of the error that can be achieved by the PD model class of \secref{sec:mtpd:model:singletask} for the data in \secref{sec:mtpd:experimental:data} via its \emph{in-sample} error. To estimate this, we train a single-task model for each patient $i$ on the complete sequence $t=1,\ldots, T_i$ and report its in-sample error for 20-step and 40-step ahead predictions at $t=12,24$ as per \secref{sec:mtpd:experimental}. The RMSE of the BIS channel is relatively high, in part due to noise processes, but also since the PD model class is insufficiently flexible for this channel.\footnote{BIS is known not to follow PD assumptions: it is a composition of many signals and suffers from the difficulty of defining consciousness as a scalar value; see e.g.\ \citet{lobo2011limitationsBis, schuller2015response}.} For an example of BIS violating the PD model assumptions, see the step-change in the BIS channel observations at $t=30$ in Figure \ref{fig:mtpd:BIS_example} which  may perhaps correspond to a phase change in patient state (see e.g.\  \citealt{mukamel2014transition}) or some form of hysteresis.

\subsubsection{Test set Results}   \label{sec:mtpd:results:all}

The \emph{out-of-sample} performance (LOO average) relative to the optimal performance (per the results in Table \ref{tbl:mtpd:pd-opt}) is shown in Table \ref{tbl:mtpd:diff} for all models. This is reported as Standardized RMSE (SRMSE), which is a ratio of the model RMSE to the optimal fit in Table \ref{tbl:mtpd:pd-opt}. A value of $1.00$ indicates the same level of performance as the optimal PD fit, and a higher value indicates a worse fit. Results for the three channels are listed separately. One can see that the non-adaptive approaches often proposed in the literature (both Pooled and Task-ID) are highly suboptimal; for the blood pressure channels, the RMSE is more than twice that of the optimal fit.

\begin{table}
    \centering
    \small
    \begin{tabular}{llrrrrrr}
\toprule
{} & {} && \multicolumn{2}{c}{$t=12$ m} && \multicolumn{2}{c}{$t=24$ m} \\
\cmidrule(r){4-5}\cmidrule(r){7-8}
{} &  {} && SRMSE &  SRMSE &&  SRMSE &  SRMSE \\
   & {} && 20-ahead &  40-ahead && 20-ahead & 40-ahead  \\
\cmidrule(r){1-2}\cmidrule(r){4-5}\cmidrule(r){7-8} 
      & Pooled   &&     2.86 &     2.65 &&     3.25 &     3.11 \\
      & Task-ID  &&     2.84 &     2.64 &&     3.23 &     3.10 \\
\arrayrulecolor{lightgray}\cmidrule(r){2-2}\cmidrule(r){4-5}\cmidrule(r){7-8}
  & \phantom{Offsets Task-ID} && & && & \\[-\normalbaselineskip] 
      & Pooled+$\alpha$           &&     1.23 &     1.31 &&     1.37 &     1.41 \\
BPsys & Task-ID+$\alpha$          &&     1.26 &     1.37 &&     1.55 &     1.67 \\
      & STL              &&     2.10 &     2.29 &&     1.29 &     1.60 \\
      & MTDS-5           && \bf 1.18 & \bf 1.30 && \bf 1.00 &     1.15 \\
      & MTDS-7           &&     1.19 &     1.31 && \bf 1.00 & \bf 1.12 \\
\cmidrule(r){2-2}\cmidrule(r){4-5}\cmidrule(r){7-8}\arrayrulecolor{black}
      & LSTM (open loop) &&     1.59 &     1.76 &&     1.46 &     1.48 \\
\cmidrule(r){1-2}\cmidrule(r){4-5}\cmidrule(r){7-8} 
      & Pooled   &&     2.42 &     2.37 &&     2.59 &     2.35 \\
      & Task-ID  &&     2.40 &     2.35 &&     2.57 &     2.34 \\
\arrayrulecolor{lightgray}\cmidrule(r){2-2}\cmidrule(r){4-5}\cmidrule(r){7-8}
      & Pooled+$\alpha$           && \bf 1.03 & \bf 1.13 &&     1.21 &     1.15 \\
BPdia & Task-ID+$\alpha$          &&     1.16 &     1.30 &&     1.23 &     1.24 \\
      & STL              &&     1.50 &     1.74 &&     1.57 &     1.58 \\
      & MTDS-5           &&     1.12 &     1.24 && \bf 1.01 &     1.03 \\
      & MTDS-7           &&     1.15 &     1.25 &&     1.02 & \bf 0.99 \\
\cmidrule(r){2-2}\cmidrule(r){4-5}\cmidrule(r){7-8}\arrayrulecolor{black}
      & LSTM (open loop) &&     1.61 &     1.80 &&     1.26 &     1.41 \\
\cmidrule(r){1-2}\cmidrule(r){4-5}\cmidrule(r){7-8} 
      & Pooled   &&     1.66 &     1.87 &&     2.08 &     1.82 \\
      & Task-ID  &&     1.69 &     1.90 &&     2.11 &     1.84 \\
\arrayrulecolor{lightgray}\cmidrule(r){2-2}\cmidrule(r){4-5}\cmidrule(r){7-8}\arrayrulecolor{black}
      & Pooled+$\alpha$           &&     1.45 & \bf 1.67 &&     1.47 &     1.49 \\
BIS   & Task-ID+$\alpha$          &&     1.66 &     2.01 &&     1.65 &     1.69 \\
      & STL              &&     1.97 &     3.16 &&     1.31 &     1.61 \\
      & MTDS-5           &&     1.35 &     1.81 &&     1.21 &     1.29 \\
      & MTDS-7           && \bf 1.32 &     1.85 && \bf 1.19 & \bf 1.28 \\
\arrayrulecolor{lightgray}\cmidrule(r){2-2}\cmidrule(r){4-5}\cmidrule(r){7-8}\arrayrulecolor{black}
      & LSTM (open loop) &&     1.19 &     1.55 &&     0.88 &     1.28 \\
      \bottomrule
\end{tabular}
    \vspace{5pt}
    \caption{The out-of-sample performance of each model standardized by the optimal PD results in Table \ref{tbl:mtpd:pd-opt} (lower is better). For each channel, models are split into standard PD approaches, adaptive PD approaches, and non-PD approaches. As per Table \ref{tbl:mtpd:pd-opt}, the results give the 20 and 40-step predictive RMSE after $t=12,24$. The results highlighted in bold are the best performance from a PD model.}
    \label{tbl:mtpd:diff}
\end{table}


By $t=24$ there is a clear win for the MTDS models over all other PD approaches. The 20-step performance is essentially optimal for the BPsys and BPdia channels, and further, the performance is substantially better than the flexible black-box LSTM model. For BIS, an optimal performance is not achieved by any approach within the class of PD models, but the MTDS results are a promising step forward (for the LSTM results, see below). 
The ST approach performs relatively poorly for all channels, and inference takes 1-2 orders of magnitude longer than other approaches. Nevertheless, its performance is expected to improve with increasing $t$. The adaptive Pooled-$\alpha$ model performs better initially, but it is not flexible enough to provide much customization, and shows no improvement over time. The performance of the patient covariate model Task-ID (both adaptive and non) is similar to or worse than the Pooled version. While  Task-ID models clearly suffer from overfitting, the in-sample improvement is fairly small, suggesting the available patient covariates may lack the required information to attain meaningful improvement. Note that patient covariates have already been used in the PK component, and are hence incorporated in the inputs.



There appears to be some performance advantage for using the LSTM model for the BIS channel (although the MTDS is able to reduce the gap by $t=24$). This may be due to the particularly serious mis-specification of the PD model for BIS, as mentioned above. The SRMSE of 0.88 for the BIS prediction at $t=24$ indicates that the LSTM can fit the data better than the \emph{in-sample} PD model. In particular, the 1-d latent linear dynamics of the PD model cannot fit the stepped level changes and hysteresis sometimes observed. The latter results in correlated noise to the fit, which further misleads online inference for the MTDS. For further experimental results and discussion see \citet{bird21thesis}, \S6.

\subsection{Conclusion}
\label{sec:mtpd:conclusion}
The application of the MTDS framework to PK/PD modelling problem provides a novel approach to personalized medicine, which (for our propofol data set) shows substantial promise over traditional approaches using patient covariates. For a novel patient, our approach initializes with a close approximation to the current state-of-the-art, but can achieve increasing personalization over time, without deviating from the model class accepted by clinical practitioners. This is performed via an efficient online Bayesian filtering approach, which reduces the risk of overfitting compared to a point estimate or amortized approach.

The experiments have highlighted a number of areas for further improvement. Firstly, the BIS channel may benefit from a more flexible PD model class. This can be seen from the relatively poor performance of optimal PD fits, visual inspection, and comparison with LSTM models.  Secondly, model predictions can likely be improved via better experimental design of the infusion sequence. Finally, incorporating artefact models \citep[see e.g.][]{quinn2009factorial} should reduce the sensitivity of the inference to correlated noise and artefacts, reducing suboptimal predictions.

\section{Conclusion}


This paper has defined a new class of time series model: the multi-task dynamical system, together with methods of learning and inference. We have demonstrated the efficacy of these via two detailed experiments. Our results show that the MTDS can indeed learn an embedding of sequence characteristics, resulting in a family of dynamical systems which span the inter-sequence variation of the training set. This latent encoding can be used to personalize forecasts---leading to increased accuracy---or to modulate their characteristics at will, depending on the goals of an end user. 

We believe that datasets containing multiple sequences often manifest the inter-sequence variation discussed in this paper. This extends to natural language or video data; other examples might often be found in a business context, albeit rarely in the public domain. If this belief is correct, predicting time series data accurately must require task inference (in the sense described here), unless one can use a separate model for each time series. For task inference, the only general-purpose alternatives to the MTDS are neural models (such as RNNs), which perform task inference implicitly. In contrast, our MTDS approach performs this \emph{explicitly} and encompasses a number of existing models in the literature. We have shown a number of advantages of explicit inference in our experiments including  interpretability, data efficiency, robustness to dataset shift, end-user control, and visualization.

Future work may consider extending the MTDS to handle non-stationary data explicitly via a time-varying dynamic task variable $\z_t$. Also, while the MTDS appears to interpolate well between tasks, it may not be able to extrapolate so well, as evidenced in Section \ref{sec:mocap:expmt2}. This can be handled by retraining, but it may be advantageous to consider extensions that can adapt faster to novel tasks.


\acks{The authors would like to thank Stefan Schraag, Shiona McKelvie, Mani
Chandra and Nick Sutcliffe for the original study design and data
collection for the drug response data. They also thank Ian Mason for
making available the mocap data, as well as a helpful discussion.  This
work was supported by The Alan Turing Institute under the EPSRC grant
EP/N510129/1.}


\appendix
\section{Further Details of General MTDS Models}
\label{appdx:intro}

In this section we provide further details about the construction of MTDS models.

\subsection{Choice of Prior for Large $d$} \label{appdx:mtds:prior}
 
For modern `neural' applications, the dimension $d$ of the output space $\Theta = \R^d$, corresponding to the system and emission parameters, may be very large, e.g.\ $\gO(10^6)$ for RNNs. The parameter $\bfphi$ will therefore be even larger; in the case of an affine $\hphi$, the parameter will have $(k{+}1){\times}d$ dimensions, and for an MLP, $\bfphi$ could be an order of magnitude larger even than this. Practical choices of prior will restrict the final layer of such an MLP to be relatively small, reducing the flexibility of this nonlinear approach. Nevertheless, use of an MLP may still be advantageous since, for example:
\begin{enumerate}[label=(\roman*)]
    \item The MLP can result in non-Gaussian densities in parameter space, even if the manifold of the support is relatively simple.
    \item A linear space of recurrent model parameters can yield highly non-linear changes even to simple dynamical systems via bifurcation \citep[see e.g.][\S 8]{strogatz2018nonlinear}. We speculate it might be advantageous to curve the manifold to avoid such phenomena.
    \item More expressive choices may help utilization of the latent space \citep[see e.g.][]{chen2016variational}. 
\end{enumerate}
Nonetheless, it may often be reasonable to use a linear factor analysis model (i.e.\ $\hphi$ is affine) when $\bftheta$ is large. Empirically we have observed higher marginal likelihoods for smaller state space models when using a nonlinear manifold, but affine and nonlinear manifolds may work equally well for larger RNN models.

\subsection{Use of Deterministic State Models} \label{appdx:mtds:deterministic}

In this paper, we restrict our focus to deterministic state dynamical systems because the applications do not demand stochastic state models, and long-term predictions appear to benefit from the deterministic state. Some intuition of the latter can be found in \citet{martinez2017human} where use of a deterministic state forces the models to learn to `recover' from their mistakes. Similar observations can be found in \citet{bengio2015scheduled} or \citet{chiappa2017recurrent} for example.

We also propose that the choice of deterministic state models is less limiting than it may first appear. Consider a simple linear dynamical system with no inputs:
\begin{alignat}{3}
&& \x_t \,&=\,A\,\x_{t-1} + \bv_t, \qquad &&\bv_t \sim \Normal{\mbf 0, R}, \label{eq:ts:lds:system}\\
&& \y_t \,&=\,C\,\x_{t} + \bw_t, \qquad &&\bw_t \sim \Normal{\mbf 0, S}. \label{eq:ts:lds:emission}
\end{alignat}
Now consider a Kalman filter initialized from $p(\x_{t-1} | \x_{1:t-2}) = \mathcal{N}(\x_{t-1} \,|\, \mbf m_{t-1}, \, P_{t-1})$ which infers the latent variables recursively \citep[following][]{sarkka2013bayesian} via the following steps:
\begin{align*}
 \mbf m_t^- &\,=\, A \mbf m_{t-1}, \\
 P_t^- &\,=\, A P_{t-1}A\Tr + R. \\
  K_t &\,=\, P_t^- C\Tr (C P_t^- C\Tr + S)^{-1}. \\
 \mbf m_t &\,=\, \mbf m_t^- + K_t (\y_t - C \mbf m_t^-),  \\
 P_t &\,=\, P_t^- - K_t C P_t^-.
\end{align*}
The conditional likelihood of the observations is $p(\y_t \,|\,\y_{1:t-1}) \,=\, \mathcal{N}(C \mbf m_t^- \,|\, C P_t^- C\Tr + S)$, obtained from \eqref{eq:ts:lds:emission}. The covariance $P_t$ typically converges quickly to a fixed point \citep[][ch.\ 24]{barber2012bayesian}, in which case the quantities $P_t, K_t$ converge to time-independent quantities $P, K$ \citep[for more details see][\S3.3.3]{harvey1990forecasting}. Therefore at steady state the one-state conditional likelihoods can be described by the following system:
\begin{align}
    \mbf m_t &\,=\, A \mbf  m_{t-1} + K (\y_t - C A \mbf m_{t-1})  \label{eq:ts:kf:pred:deterministicstate}\\
    \y_{t+1}\,|\, \y_{1:t} &\,\sim\, \Normal{C A \mbf m_{t},\, C (A PA\Tr + R) C\Tr + S}, \label{eq:ts:kf:pred:deterministicemission}
\end{align}
which is the form of a deterministic-state LDS. Hence for every stochastic-state LDS at steady state, there exists a \emph{deterministic-state} LDS with the same distribution over $\y_{1:T}$. Notably however, the latter interprets the preceding observations $\y_{1:t-1}$ (at time $t$) as inputs, whereas the former may have no control inputs. While the steady-state assumption cannot always be made, this idea suggests that learning a deterministic LDS using observations $\y_{1:T-1}$ as inputs can approximate the likelihood of a stochastic model. Deterministic RNNs use a similar idea (sometimes called `teacher forcing') which enables them to model notionally stochastic time series with high accuracy. 

If one does wish to learn a stochastic state MTDS, the latent dynamical state must be integrated out for learning and inference over $\z$. This is a more difficult inference problem. Nevertheless, it is relatively easy to extend the MTDS to stochastic-state LDS models as this can be performed in closed form (as above), and the benefit of this closed-form inference may be further extended in principle to nonlinear models too as in e.g.\ \citet{karl2016dvbfilter}.

\subsection{Implementations} \label{appdx:mtds:code}
An example implementation of the MTDS in PyTorch is provided at \url{https://github.com/ornithos/pytorch-mtds-mocap}, together with the data for the Mocap experiments. At the time of writing, the pharmacodynamic data is not yet publicly available, but a reference implementation of the model is available at \url{https://gist.github.com/ornithos/71abb7349e91db633ce15971785bbae1}.

A Julia implementation of AdaIS for performing sequential inference in models with static latent variables (such as the MTDS) can be found at \url{https://github.com/ornithos/SeqAdaptiveIS}.

\section{Application to Mocap Data} \label{appdx:mocap}
This section provides additional details pertaining to the mocap application in \secref{sec:mocap}.

\subsection{Model Inputs} \label{sec:mocap:data:inputs}

Our choice of inputs reflects controls that an animator may wish to manipulate. The first input is the trajectory that the skeleton is to follow. As in \citet{holden2017phase}, we provide the trajectory over the next second (30 frames), sampled every 5 frames. Unlike previous work, the trajectory history of the prior 30 frames is omitted, since it can be retained in the dynamic state of the model. The (2-d) trajectory co-ordinates are given with respect to the Lagrangian co-ordinate frame, and hence can vary rapidly when the skeleton turns quickly. We choose to provide an Eulerian representation of the trajectory too, which sometimes resulted in a smoother prediction.

The velocity implied by the trajectory does not disambiguate the gait frequency vs.\ stride length. The same velocity may be achieved with fast short steps, or slower long strides. We therefore provide the gait frequency via a phasor \citep[as in][]{holden2017phase}. This is provided by sine and cosine components to avoid the discontinuity at $2\pi$. This may also be useful to an animator. A final ambiguity exists from the trajectory at tight corners: the skeleton can rotate either towards the focus of the corner, or in the opposite direction. Figure \ref{fig:mocap:intro:reverse} demonstrates the latter, which is not infrequently performed in the data. We provide a boolean indicator for each of the 6 sampled trajectory timesteps, indicating corners for which this happens. 

Altogether we have $\bu_t \in \R^{35}$: 12 inputs each for the Lagrangian trajectory and the differenced Eulerian trajectory, 2 inputs for the gait phase, 6 inputs for the turning indicators and 3 additional inputs of the instantaneous velocity\footnote{We include 3 inputs consisting of the first difference of the trajectory evaluated at the current position, and the current rotational velocity. While this can in theory be derived by the model from the preceding inputs, we found its inclusion resulted in a smoother prediction.}. These inputs $\{\bu_t\}$ are standardized to have zero mean and unit variance. A particular problem in producing the inputs was the stylistic swaying/movement of the root joint which leaked information about the style in the trajectory and prevented customization of the pelvic movement. This was removed during pre-processing, for more on which see \citet{bird21thesis}.

\subsection{Recurrent Cell Definitions} \label{appdx:mocap:recurrentcells}
Equations (\ref{eq:mocap-gru}) and (\ref{eq:mocap-mtrnn}) refer to a $\textrm{GRUCell}$ and a $\textrm{RNNCell}$ which we define below.
\begin{align}
    \textrm{\small RNNCell}_{n_x}\left(\x_{t-1},\, \bu_t;\,\,\, \bfpsi \right) \,:&=\, \tanh\left(A \x_{t-1} + B \bu_t + \mbf b\right), \\
    \intertext{where in this case $\bfpsi$ comprises the parameters $A \in \R^{n_x\times n_x}, B \in \R^{n_x\times n_u}, \mbf b \in \R^{n_x}$. The GRU cell can be written via the following four equations:}
    \textrm{\small GRUCell}_{n_x}\left(\x_{t-1},\, \bu_t;\,\,\, \bfpsi \right) \,:&=\, (1 - \mbf g^s_t) \odot \x_{t-1} + \mbf g^s_t \odot \hat{\x}_t, \nonumber\\
    \hat{\x}_t &\,=\, \tanh\left(A^\x \mspace{1mu}(\mbf g^r_t \odot \x_{t-1}) + B^\x \mspace{1mu} \bu_t + \mbf b^{\x}\right) \nonumber\\
    \mbf g^r_t &\,=\, \mbf \sigma\left(A^r \mspace{1mu} \x_{t-1} + B^r \mspace{1mu} \bu_t + \mbf b^{r}\right) \nonumber\\
    \mbf g^s_t &\,=\, \mbf \sigma\left(A^s \mspace{1mu} \x_{t-1} + B^s \mspace{1mu} \bu_t + \mbf b^{s}\right), \nonumber
\end{align}
where $\odot$ represents elementwise multiplication and $\bfpsi$ comprises the parameters $A^r, A^s, A^\x \in \R^{n_x\times n_x},\, B^r, B^s, B^\x \in \R^{n_x\times n_u},\, \mbf b^r, \mbf b^s, \mbf b^\x \in \R^{n_x}$.

\subsection{Additional Learning Details} \label{appdx:mtds:expmt-details:learn-inf}

\begin{table}[]
    \centering
    \footnotesize
    \begin{tabular}{c c c c c}
    \toprule
        Model & Optimizer & $\eta$ & Multi-task $\eta$ & Regularization \\ \midrule
        MT-RNN & Adam & $3 \times 10^{-5}$ & $1 \times 10^{-3}$ & $1 \times 10^{-2}$ \\ \addlinespace[0.15cm]
        GRU-1L (closed loop) & Adam & $5 \times 10^{-4}$ & - & $5 \times 10^{-4}$ \\ 
        GRU-2L (closed loop) & Adam & $1 \times 10^{-4}$ & - & $5 \times 10^{-4}$ \\ 
        GRU-1L (open loop) & Adam & $5 \times 10^{-4}$ & - & 0 \\
        GRU-2L (open loop) & Adam & $1 \times 10^{-4}$ & - & 0 \\
        \bottomrule
    \end{tabular}
    \caption{Hyper-parameters of mocap models. $\eta$ denotes the learning rate.}
    \label{tbl:Mocap-supp:model-hyperparams}
\end{table}

\paragraph{MTDS, MT-Bias models:} Each sequence was broken into overlapping segments of length 64 (approx.\ two second intervals), with a different $\z$ per segment. Unlike open-loop training in \citet{martinez2017human}, predictions do not use the previous modelled output $\hat{\y}_{t-1}$ when predicting the following time step. This is required in previous work since the observations $\y_{1:\Tenc}$ are required as inputs in the encoding stage, and hence predictions are used in their place for the decoding stage. By use of an explicit latent $\z$, we avoid the need to perform encoding via the sequence model, and hence avoid the need to append the predictions to the inputs; the information from the inputs is already contained in the recurrent state.

The model was optimized for 20\,000 iterations with $N_{\textrm{batch}} = 16$ using the variational procedure in \secref{sec:MTDS:learning}. We used standard variational inference for each $\z^\ii$ (i.e.\ each posterior is parameterized directly), which worked better in general than amortized inference using an inference network. A form of KL annealing \citep{bowman2016generating} was also used for improving the quality of the latent description. Hyperparameters were chosen primarily via the ELBO, but the \emph{qualitative} training set fit was also considered where values were similar; specifically we compared the smoothness of animation, and the results of style transfer on a few candidate tasks. The choices are shown in Table \ref{tbl:Mocap-supp:model-hyperparams}. The main learning rate $\eta$ applies to the fixed parameters wrt.\ $\z$ (i.e. $\bfpsi_1, H$), and the multi-task learning rate applies to the parameter generation parameters $\bfphi$ (i.e. $\bfpsi_1, H, C_1$) and inference parameters $\bflambda$. L2 regularization was applied to $\bfphi, \bfpsi_1, H$. Models were learned for $k=3,5,7,9$, for which the optimal choice via the ELBO varied across experiments, but was most frequently either $k=5$ or 7. We report multiple values in the results to give the reader some insight into the importance of this hyperparameter. The model was implemented in PyTorch \citep{paszke2017automatic} and trained on an NVIDIA K80 GPU.

\paragraph{Benchmark Models:}  The models are trained on predicting length 64 sequences (chosen at random at each iteration), using the encoding from the previous $\Tenc=64$ frames. The hyper-parameters were chosen using a grid search over learning rate, regularization, and optimizer \{Adam, SGD\}. We perform the search over the pooled data for all 8 styles, with a stratified sample of 12.5\% held out for a validation set. The benchmark models were trained for a maximum of 20\,000 iterations with early stopping. See Table \ref{tbl:Mocap-supp:model-hyperparams} for the chosen hyperparameters.


\subsection{Additional Details of Results} \label{appdx:mocap:results}
This section provides further details of the qualitative results: the latent representation (\secref{appdx:mocap:results:qual}) and style transfer experiment (\secref{appdx:mocap:results:style-tf})

\subsubsection{Qualitative Representation} \label{appdx:mocap:results:qual}

\begin{figure}
    \centering
        \begin{tikzpicture}
            \node[anchor=south west,inner sep=0] (image) at (0,0) {\includegraphics[trim={56 280 0 470}, clip, width=0.82\linewidth]{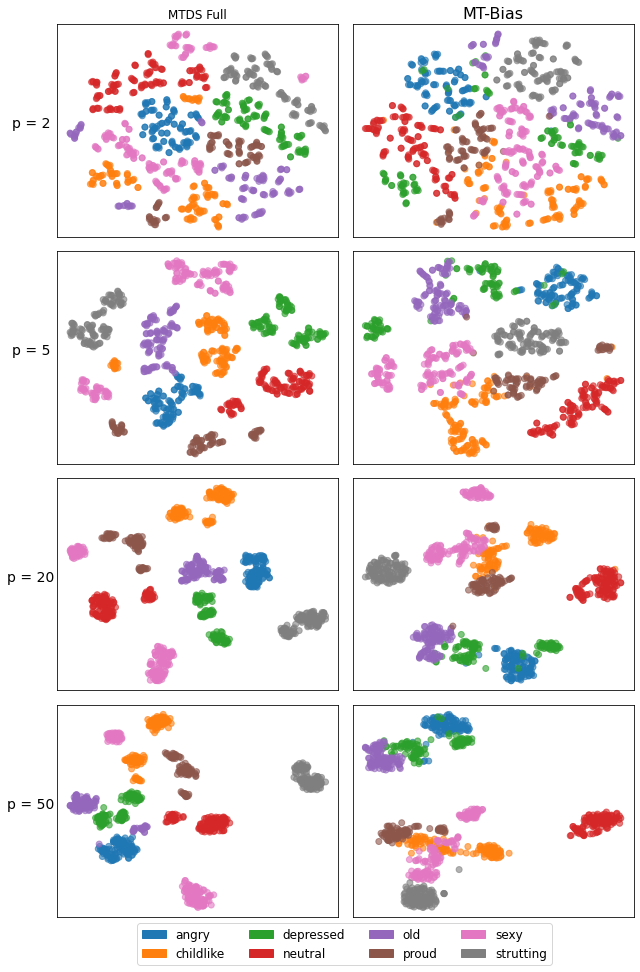}};
            \begin{scope}[x={(image.south east)},y={(image.north west)}]
                \node[anchor=south west,inner sep=0] (xlbl) at (0.2,1.00)  {\small MT Full};
                \node[anchor=south west,inner sep=0] (xlbl) at (0.7,1.00)  {\small MT-Bias};
            \end{scope}
        \end{tikzpicture}
    \includegraphics[trim={56 0 0 920}, clip, width=0.82\linewidth]{img/mocap/results/tsne_mtds_vs_mtbias_4x2.png}
    \caption{Mean embedding of latent $\z$ coloured by true style of the proposed MTDS model vs.\ a MT-Bias approach. Figure shows t-SNE plot for perplexity 20. The original $\z$ data have $k=8$ dimensions, and are standardized to have zero mean, unit variance.}
    \label{fig:Mocap-supp:mocap-tsne-comparison}
\end{figure}

Figure \ref{fig:Mocap-supp:mocap-tsne-comparison} compares the t-SNE embeddings for the MTDS and MT-Bias models.  The MTDS maintains well-defined clusters for different styles, which are not only homogeneous in terms of the original style label, but obtain a further split into qualitatively different sub-styles. In the MT-Bias case, there are many clusters which `bleed into each other'. This means that some differing styles are closer to each other in latent space than some instances of the same style. The representation of the sub-styles is also much weaker. These observations are valid for all values of the `perplexity' parameter in t-SNE, except for small values where little-to-no clustering occurs.

\subsubsection{Style Transfer} \label{appdx:mocap:results:style-tf}

\paragraph{Experimental Setup:} We pre-process the mocap sequences to standardize the gait frequency across all styles; some styles can be identified purely by calculating the frequency. It is critical for these experiments that the classifier cannot use this information. All sequences are standardized to use the inter-style mean frequency (1 cycle per 33 frames, or 1.1 Hz), which is applied via linear interpolation. This process is made straight-forward by the presence of the phase given in the input sequence $U$.

The experimental setup is as follows. We choose four segments of length 64 for each of the styles $s_1 = 1,\ldots,8$, being careful to represent the variability within each style. These segments are used as the input data $U_j^{(s_1)}$ for each source style $s_1$ with $j=1,\ldots,4$ examples. We next seek the `canonical' latent code $\z$ to be associated with each target style $s_2$. To this end, we consider the posterior mean ($\mulam$) of each training sequence in $\gD$, but since many styles have multiple clusters (as can be seen in Figure \ref{fig:mocap:latent-viz}), the mean of the $\{\mulam\}$ within each style may therefore not correspond to any existing style. We circumvent this problem by optimizing the latent code $\z^{(s_2)}$ as a discrete choice from among the relevant $\{\mulam\}$, choosing the value that performs the best at our style transfer task. The 32 highly varied input sequences guard against overfitting---the `canonical' latent codes for each style must perform well for all styles and sub-styles; the majority of the variability of the original dataset. We perform this independently for both the MT-Bias and MTDS model.

We provide a scalar measurement of the `success' of style transfer for each pair ($s_1$, $s_2$) by using the resulting `probability' that the classifier assigns the target style $s_2$, averaged across the four input sequences for the source $s_1$. A score of 1.00 thus requires the classifier to output a highly certain prediction of the target style $s_2$ for all of the 32 different input sequences $\{U_j^{(s_1)}\}$.

\begin{figure}[p]
    \centering
    
    \begin{subfigure}{.45\textwidth}
        \centering
        \includegraphics[trim={40 0 60 30}, clip, width=\textwidth]{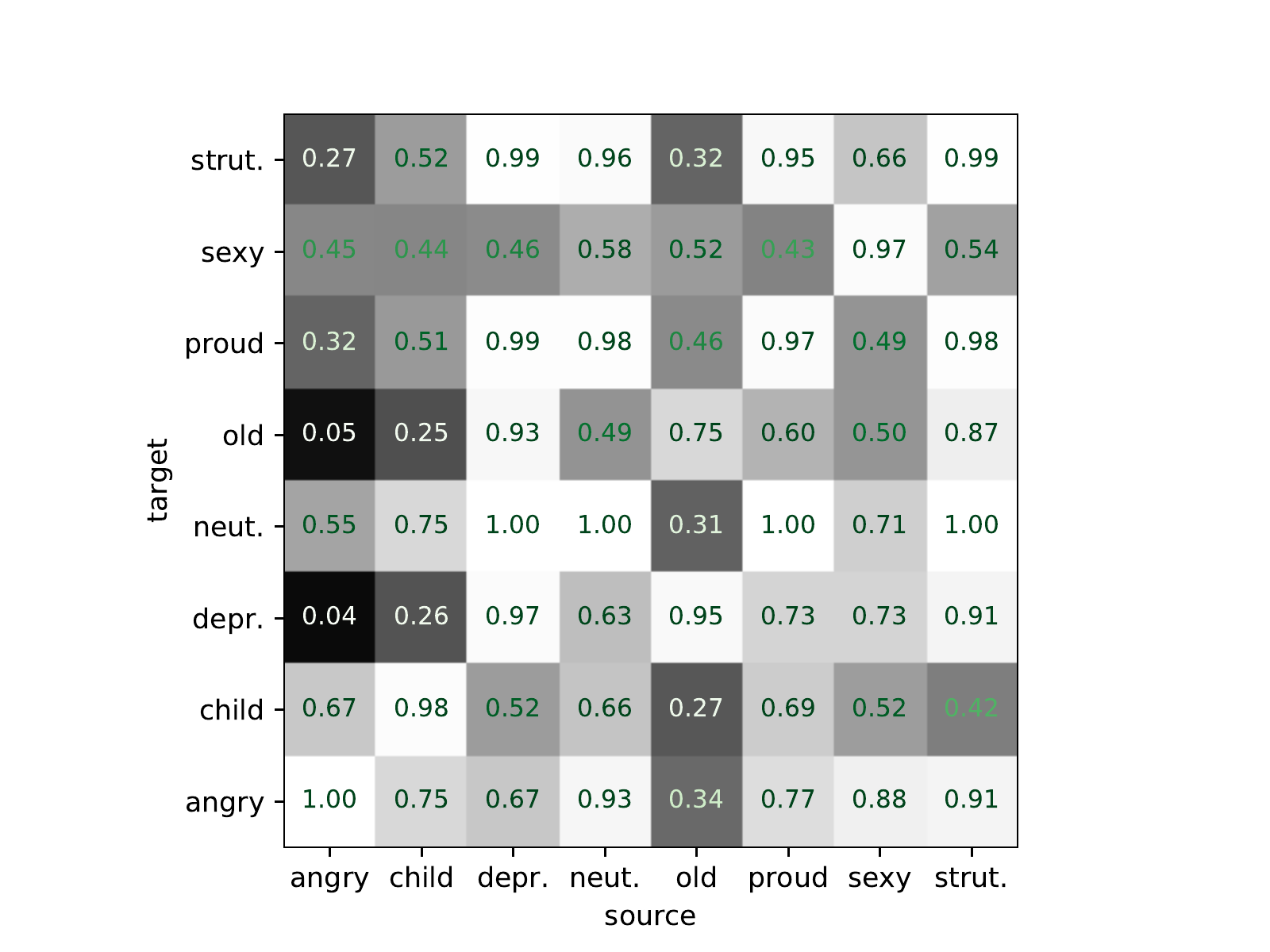}
        \caption{MT-Bias}
    \label{fig:mocap:expmt3:bdown4:mtbias}
    \end{subfigure} %
    ~
    \begin{subfigure}{.45\textwidth}
        \centering
        \includegraphics[trim={40 0 60 30}, clip, width=\textwidth]{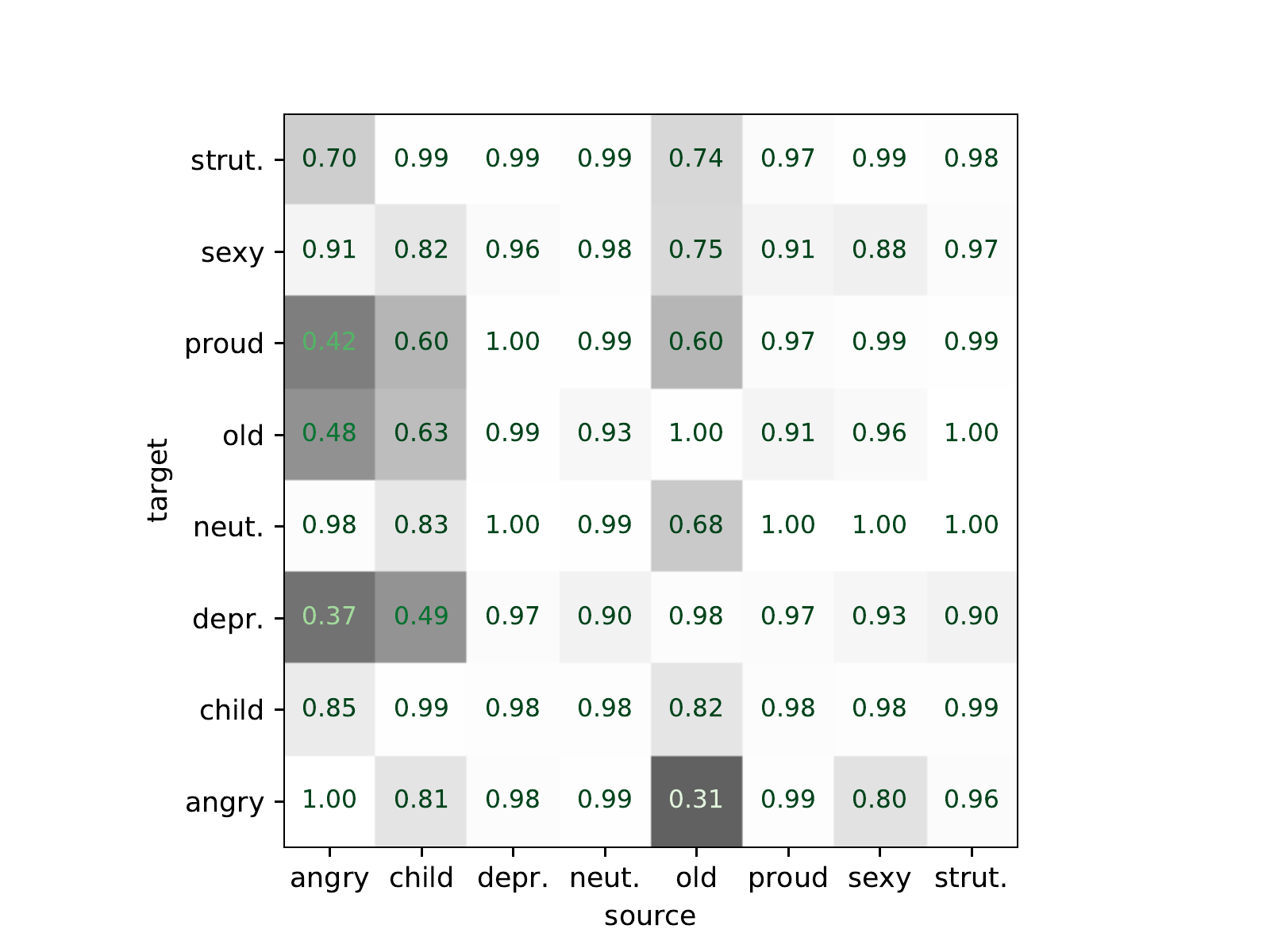}
        \caption{MTDS}
    \label{fig:mocap:expmt3:bdown4:mtds}  
    \end{subfigure} %
    
    \caption{Average classification accuracy for style transfer using inputs from source style (columns) and latent code $\z$ from target style (rows). There is no style transfer on the diagonal.
    }
    \label{fig:mocap:expmt3:bdown4}
\end{figure}

\begin{figure}
    \centering
    
    \begin{subfigure}{.45\textwidth}
        \centering
        \includegraphics[trim={40 0 60 30}, clip, width=\textwidth]{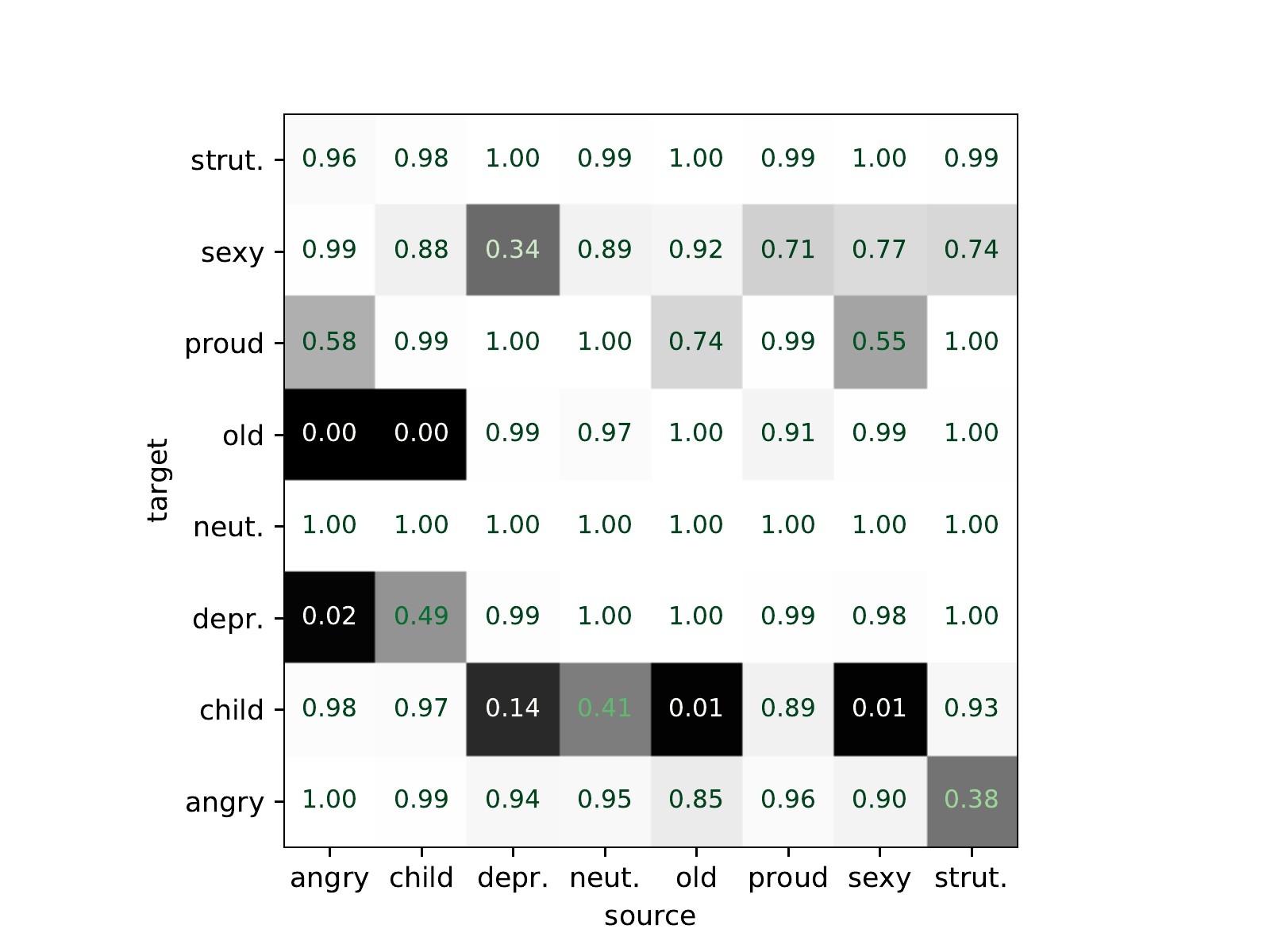}
        \caption{MT-Bias}
    \label{fig:mocap:expmt3:bdown1:mtbias}
    \end{subfigure} %
    ~
    \begin{subfigure}{.45\textwidth}
        \centering
        \includegraphics[trim={40 0 60 30}, clip, width=\textwidth]{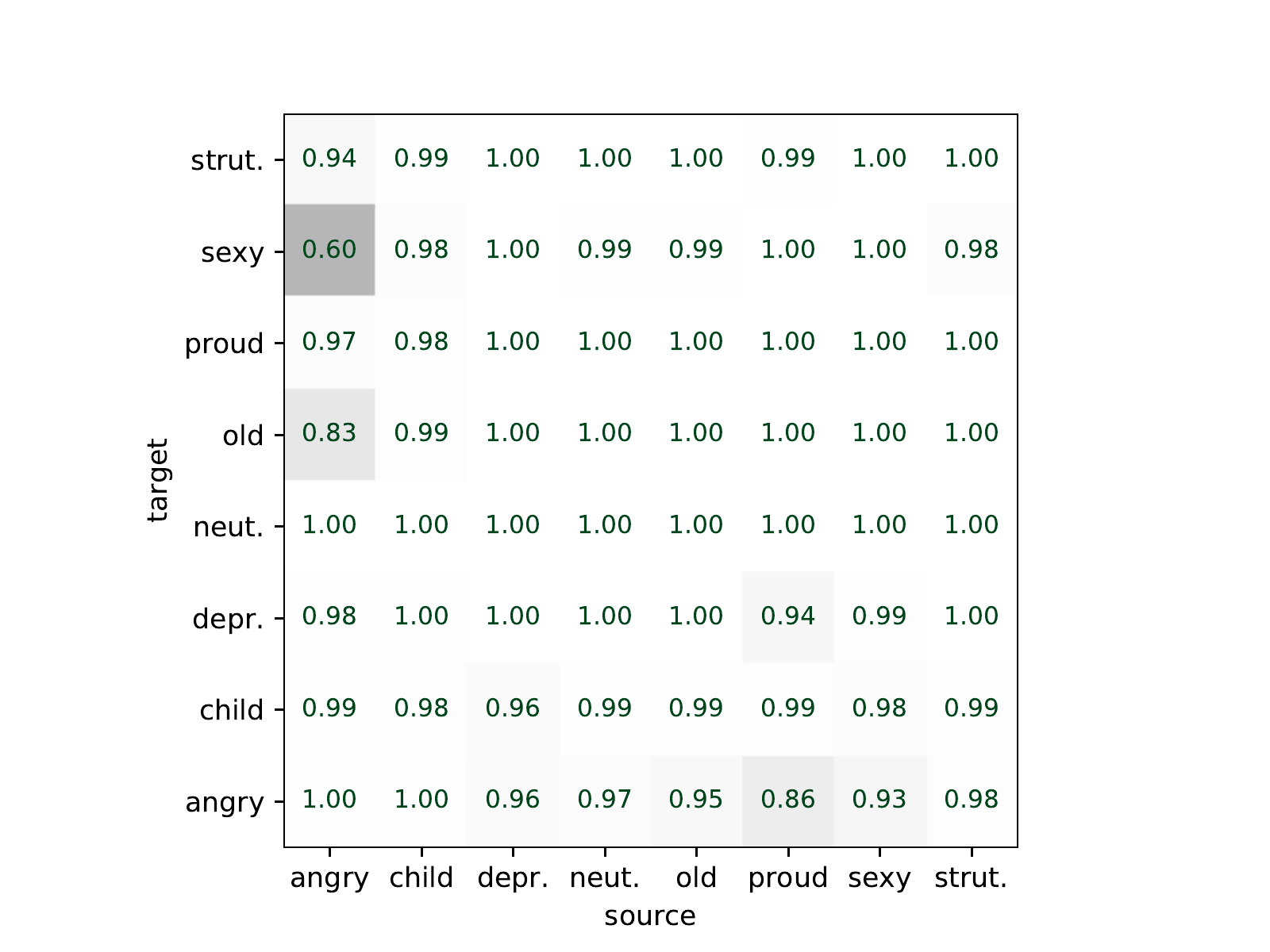}
        \caption{MTDS}
    \label{fig:mocap:expmt3:bdown1:mtds}  
    \end{subfigure} %
    
    \caption{Average classification accuracy for style transfer where only a single source input is used for each (source, target) pair. The configuration of the matrix is the same as Figure \ref{fig:mocap:expmt3:bdown4}.
    }
    \label{fig:mocap:expmt3:bdown1}
\end{figure}

\paragraph{Results Breakdown:} The breakdown of these results into their (source, target) pairs are given in Figure \ref{fig:mocap:expmt3:bdown4}. The cells provide the average classifier probability for the target style over each combination (averaged over the four source sequences). Successful style transfer should result in a high score in every cell.
For most (source, target) pairs, the full MTDS model substantially outperforms the MT-Bias model, resulting in superior user control in the majority of cases. We can gain some insight into the MT-Bias performance via its latent representation (Figure \ref{fig:Mocap-supp:mocap-tsne-comparison}), where we see it has conflated a variety of styles, which are disambiguated therefore only via the inputs. It is therefore not surprising that changing the latent $\z$ often results in unrecognizable changes.

It is notable that both models exhibit worse results when styles are associated with extremes of the input distribution. Specifically, both the `childlike' and `angry' styles have unusually fast trajectories, and the `old' style has unusually slow ones. The lowest scores tend to involve these styles as either the source or the target. This suggests that the first layer (GRU) fails to provide a coherent shared representation of these behaviours, and/or there is some information leakage from the inputs. We leave further investigation and improvements to future work.

Providing style transfer from a wide variety of source styles is a challenging task. We are attempting to find a single latent code which can reliably transfer style from 28 different source sequences, many of which may be mismatched to the target style. We consider a more pragmatic experiment where the variety of source styles is reduced to a single example each. Nonetheless, the same $\z^{(s_2)}$ is  used across all sources $s_1$. The results of this secondary experiment are provided in Figure \ref{fig:mocap:expmt3:bdown1}. In this case, the MTDS achieves successful style transfer for almost all (source, target) combinations. The MT-Bias model still has many notable failures.

\section{Application to drug response data} \label{appdx:mtpd}
In this section we provide additional details pertaining to the pharmacodynamic application in \secref{sec:mtpd}.



\subsection{Data: Infusion Sequences} \label{sec:MTPD-supp:data}

\begin{figure}
    \centering
    \includegraphics[trim={0 0 0 23}, clip, height=5cm]{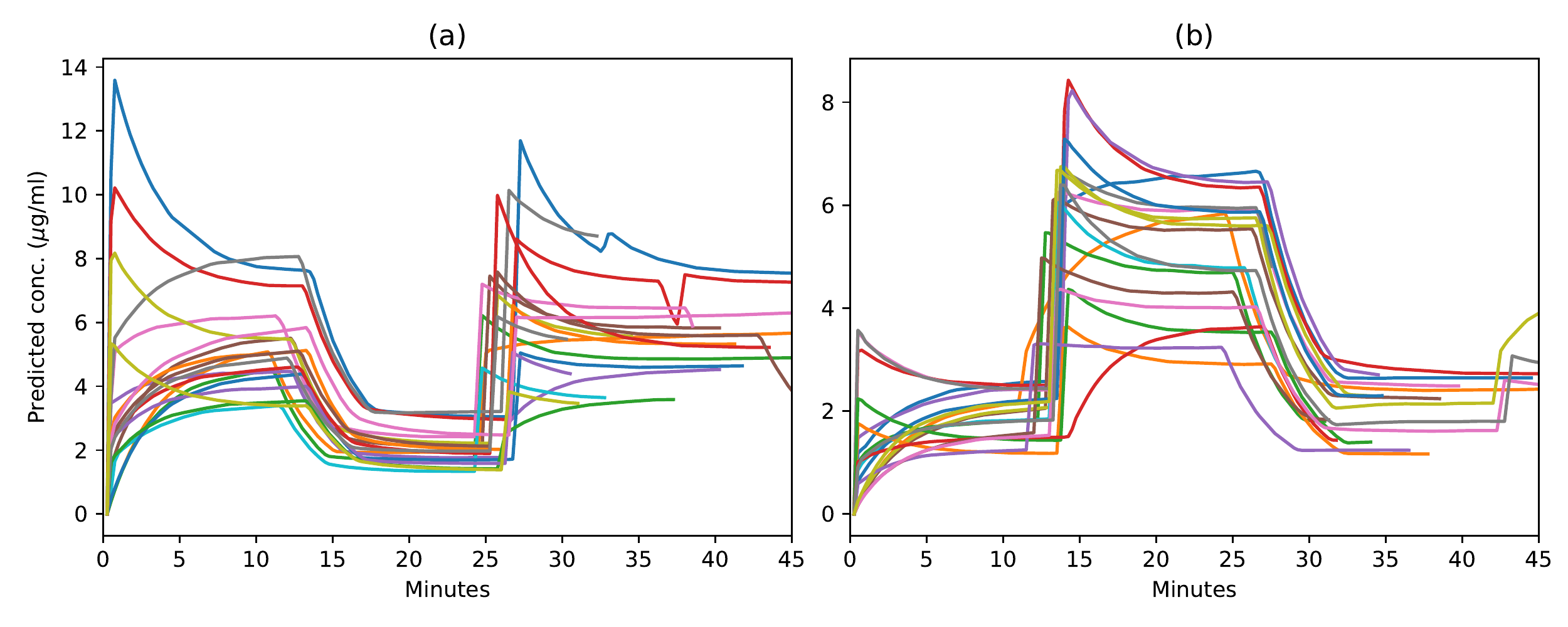}
    \caption{Predictions of \citet{white2008use} PK model for each of the 40 patients, split by infusion schedule type: (\emph{left}) high-low-high, (\emph{right}) low-high-low. The inter-individual differences seen here are primarily due to the raw infusion chosen by the anaesthetist, rather than the modelled inter-patient variation. }
    \label{fig:mtpd:data:inputs}
\end{figure}

The $N=40$ patients are split into two groups, each of which is given a different intravenous schedule of propofol. The two different schedules split the time into three consecutive segments of 10-15 minutes each, within which the TCI pump targeted a high-low-high, or a low-high-low concentration of propofol. The two schedules are visualized in Figure \ref{fig:mtpd:data:inputs} using the propofol concentration predicted via the \citet{white2008use} model.

\subsection{Derivation of Discrete-Time Relationship for Piecewise Constant $\mbf u(t)$}
\label{sec:MTPD-supp:model:wlog-pwise-const}

Equation (\ref{eq:mtpd:PD-ODE}) can be solved by integration using Laplace transformations. Let $U(s)$ be the Laplace transform of $c(t)$, and $X(s)$ the Laplace transform of $x(t)$. Define further a piecewise constant $u(t) = \sum_{i=0}^{T-1} R(t-i) u_i$ with a slight abuse of notation for $u$, and for $R(\cdot)$ the unit rectangle function. Then we have:
\begin{align}
    s X(s) &= k_{1e} U(s) - k_{e0} X(s) \\\Rightarrow X(s) &= \frac{k_{1e} U(s)}{s + k_{e0}}.
    \intertext{Recognising the RHS as the product of two known Laplace transformations gives:}
    \mathcal{L}(x(t)) &= \mathcal{L}(k_{1e} u(t)) \cdot \mathcal{L}(e^{-k_{e0} t} H(t))
    \intertext{for the Heaviside step function $H(t)$. Then using the convolution theorem:}
\Rightarrow x(t) &= k_{1e} \int_{0}^t e^{-k_{e0}\tau} u(t - \tau) \dif \tau\\
    &= k_{1e} \left[ \sum_{i=0}^{t-1} u_i \int_{0}^t e^{-k_{e0}\tau} R(t - \tau - i) \dif \tau \right] \\
    &= k_{1e} \left[ \sum_{i=0}^{t-1} u_i \int_{t-i-1}^{t-i} e^{-k_{e0}\tau} \dif \tau \right] \\
    &= k_{1e} \sum_{i=0}^{t-1} u_i \frac{1}{k_{e0}}\left[ e^{-k_{e0}(t-i-1)} - e^{-k_{e0}(t-i)} \right]\\
    &= \sum_{i=0}^{t-1} \frac{k_{1e}}{k_{e0}} u_i \left(1 - e^{-k_{e0}}\right) e^{-k_{e0}(t-i-1)} \\
    &= \frac{k_{1e}}{k_{e0}} \left(1 - e^{-k_{e0}}\right) \sum_{i=0}^{t-1} u_i (e^{-k_{e0}})^{t-i-1}. \label{eq:deriv-discrete-beta}
\end{align}
Equation (\ref{eq:mtpd:vanilla-pd-recurrence}) can be written explicitly as $x_{tj} = \beta_{1j} \sum_{i=0}^{t-1} \beta_{2j}^{t-i-1} u_{i}$ by unrolling the recursion. The relationships given in Section \ref{sec:mtpd:model:singletask} are then derived by comparison with eq. (\ref{eq:deriv-discrete-beta}).

\vskip 0.2in
\bibliography{biblio}

\begin{thebibliography}{103}
\providecommand{\natexlab}[1]{#1}
\providecommand{\url}[1]{\texttt{#1}}
\expandafter\ifx\csname urlstyle\endcsname\relax
  \providecommand{\doi}[1]{doi: #1}\else
  \providecommand{\doi}{doi: \begingroup \urlstyle{rm}\Url}\fi

\bibitem[Alaa et~al.(2018)Alaa, Yoon, Hu, and van~der
  Schaar]{alaa2018personalized}
Ahmed~M Alaa, Jinsung Yoon, Scott Hu, and Mihaela van~der Schaar.
\newblock {Personalized Risk Scoring for Critical Care Prognosis using Mixtures
  of Gaussian Processes}.
\newblock \emph{IEEE Transactions on Biomedical Engineering}, 65\penalty0
  (1):\penalty0 207--218, 2018.

\bibitem[\'Alvarez et~al.(2012)\'Alvarez, Rosasco, and
  Lawrence]{alvarez2012kernels}
Mauricio~A \'Alvarez, Lorenzo Rosasco, and Neil~D Lawrence.
\newblock {Kernels for Vector-Valued Functions: A Review}.
\newblock \emph{Foundations and Trends{\textregistered} in Machine Learning},
  4\penalty0 (3):\penalty0 195--266, 2012.

\bibitem[\'Alvarez et~al.(2013)\'Alvarez, Luengo, and
  Lawrence]{alvarez2013linear}
Mauricio~A \'Alvarez, David Luengo, and Neil~D Lawrence.
\newblock {Linear Latent Force Models Using Gaussian Processes}.
\newblock \emph{IEEE Transactions on Pattern Analysis and Machine
  Intelligence}, 35:\penalty0 2693--2705, 2013.

\bibitem[Ando and Zhang(2005)]{ando2005framework}
Rie~Kubota Ando and Tong Zhang.
\newblock {A Framework for Learning Predictive Structures from Multiple Tasks
  and Unlabeled Data}.
\newblock \emph{Journal of Machine Learning Research}, 6\penalty0
  (Nov):\penalty0 1817--1853, 2005.

\bibitem[Astr{\"o}m and Murray(2010)]{astrom2010feedback}
Karl~Johan Astr{\"o}m and Richard~M Murray.
\newblock \emph{{Feedback Systems: an Introduction for Scientists and
  Engineers}}.
\newblock Princeton University Press, 2010.

\bibitem[Bailey and Haddad(2005)]{bailey2005drug}
James~M Bailey and Wassim~M Haddad.
\newblock {Drug Dosing Control in Clinical Pharmacology}.
\newblock \emph{IEEE Control Systems}, 25\penalty0 (2):\penalty0 35--51, 2005.

\bibitem[Bakker and Heskes(2003)]{bakker2003task}
Bart Bakker and Tom Heskes.
\newblock {Task Clustering and Gating for Bayesian Multitask Learning}.
\newblock \emph{Journal of Machine Learning Research}, 4\penalty0
  (May):\penalty0 83--99, 2003.

\bibitem[Barber(2012)]{barber2012bayesian}
David Barber.
\newblock \emph{{Bayesian Reasoning and Machine Learning}}.
\newblock Cambridge University Press, 2012.

\bibitem[Bengio et~al.(2015)Bengio, Vinyals, Jaitly, and
  Shazeer]{bengio2015scheduled}
Samy Bengio, Oriol Vinyals, Navdeep Jaitly, and Noam Shazeer.
\newblock {Scheduled Sampling for Sequence Prediction with Recurrent Neural
  Networks}.
\newblock In \emph{Advances in Neural Information Processing Systems (NeurIPS)
  29}, pages 1171--1179, 2015.

\bibitem[Bird(2021)]{bird21thesis}
Alex Bird.
\newblock \emph{Multi-Task Dynamical Systems}.
\newblock PhD thesis, School of Informatics, University of Edinburgh, 2021.
\newblock URL \url{https://era.ed.ac.uk/handle/1842/38267}.

\bibitem[Bird et~al.(2019)Bird, Williams, and Hawthorne]{bird19mtpd}
Alex Bird, Christopher K.~I. Williams, and Christopher Hawthorne.
\newblock {Multi-Task Time Series Analysis applied to Drug Response Modelling}.
\newblock In \emph{International Conference on Artificial Intelligence and
  Statistics (AISTATS)}, 2019.

\bibitem[Bonilla et~al.(2008)Bonilla, Chai, and Williams]{bonilla2008multi}
E.~Bonilla, K.~M.~A. Chai, and C.~K.~I. Williams.
\newblock {Multi-task Gaussian Process Prediction}.
\newblock In J.C. Platt, D.~Koller, Y.~Singer, and S.~Roweis, editors,
  \emph{Advances in Neural Information Processing Systems (NeurIPS) 21}. MIT
  Press, Cambridge, MA, 2008.

\bibitem[Bowman et~al.(2016)Bowman, Vilnis, Vinyals, Dai, Jozefowicz, and
  Bengio]{bowman2016generating}
Samuel Bowman, Luke Vilnis, Oriol Vinyals, Andrew~M Dai, Rafal Jozefowicz, and
  Samy Bengio.
\newblock {Generating Sentences from a Continuous Space}.
\newblock In \emph{Conference on Computational Natural Language Learning
  (CoNLL)}, 2016.

\bibitem[Burda et~al.(2016)Burda, Grosse, and Salakhutdinov]{burda2016iwae}
Yuri Burda, Roger Grosse, and Ruslan Salakhutdinov.
\newblock {Importance Weighted Autoencoders}.
\newblock In \emph{International Conference on Learning Representations
  (ICLR)}, 2016.

\bibitem[Capp{\'e} et~al.(2008)Capp{\'e}, Douc, Guillin, Marin, and
  Robert]{cappe2008adaptive}
Olivier Capp{\'e}, Randal Douc, Arnaud Guillin, Jean-Michel Marin, and
  Christian~P Robert.
\newblock {Adaptive Importance Sampling in General Mixture Classes}.
\newblock \emph{Statistics and Computing}, 18\penalty0 (4):\penalty0 447--459,
  2008.

\bibitem[Carpenter et~al.(2016)Carpenter, Gelman, Hoffman, Lee, Goodrich,
  Betancourt, Brubaker, Guo, Li, Riddell, et~al.]{carpenter2016stan}
Bob Carpenter, Andrew Gelman, Matt Hoffman, Daniel Lee, Ben Goodrich, Michael
  Betancourt, Michael~A Brubaker, Jiqiang Guo, Peter Li, Allen Riddell, et~al.
\newblock {Stan: A Probabilistic Programming Language}.
\newblock \emph{Journal of Statistical Software}, 20\penalty0 (2):\penalty0
  1--37, 2016.

\bibitem[Caruana(1998)]{caruana1998multitask}
R~Caruana.
\newblock \emph{{Multitask Learning}}.
\newblock PhD thesis, Carnegie Mellon University, 1998.

\bibitem[Chen et~al.(2017)Chen, Kingma, Salimans, Duan, Dhariwal, Schulman,
  Sutskever, and Abbeel]{chen2016variational}
Xi~Chen, Diederik~P Kingma, Tim Salimans, Yan Duan, Prafulla Dhariwal, John
  Schulman, Ilya Sutskever, and Pieter Abbeel.
\newblock {Variational Lossy Autoencoder}.
\newblock In \emph{International Conference on Learning Representations
  (ICLR)}, 2017.

\bibitem[Cheng et~al.(2020)Cheng, Dumitrascu, Zhang, Chivers, Draugelis, Li,
  and Engelhardt]{cheng2020patient}
Li-Fang Cheng, Bianca Dumitrascu, Michael Zhang, Corey Chivers, Michael
  Draugelis, Kai Li, and Barbara Engelhardt.
\newblock {Patient-Specific Effects of Medication Using Latent Force Models
  with Gaussian Processes}.
\newblock In \emph{International Conference on Artificial Intelligence and
  Statistics (AISTATS)}, 2020.

\bibitem[Chiappa and Barber(2007)]{chiappa2007output}
Silvia Chiappa and David Barber.
\newblock {Output Grouping using Dirichlet Mixtures of Linear Gaussian
  State-Space Models}.
\newblock In \emph{International Symposium on Image and Signal Processing and
  Analysis}. IEEE, 2007.

\bibitem[Chiappa et~al.(2017)Chiappa, Racaniere, Wierstra, and
  Mohamed]{chiappa2017recurrent}
Silvia Chiappa, S{\'e}bastien Racaniere, Daan Wierstra, and Shakir Mohamed.
\newblock {Recurrent Environment Simulators}.
\newblock In \emph{International Conference on Learning Representations
  (ICLR)}, 2017.

\bibitem[Cho et~al.(2014)Cho, van Merrienboer, Gulcehre, Bahdanau, Bougares,
  Schwenk, and Bengio]{cho2014gru}
Kyunghyun Cho, Bart van Merrienboer, Caglar Gulcehre, Dzmitry Bahdanau, Fethi
  Bougares, Holger Schwenk, and Yoshua Bengio.
\newblock {Learning Phrase Representations using RNN Encoder--Decoder for
  Statistical Machine Translation}.
\newblock In \emph{Conference on Empirical Methods in Natural Language
  Processing (EMNLP)}, 2014.

\bibitem[Chopin(2002)]{chopin2002sequential}
Nicolas Chopin.
\newblock {A Sequential Particle Filter Method for Static Models}.
\newblock \emph{Biometrika}, 89\penalty0 (3):\penalty0 539--552, 2002.

\bibitem[Chopin et~al.(2013)Chopin, Jacob, and
  Papaspiliopoulos]{chopin2013smc2}
Nicolas Chopin, Pierre~E Jacob, and Omiros Papaspiliopoulos.
\newblock {SMC2: an Efficient Algorithm for Sequential Analysis of State Space
  Models}.
\newblock \emph{Journal of the Royal Statistical Society: Series B (Statistical
  Methodology)}, 75\penalty0 (3):\penalty0 397--426, 2013.

\bibitem[De~la Torre et~al.(2009)De~la Torre, Hodgins, Bargteil, Martin, Macey,
  Collado, and Beltran]{cmu2009guide}
Fernando De~la Torre, Jessica Hodgins, Adam Bargteil, Xavier Martin, Justin
  Macey, Alex Collado, and Pep Beltran.
\newblock {Guide to the Carnegie Mellon University Multimodal Activity
  (CMU-MMAC) Database}.
\newblock 2009.

\bibitem[Denton and Birodkar(2017)]{denton2017video}
Emily~L Denton and Vighnesh Birodkar.
\newblock {Unsupervised Learning of Disentangled Representations from Video}.
\newblock In \emph{Advances in Neural Information Processing Systems (NeurIPS)
  30}, pages 4414--4423. 2017.

\bibitem[Doucet et~al.(2000)Doucet, Godsill, and Andrieu]{doucet2000onsmc}
Arnaud Doucet, Simon Godsill, and Christophe Andrieu.
\newblock {On Sequential Monte Carlo Sampling Methods for Bayesian Filtering}.
\newblock \emph{Statistics and Computing}, 10\penalty0 (3), 2000.

\bibitem[D{\"u}richen et~al.(2015)D{\"u}richen, Pimentel, Clifton, Schweikard,
  and Clifton]{durichen2015multitask}
Robert D{\"u}richen, Marco~AF Pimentel, Lei Clifton, Achim Schweikard, and
  David~A Clifton.
\newblock {Multitask Gaussian Processes for Multivariate Physiological
  Time-Series Analysis}.
\newblock \emph{IEEE Transactions on Biomedical Engineering}, 62\penalty0
  (1):\penalty0 314--322, 2015.

\bibitem[Eleveld et~al.(2018)Eleveld, Colin, Absalom, and
  Struys]{eleveld2018pharmacokinetic}
DJ~Eleveld, P~Colin, AR~Absalom, and MMRF Struys.
\newblock {Pharmacokinetic--Pharmacodynamic Model for Propofol for Broad
  Application in Anaesthesia and Sedation}.
\newblock \emph{British Journal of Anaesthesia}, 120\penalty0 (5):\penalty0
  942--959, 2018.

\bibitem[Fabius and van Amersfoort(2015)]{fabius2015variational}
Otto Fabius and Joost~R van Amersfoort.
\newblock {Variational Recurrent Auto-encoders}.
\newblock In \emph{International Conference on Learning Representations
  (ICLR)}, 2015.

\bibitem[Fragkiadaki et~al.(2015)Fragkiadaki, Levine, Felsen, and
  Malik]{fragkiadaki2015recurrent}
Katerina Fragkiadaki, Sergey Levine, Panna Felsen, and Jitendra Malik.
\newblock {Recurrent Network Models for Human Dynamics}.
\newblock In \emph{IEEE International Conference on Computer Vision (ICCV)},
  pages 4346--4354, 2015.

\bibitem[Futoma et~al.(2016)Futoma, Sendak, Cameron, and
  Heller]{futoma2016predicting}
Joseph Futoma, Mark Sendak, Blake Cameron, and Katherine Heller.
\newblock {Predicting Disease Progression with a Model for Multivariate
  Longitudinal Clinical Data}.
\newblock In \emph{Machine Learning for Healthcare Conference}, 2016.

\bibitem[Georgatzis et~al.(2016)Georgatzis, Williams, and
  Hawthorne]{georgatzis2016ionlds}
Konstantinos Georgatzis, Christopher K.~I. Williams, and Christopher Hawthorne.
\newblock {Input-Output Non-Linear Dynamical Systems applied to Physiological
  Condition Monitoring}.
\newblock \emph{Machine Learning for Healthcare}, 2016.

\bibitem[Ghosh et~al.(2017)Ghosh, Song, Aksan, and Hilliges]{ghosh2017learning}
Partha Ghosh, Jie Song, Emre Aksan, and Otmar Hilliges.
\newblock {Learning Human Motion Models for Long-Term Predictions}.
\newblock In \emph{IEEE International Conference on 3D Vision (3DV)}, pages
  458--466, 2017.

\bibitem[Gilks and Berzuini(2001)]{gilks2001following}
Walter~R Gilks and Carlo Berzuini.
\newblock {Following a Moving Target -- Monte Carlo Inference for Dynamic
  Bayesian Models}.
\newblock \emph{Journal of the Royal Statistical Society: Series B (Statistical
  Methodology)}, 63\penalty0 (1):\penalty0 127--146, 2001.

\bibitem[Glen and White(2014)]{glen2014comparison}
JB~Glen and M~White.
\newblock {A Comparison of the Predictive Performance of Three Pharmacokinetic
  Models for Propofol Using Measured Values Obtained During Target-Controlled
  Infusion}.
\newblock \emph{Anaesthesia}, 69\penalty0 (6):\penalty0 550--557, 2014.

\bibitem[Goodfellow et~al.(2016)Goodfellow, Bengio, and
  Courville]{goodfellow2016deep}
Ian Goodfellow, Yoshua Bengio, and Aaron Courville.
\newblock \emph{{Deep Learning}}.
\newblock MIT Press, 2016.

\bibitem[Goyal et~al.(2017)Goyal, Sordoni, C{\^o}t{\'e}, Ke, and
  Bengio]{goyal2017zforce}
Anirudh Goyal, Alessandro Sordoni, Marc-Alexandre C{\^o}t{\'e}, Nan~Rosemary
  Ke, and Yoshua Bengio.
\newblock {Z-Forcing: Training Stochastic Recurrent Networks}.
\newblock In \emph{Advances in Neural Information Processing Systems (NeurIPS)
  31}, pages 6713--6723, 2017.

\bibitem[Harvey(1990)]{harvey1990forecasting}
Andrew~C Harvey.
\newblock \emph{{Forecasting, Structural Time Series Models and the Kalman
  Filter}}.
\newblock Cambridge University Press, 1990.

\bibitem[Hochreiter and Schmidhuber(1997)]{hochreiter1997long}
Sepp Hochreiter and J{\"u}rgen Schmidhuber.
\newblock {Long Short-Term Memory}.
\newblock \emph{Neural Computation}, 9\penalty0 (8):\penalty0 1735--1780, 1997.

\bibitem[Hoffman and Gelman(2014)]{hoffman2014no}
Matthew~D Hoffman and Andrew Gelman.
\newblock {The No-U-Turn Sampler: Adaptively Setting Path Lengths in
  Hamiltonian Monte Carlo.}
\newblock \emph{Journal of Machine Learning Research}, 15\penalty0
  (1):\penalty0 1593--1623, 2014.

\bibitem[Holden et~al.(2016)Holden, Saito, and Komura]{holden2016deep}
Daniel Holden, Jun Saito, and Taku Komura.
\newblock {A Deep Learning Framework for Character Motion Synthesis and
  Editing}.
\newblock \emph{ACM Transactions on Graphics (TOG)}, 35\penalty0 (4):\penalty0
  138, 2016.

\bibitem[Holden et~al.(2017)Holden, Komura, and Saito]{holden2017phase}
Daniel Holden, Taku Komura, and Jun Saito.
\newblock {Phase-Functioned Neural Networks for Character Control}.
\newblock \emph{ACM Transactions on Graphics (TOG)}, 36\penalty0 (4):\penalty0
  42, 2017.

\bibitem[Holford(2018)]{holford2018pharmacodynamic}
Nick Holford.
\newblock {Pharmacodynamic Principles and the Time Course of Delayed and
  Cumulative Drug Effects}.
\newblock \emph{Translational and Clinical Pharmacology}, 26\penalty0
  (2):\penalty0 56--59, 2018.

\bibitem[Hsieh et~al.(2018)Hsieh, Liu, Huang, Fei-Fei, and
  Niebles]{hsieh2018learning}
Jun-Ting Hsieh, Bingbin Liu, De-An Huang, Li~F Fei-Fei, and Juan~Carlos
  Niebles.
\newblock {Learning to Decompose and Disentangle Representations for Video
  Prediction}.
\newblock In \emph{Advances in Neural Information Processing Systems (NeurIPS)
  31}, pages 517--526, 2018.

\bibitem[Hsu et~al.(2017)Hsu, Zhang, and Glass]{hsu2017disentangledseq}
Wei-Ning Hsu, Yu~Zhang, and James Glass.
\newblock {Unsupervised Learning of Disentangled and Interpretable
  Representations from Sequential Data}.
\newblock In \emph{Advances in Neural Information Processing Systems (NeurIPS)
  30}, pages 1876--1887. 2017.

\bibitem[H\"uppe et~al.(2019)H\"uppe, Maurer, Sessler, Volk, and
  Kreuer]{huppe2019eleveldretrospective}
Tobias H\"uppe, Felix Maurer, Daniel~I. Sessler, Thomas Volk, and Sascha
  Kreuer.
\newblock {Retrospective comparison of Eleveld, Marsh, and Schnider propofol
  pharmacokinetic models in 50 patients}.
\newblock \emph{British Journal of Anaesthesia}, 2019.

\bibitem[Inoue et~al.(2007)Inoue, Neira, Nelson, Gleave, and
  Etzioni]{inoue2007cluster}
Lurdes~YT Inoue, Mauricio Neira, Colleen Nelson, Martin Gleave, and Ruth
  Etzioni.
\newblock {Cluster-Based Network Model for Time-Course Gene Expression Data}.
\newblock \emph{Biostatistics}, 8:\penalty0 507--525, 2007.

\bibitem[Jeleazcov et~al.(2015)Jeleazcov, Lavielle, Sch{\"u}ttler, and
  Ihmsen]{jeleazcov2015pharmacodynamic}
Christian Jeleazcov, Marc Lavielle, J{\"u}rgen Sch{\"u}ttler, and Harald
  Ihmsen.
\newblock {Pharmacodynamic Response Modelling of Arterial Blood Pressure in
  Adult Volunteers During Propofol Anaesthesia}.
\newblock \emph{British Journal of Anaesthesia}, 115\penalty0 (2):\penalty0
  213--226, 2015.

\bibitem[Karl et~al.(2017)Karl, Soelch, Bayer, and van~der
  Smagt]{karl2016dvbfilter}
Maximilian Karl, Maximilian Soelch, Justin Bayer, and Patrick van~der Smagt.
\newblock {Deep Variational Bayes Filters: Unsupervised Learning of State Space
  Models from Raw Data}.
\newblock In \emph{International Conference on Learning Representations
  (ICLR)}, 2017.

\bibitem[Kingma and Welling(2014)]{kingma2014vae}
Diederik~P Kingma and Max Welling.
\newblock {Stochastic Gradient VB and the Variational Auto-Encoder}.
\newblock In \emph{{International Conference on Learning Representations
  (ICLR)}}, 2014.

\bibitem[Le et~al.(2018)Le, Igl, Rainforth, Jin, and Wood]{anh2018autoencoding}
Tuan~Anh Le, Maximilian Igl, Tom Rainforth, Tom Jin, and Frank Wood.
\newblock {Auto-Encoding Sequential Monte Carlo}.
\newblock In \emph{International Conference on Learning Representations
  (ICLR)}, 2018.

\bibitem[Lee and Grimson(2002)]{lee2002gait}
Lily Lee and W~Eric~L Grimson.
\newblock {Gait Analysis for Recognition and Classification}.
\newblock In \emph{IEEE International Conference on Automatic Face Gesture
  Recognition}. IEEE, 2002.

\bibitem[Lin et~al.(2019)Lin, Zhang, Heng, Allsop, Tye, Jacob, and
  Ba]{lin2019clustering}
Alexander Lin, Yingzhuo Zhang, Jeremy Heng, Stephen~A Allsop, Kay~M Tye,
  Pierre~E Jacob, and Demba Ba.
\newblock {Clustering Time Series with Nonlinear Dynamics: A Bayesian
  Non-Parametric and Particle-Based Approach}.
\newblock In \emph{International Conference on Artificial Intelligence and
  Statistics (AISTATS)}, 2019.

\bibitem[Linderman et~al.(2017)Linderman, Johnson, Miller, Adams, Blei, and
  Paninski]{linderman2017bayesian}
Scott Linderman, Matthew Johnson, Andrew Miller, Ryan Adams, David Blei, and
  Liam Paninski.
\newblock {Bayesian Learning and Inference in Recurrent Switching Linear
  Dynamical Systems}.
\newblock In \emph{International Conference on Artificial Intelligence and
  Statistics (AISTATS)}, pages 914--922, 2017.

\bibitem[Lobo and Schraag(2011)]{lobo2011limitationsBis}
Francisco~A Lobo and Stefan Schraag.
\newblock {Limitations of Anaesthesia Depth Monitoring}.
\newblock \emph{Current Opinion in Anesthesiology}, 24\penalty0 (6):\penalty0
  657--664, 2011.

\bibitem[Maddison et~al.(2017)Maddison, Lawson, Tucker, Heess, Norouzi, Mnih,
  Doucet, and Teh]{maddison2017filtering}
Chris~J Maddison, John Lawson, George Tucker, Nicolas Heess, Mohammad Norouzi,
  Andriy Mnih, Arnaud Doucet, and Yee Teh.
\newblock {Filtering Variational Objectives}.
\newblock In \emph{Advances in Neural Information Processing Systems (NeurIPS)
  31}, pages 6573--6583, 2017.

\bibitem[Mager et~al.(2003)Mager, Wyska, and Jusko]{mager2003diversity}
Donald~E Mager, Elzbieta Wyska, and William~J Jusko.
\newblock {Diversity of Mechanism-Based Pharmacodynamic Models}.
\newblock \emph{Drug Metabolism and Disposition}, 31\penalty0 (5):\penalty0
  510--518, 2003.

\bibitem[Marsh et~al.(1991)Marsh, White, Morton, and
  Kenny]{marsh1991pharmacokinetic}
B~Marsh, M~White, N~Morton, and GNC Kenny.
\newblock {Pharmacokinetic Model Driven Infusion of Propofol in Children}.
\newblock \emph{British Journal of Anaesthesia}, 67\penalty0 (1):\penalty0
  41--48, 1991.

\bibitem[Martinez et~al.(2017)Martinez, Black, and Romero]{martinez2017human}
Julieta Martinez, Michael~J Black, and Javier Romero.
\newblock {On Human Motion Prediction Using Recurrent Neural Networks}.
\newblock In \emph{IEEE Conference on Computer Vision and Pattern Recognition
  (CVPR)}, pages 2891--2900, 2017.

\bibitem[Mason et~al.(2018)Mason, Starke, Zhang, Bilen, and
  Komura]{mason2018few}
Ian Mason, Sebastian Starke, He~Zhang, Hakan Bilen, and Taku Komura.
\newblock {Few-Shot Learning of Homogeneous Human Locomotion Styles}.
\newblock In \emph{Computer Graphics Forum}, volume~37, pages 143--153. Wiley
  Online Library, 2018.

\bibitem[Masui et~al.(2010)Masui, Upton, Doufas, Coetzee, Kazama, Mortier, and
  Struys]{masui2010performance}
Kenichi Masui, Richard~N Upton, Anthony~G Doufas, Johan~F Coetzee, Tomiei
  Kazama, Eric~P Mortier, and Michel~M Struys.
\newblock {The Performance of Compartmental and Physiologically Based
  Recirculatory Pharmacokinetic Models for Propofol: a Comparison Using Bolus,
  Continuous, and Target-Controlled Infusion Data}.
\newblock \emph{Anesthesia \& Analgesia}, 111\penalty0 (2):\penalty0 368--379,
  2010.

\bibitem[Memisevic and Hinton(2007)]{memisevic2007unsupervised}
Roland Memisevic and Geoffrey Hinton.
\newblock {Unsupervised Learning of Image Transformations}.
\newblock In \emph{IEEE Conference on Computer Vision and Pattern Recognition
  (CVPR)}, pages 1--8. IEEE, 2007.

\bibitem[Memisevic and Hinton(2010)]{memisevic2010learning}
Roland Memisevic and Geoffrey~E Hinton.
\newblock {Learning to Represent Spatial Transformations with Factored
  Higher-Order Boltzmann Machines}.
\newblock \emph{Neural Computation}, 22\penalty0 (6):\penalty0 1473--1492,
  2010.

\bibitem[{Miladinovi{\'c}} et~al.(2019){Miladinovi{\'c}}, {Waleed Gondal},
  {Sch{\"o}lkopf}, {Buhmann}, and {Bauer}]{miladinovic2019dssm}
{\DJ}or{\dj}e {Miladinovi{\'c}}, Muhammad {Waleed Gondal}, Bernhard
  {Sch{\"o}lkopf}, Joachim~M. {Buhmann}, and Stefan {Bauer}.
\newblock {Disentangled State Space Representations}.
\newblock \emph{arXiv e-prints}, art. 1906.03255, Jun 2019.

\bibitem[Mukamel et~al.(2014)Mukamel, Pirondini, Babadi, Wong, Pierce, Harrell,
  Walsh, Salazar-Gomez, Cash, Eskandar, et~al.]{mukamel2014transition}
Eran~A Mukamel, Elvira Pirondini, Behtash Babadi, Kin Foon~Kevin Wong, Eric~T
  Pierce, P~Grace Harrell, John~L Walsh, Andres~F Salazar-Gomez, Sydney~S Cash,
  Emad~N Eskandar, et~al.
\newblock {A Transition in Brain State During Propofol-Induced
  Unconsciousness}.
\newblock \emph{Journal of Neuroscience}, 34\penalty0 (3):\penalty0 839--845,
  2014.

\bibitem[Myles et~al.(2004)Myles, Leslie, McNeil, Forbes, Chan, Group,
  et~al.]{myles2004bispectral}
PS~Myles, K~Leslie, J~McNeil, A~Forbes, MTV Chan, B-Aware~Trial Group, et~al.
\newblock {Bispectral Index Monitoring to Prevent Awareness During Anaesthesia:
  the B-Aware Randomised Controlled Trial}.
\newblock \emph{The Lancet}, 363\penalty0 (9423):\penalty0 1757--1763, 2004.

\bibitem[Naesseth et~al.(2018)Naesseth, Linderman, Ranganath, and
  Blei]{naesseth2018variational}
Christian Naesseth, Scott Linderman, Rajesh Ranganath, and David Blei.
\newblock {Variational Sequential Monte Carlo}.
\newblock In \emph{International Conference on Artificial Intelligence and
  Statistics (AISTATS)}, pages 968--977, 2018.

\bibitem[Opper and Winther(1998)]{opper1998bayesian}
Manfred Opper and Ole Winther.
\newblock {A Bayesian Approach to Online Learning}.
\newblock \emph{On-line Learning in Neural Networks}, 6:\penalty0 363--378,
  1998.

\bibitem[Osborne et~al.(2008)Osborne, Roberts, Rogers, Ramchurn, and
  Jennings]{osborne2008towards}
Michael~A Osborne, Stephen~J Roberts, Alex Rogers, Sarvapali~D Ramchurn, and
  Nicholas~R Jennings.
\newblock {Towards Real-Time Information Processing of Sensor Network Data
  Using Computationally Efficient Multi-Output Gaussian Processes}.
\newblock In \emph{International Conference on Information Processing in Sensor
  Networks}, pages 109--120. IEEE Computer Society, 2008.

\bibitem[Owen(2013)]{owen2013mcbook}
Art~B. Owen.
\newblock \emph{{Monte Carlo Theory, Methods and Examples}}.
\newblock 2013.
\newblock Published online at time of writing, retrieved from
  https://artowen.su.domains/mc/.

\bibitem[Paszke et~al.(2017)Paszke, Gross, Chintala, Chanan, Yang, DeVito, Lin,
  Desmaison, Antiga, and Lerer]{paszke2017automatic}
Adam Paszke, Sam Gross, Soumith Chintala, Gregory Chanan, Edward Yang, Zachary
  DeVito, Zeming Lin, Alban Desmaison, Luca Antiga, and Adam Lerer.
\newblock {Automatic Differentiation in PyTorch}.
\newblock 2017.

\bibitem[Pavllo et~al.(2018)Pavllo, Grangier, and Auli]{pavllo2018quaternet}
Dario Pavllo, David Grangier, and Michael Auli.
\newblock {Quaternet: A Quaternion-Based Recurrent Model for Human Motion}.
\newblock In \emph{British Machine Vision Conference ({BMVC})}, 2018.

\bibitem[Quinn et~al.(2009)Quinn, Williams, and McIntosh]{quinn2009factorial}
John~A Quinn, Christopher K~I Williams, and Neil McIntosh.
\newblock {Factorial Switching Linear Dynamical Systems applied to
  Physiological Condition Monitoring}.
\newblock \emph{IEEE Transactions on Pattern Analysis and Machine
  Intelligence}, 31\penalty0 (9):\penalty0 1537--1551, 2009.

\bibitem[Rangapuram et~al.(2018)Rangapuram, Seeger, Gasthaus, Stella, Wang, and
  Januschowski]{rangapuram2018deep}
Syama~Sundar Rangapuram, Matthias~W Seeger, Jan Gasthaus, Lorenzo Stella,
  Yuyang Wang, and Tim Januschowski.
\newblock {Deep State Space Models for Time Series Forecasting}.
\newblock In \emph{Advances in Neural Information Processing Systems (NeurIPS)
  31}, pages 7785--7794, 2018.

\bibitem[Rezende et~al.(2014)Rezende, Mohamed, and
  Wierstra]{rezende2014stochastic}
Danilo~Jimenez Rezende, Shakir Mohamed, and Daan Wierstra.
\newblock {Stochastic Backpropagation and Approximate Inference in Deep
  Generative Models}.
\newblock In \emph{{International Conference on Machine Learning (ICML)}},
  pages 1278--1286, 2014.

\bibitem[Roberts et~al.(2013)Roberts, Osborne, Ebden, Reece, Gibson, and
  Aigrain]{roberts2013gpts}
Stephen Roberts, Michael Osborne, Mark Ebden, Steven Reece, Neale Gibson, and
  Suzanne Aigrain.
\newblock {Gaussian Processes for Time-Series Modelling}.
\newblock \emph{Philosophical Transactions of the Royal Society A:
  Mathematical, Physical and Engineering Sciences}, 371:\penalty0 20110550,
  2013.

\bibitem[Salinas et~al.(2017)Salinas, Flunkert, and
  Gasthaus]{salinas2017deepar}
David Salinas, Valentin Flunkert, and Jan Gasthaus.
\newblock {DeepAR: Probabilistic Forecasting with Autoregressive Recurrent
  Networks}.
\newblock In \emph{International Conference on Machine Learning (ICML)}, 2017.

\bibitem[S{\"a}rkk{\"a}(2013)]{sarkka2013bayesian}
Simo S{\"a}rkk{\"a}.
\newblock \emph{{Bayesian Filtering and Smoothing}}, volume~3.
\newblock Cambridge University Press, 2013.

\bibitem[Schnider et~al.(1998)Schnider, Minto, Gambus, Andresen, Goodale,
  Shafer, and Youngs]{schnider1998influence}
Thomas~W Schnider, Charles~F Minto, Pedro~L Gambus, Corina Andresen, David~B
  Goodale, Steven~L Shafer, and Elizabeth~J Youngs.
\newblock {The Influence of Method of Administration and Covariates on the
  Pharmacokinetics of Propofol in Adult Volunteers}.
\newblock \emph{The Journal of the American Society of Anesthesiologists},
  88\penalty0 (5):\penalty0 1170--1182, 1998.

\bibitem[Schulam and Saria(2015)]{schulam2015framework}
Peter Schulam and Suchi Saria.
\newblock {A Framework for Individualizing Predictions of Disease Trajectories
  by Exploiting Multi-Resolution Structure}.
\newblock In \emph{Advances in Neural Information Processing Systems (NeurIPS)
  29}, pages 748--756, 2015.

\bibitem[Schuller et~al.(2015)Schuller, Newell, Strickland, and
  Barry]{schuller2015response}
PJ~Schuller, S~Newell, PA~Strickland, and JJ~Barry.
\newblock {Response of Bispectral Index to Neuromuscular Block in Awake
  Volunteers}.
\newblock \emph{British Journal of Anaesthesia}, 115, 2015.

\bibitem[Soleimani et~al.(2017)Soleimani, Subbaswamy, and
  Saria]{soleimani2017treatment}
Hossein Soleimani, Adarsh Subbaswamy, and Suchi Saria.
\newblock {Treatment-Response Models for Counterfactual Reasoning with
  Continuous-Time, Continuous-Valued Interventions}.
\newblock \emph{arXiv preprint arXiv:1704.02038}, 2017.

\bibitem[Spieckermann et~al.(2015)Spieckermann, D{\"u}ll, Udluft, Hentschel,
  and Runkler]{spieckermann2015exploiting}
Sigurd Spieckermann, Siegmund D{\"u}ll, Steffen Udluft, Alexander Hentschel,
  and Thomas Runkler.
\newblock {Exploiting Similarity in System Identification Tasks with Recurrent
  Neural Networks}.
\newblock \emph{Neurocomputing}, 169:\penalty0 343--349, 2015.

\bibitem[Strogatz(2018)]{strogatz2018nonlinear}
Steven~H Strogatz.
\newblock \emph{{Nonlinear Dynamics and Chaos: with Applications to Physics,
  Biology, Chemistry, and Engineering}}.
\newblock CRC Press, 2018.

\bibitem[Sussillo and Barak(2013)]{sussillo2013opening}
David Sussillo and Omri Barak.
\newblock {Opening the Black Box: Low-Dimensional Dynamics in High-Dimensional
  Recurrent Neural Networks}.
\newblock \emph{Neural Computation}, 25:\penalty0 626--649, 2013.

\bibitem[Taylor et~al.(2010)Taylor, Sigal, Fleet, and
  Hinton]{taylor2010dynamical}
Graham~W Taylor, Leonid Sigal, David~J Fleet, and Geoffrey~E Hinton.
\newblock {Dynamical Binary Latent Variable Models for 3D Human Pose Tracking}.
\newblock In \emph{IEEE Conference on Computer Vision and Pattern Recognition
  (CVPR)}, 2010.

\bibitem[Titsias and L{\'a}zaro-Gredilla(2011)]{titsias2011spike}
Michalis~K Titsias and Miguel L{\'a}zaro-Gredilla.
\newblock {Spike and Slab Variational Inference for Multi-Task and Multiple
  Kernel Learning}.
\newblock In \emph{Advances in Neural Information Processing Systems (NeurIPS)
  24}, pages 2339--2347, 2011.

\bibitem[Tomasetti et~al.(2019)Tomasetti, Forbes, Panagiotelis,
  et~al.]{tomasetti2019updating}
Nathaniel Tomasetti, Catherine Forbes, Anastasios Panagiotelis, et~al.
\newblock {Updating Variational Bayes: Fast Sequential Posterior Inference}.
\newblock Technical report, Monash University, Department of Econometrics and
  Business Statistics, 2019.

\bibitem[Toqu{\'e} et~al.(2016)Toqu{\'e}, C{\^o}me, El~Mahrsi, and
  Oukhellou]{toque2016forecasting}
Florian Toqu{\'e}, Etienne C{\^o}me, Mohamed~Khalil El~Mahrsi, and Latifa
  Oukhellou.
\newblock {Forecasting Dynamic Public Transport Origin-Destination Matrices
  with Long-Short Term Memory Recurrent Neural Networks}.
\newblock In \emph{IEEE International Conference on Intelligent Transportation
  Systems (ITSC)}. IEEE, 2016.

\bibitem[Transtrum et~al.(2011)Transtrum, Machta, and
  Sethna]{transtrum2011geometry}
Mark~K Transtrum, Benjamin~B Machta, and James~P Sethna.
\newblock {The Geometry of Nonlinear Least Squares with Applications to Sloppy
  Models and Optimization}.
\newblock \emph{Physical Review E}, 83\penalty0 (3):\penalty0 036701, 2011.

\bibitem[Tsimikas and Ledolter(1997)]{tsimikas1997mixed}
John~V Tsimikas and Johannes Ledolter.
\newblock {Mixed Model Representation of State Space Models: New Smoothing
  Results and their Application to REML Estimation}.
\newblock \emph{Statistica Sinica}, 7:\penalty0 973--991, 1997.

\bibitem[Tulyakov et~al.(2018)Tulyakov, Liu, Yang, and
  Kautz]{tulyakov2018mocogan}
Sergey Tulyakov, Ming-Yu Liu, Xiaodong Yang, and Jan Kautz.
\newblock {MoCoGAN: Decomposing Motion and Content for Video Generation}.
\newblock In \emph{IEEE Conference on Computer Vision and Pattern Recognition
  (CVPR)}, pages 1526--1535, 2018.

\bibitem[Van~der Maaten and Hinton(2008)]{van2008visualizing}
Laurens Van~der Maaten and Geoffrey Hinton.
\newblock {Visualizing Data Using t-SNE}.
\newblock \emph{Journal of Machine Learning Research}, 9\penalty0 (11), 2008.

\bibitem[Van~Hese et~al.(2020)Van~Hese, Theys, Absalom, Rex, and
  Cuypers]{van2020comparison}
L~Van~Hese, T~Theys, AR~Absalom, S~Rex, and E~Cuypers.
\newblock {Comparison of Predicted and Real Propofol and Remifentanil
  Concentrations in Plasma and Brain Tissue During Target-Controlled Infusion:
  a Prospective Observational Study}.
\newblock \emph{Anaesthesia}, 75:\penalty0 1626--1634, 2020.

\bibitem[Villegas et~al.(2017)Villegas, Yang, Hong, Lin, and
  Lee]{villegas2017decomposing}
Ruben Villegas, Jimei Yang, Seunghoon Hong, Xunyu Lin, and Honglak Lee.
\newblock {Decomposing Motion and Content for Natural Video Sequence
  Prediction}.
\newblock In \emph{International Conference on Learning Representations
  (ICLR)}, 2017.

\bibitem[Wagner(1968)]{wagner1968kinetics}
JG~Wagner.
\newblock {Kinetics of Pharmacologic Response I. Proposed Relationships Between
  Response and Drug Concentration in the Intact Animal and Man}.
\newblock \emph{Journal of Theoretical Biology}, 20\penalty0 (2):\penalty0
  173--201, 1968.

\bibitem[Wang et~al.(2008)Wang, Fleet, and Hertzmann]{wang2008gaussian}
Jack~M Wang, David~J Fleet, and Aaron Hertzmann.
\newblock {Gaussian Process Dynamical Models for Human Motion}.
\newblock \emph{IEEE Transactions on Pattern Analysis and Machine Intelligence
  (PAMI)}, 30\penalty0 (2):\penalty0 283--298, 2008.

\bibitem[White et~al.(2008)White, Kenny, and Schraag]{white2008use}
Martin White, Gavin~NC Kenny, and Stefan Schraag.
\newblock {Use of Target Controlled Infusion to Derive Age and Gender
  Covariates for Propofol Clearance}.
\newblock \emph{Clinical Pharmacokinetics}, 47\penalty0 (2):\penalty0 119--127,
  2008.

\bibitem[Xu and Jordan(1996)]{xu1996convergence}
Lei Xu and Michael~I Jordan.
\newblock {On Convergence Properties of the EM Algorithm for Gaussian
  Mixtures}.
\newblock \emph{Neural Computation}, 8\penalty0 (1):\penalty0 129--151, 1996.

\bibitem[Xu et~al.(2016)Xu, Xu, and Saria]{xu2016bayesian}
Yanbo Xu, Yanxun Xu, and Suchi Saria.
\newblock {A Bayesian Nonparametric Approach for Estimating Individualized
  Treatment-Response Curves}.
\newblock In \emph{Machine Learning for Healthcare Conference}, 2016.

\bibitem[Yingzhen and Mandt(2018)]{yingzhen2018disentangled}
Li~Yingzhen and Stephan Mandt.
\newblock {Disentangled Sequential Autoencoder}.
\newblock In \emph{International Conference on Machine Learning (ICML)}, pages
  5656--5665, 2018.

\bibitem[Zhou et~al.(2013)Zhou, Han, and Liu]{zhou2013nonlinear}
Jie Zhou, Lu~Han, and Sanyang Liu.
\newblock {Nonlinear Mixed-Effects State Space Models with Applications to HIV
  Dynamics}.
\newblock \emph{Statistics \& Probability Letters}, 83\penalty0 (5):\penalty0
  1448--1456, 2013.

\end{thebibliography}

\end{document}